\documentclass{article}

 \usepackage[main, final]{neurips_2026}

\usepackage{authblk}
\usepackage[utf8]{inputenc} %
\usepackage[T1]{fontenc}    %
\usepackage[pagebackref=true,colorlinks,citecolor=brown]{hyperref}
\usepackage{url}            %
\usepackage{booktabs}       %
\usepackage{amsfonts}       %
\usepackage{nicefrac}       %
\usepackage{microtype}      %
\usepackage[table]{xcolor}         %

\usepackage{graphicx}
\usepackage{multirow}
\usepackage{amsmath}
\usepackage{float}
\usepackage{amssymb}
\usepackage{tabularx}
\usepackage{titlesec}
\titlespacing\section{2pt}{3pt plus 1pt minus 1pt}{2pt plus 1pt minus 1pt}
\titlespacing\subsection{2pt}{2pt plus 1pt minus 1pt}{1pt plus 1pt minus 1pt}

\renewcommand{\paragraph}[1]{\noindent\textbf{#1}}
\usepackage[font=small]{caption}

\usepackage{array}
\newcolumntype{M}{>{\centering\arraybackslash}X}
\newcolumntype{C}{>{\centering\arraybackslash}m{\dimexpr(\linewidth-1.5cm-6\tabcolsep)/9\relax}}
\usepackage{tikz}
\usetikzlibrary{spy, calc}
\definecolor{myPurple}{HTML}{8000FF}
\newcolumntype{Z}{>{\centering\arraybackslash}p{0.8cm}}
\usepackage{stackengine}

\tikzset{
    nvsspy/.style={
        spy using outlines={rectangle,lens={scale=3.5},size=1.05cm}
    }
}

\title{UGG: Unified Generative Geometry Model}

\author{%
  \textbf{Amani~Kiruga$^{1,}$\thanks{Equal contribution.}}\quad
  \textbf{Zhiyi~Li$^{2,*}$}\quad
  \textbf{Ruojin~Cai$^{1,*}$}\quad
  \textbf{Hansen~Lillemark$^3$}\quad
  \textbf{Xiaoshan~Wu$^1$} \\
  \textbf{Ayush~Tewari$^4$}\quad
  \textbf{Yilun~Du$^1$} \\
  \vspace{0.5em}
  $^1$Harvard University \quad $^2$MIT \quad $^3$UC San Diego \quad $^4$University of Cambridge
}

\begin{document}

\maketitle

\vspace{-10pt}
\begin{abstract}
\vspace{-5pt}
Understanding 4D scenes is a core challenge in visual intelligence, requiring models to jointly capture scene geometry, appearance, and dynamics from partial observations. However, existing approaches are subdivided into feedforward models for direct geometric prediction, and generative models for visual synthesis. 
We introduce UGG, a unified framework for 4D scene understanding that jointly models RGB video and geometric representations such as depth and camera pose. 
By assigning different noise levels across modalities and time, UGG formulates visual generation and geometric prediction as conditional completion within a single generative model.
This flexible formulation enables the model to learn from diverse, heterogeneous datasets with different subsets of geometric and visual annotations. Furthermore, our generative formulation enables inference-time search over multiple plausible scene hypotheses, using geometric evidence extracted from observed RGB frames to improve geometric prediction without ground truth geometry. 
We demonstrate UGG's effectiveness over non-unified baselines across camera-controlled video generation, depth estimation, and camera pose prediction, including in challenging sparse observation settings. Code and visualizations are available at \href{https://ugg-model.github.io/}{ugg-model.github.io}.
\end{abstract}

\vspace{-10pt}
\section{Introduction}
\vspace{-3pt}

For embodied agents to perceive and act intelligently in the physical world, they must understand its 3D structure and predict how it will evolve over time~\citep{chen2025LVP,ye2026dreamzero,bi2025motus}. However, visual observations provide only partial 2D projections of the world, each capturing a limited set of viewpoints, visible surfaces, and changes over time. Robust 4D scene understanding therefore requires inferring what is not directly observed, namely the 3D geometry, camera viewpoint, scene appearance from unseen views, and scene evolution. 

Current methods have made strong progress on different parts of this problem, but they are often separated into geometric prediction and visual generation. 
Feedforward models can directly estimate scene geometry, including depth, camera pose, and 3D structure, from images or videos. 
These methods are efficient and often accurate under the training distribution, but they typically produce a single deterministic estimate and can struggle when the observations are sparse or out of distribution. 
Generative models, in contrast, learn strong visual priors for completing missing observations, synthesizing novel views, and predicting plausible dynamic futures. 
However, they often lack explicit geometric grounding, which can lead to outputs that are visually plausible but inconsistent in 3D. 
Recent work has begun to narrow this gap by repurposing generative video models for geometric prediction \citep{jiang2025geo4d,mione4d2025,zhu2025motioncrafter}, or by injecting geometric representations into generative models to improve 3D consistency \citep{ren2025gen3c,wu2026geometryforcing,huang2025voyager}. Nevertheless, generation and prediction are still commonly treated as separate capabilities, rather than as two forms of inference in the same underlying scene model.

\looseness=-1
In this work, we introduce UGG, a unified generative geometry model that treats visual generation and geometric prediction as different conditional queries of the same underlying 4D scene representation. The appearance (RGB video), 3D structure (depth), and viewpoint (camera geometry) all provide complementary observations of a dynamic scene. Depending on what is observed, a joint model can infer missing scene components from the observed ones. 

\begin{figure}[t]
    \centering
    \includegraphics[width=0.98\textwidth]{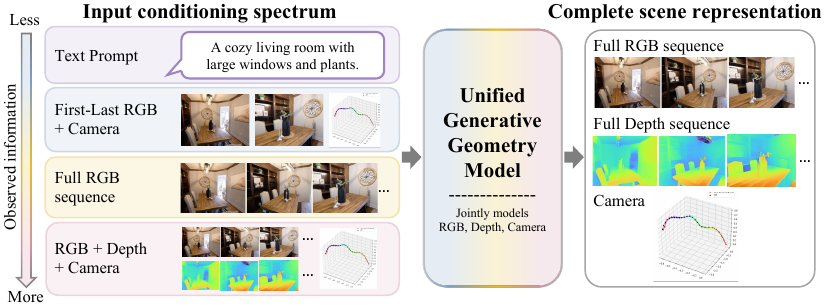}
    \vspace{-5pt}
    \caption{\small \textbf{A unified model for generation and prediction.} We formulate visual generation and geometric prediction as conditional completion in a single joint model over RGB, depth, and camera pose. Given arbitrary partial observations, ranging from a text prompt to RGB sequences with geometric cues, 
    our model jointly uses these modalities to infer a complete 4D scene representation, enabling both generation from sparse inputs and geometric prediction from observations.}
    \vspace{-18pt}
\end{figure}

We instantiate this formulation by representing RGB, depth, and camera geometry as spatially aligned modalities, with camera pose encoded as dense ray maps. A multimodal transformer jointly processes these representation through shared self-attention and modality-specific layers. Modality flexibility is achieved through an extension of Diffusion Forcing~\citep{chen2024diffusionforcing}: different tasks are specified through noise patterns over both modality and time, where observed components serve as conditioning context and missing components are denoised. This conditional completion mechanism enables the same model to condition on a text prompt, a single RGB frame, an RGB sequence, camera poses, depth maps, or any combination of these, and complete the remaining modalities. Moreover, the model can learn from heterogeneous datasets by applying the loss only on available modalities, instead of requiring every example to contain complete annotations of pose and depth. 

Our generative formulation also enables inference-time search over multiple plausible scene completions. Instead of committing to a single geometric prediction, we sample multiple candidates and select the one with the strongest cross-modal consistency, measured through photometric reprojection. This inference-time search capability is also consistent with the cognitive theory of perception as inference: human scene understanding requires inferring latent scene causes from incomplete observations, and may rely on internal simulation to evaluate possible scene explanations~\citep{kersten2004object,yuille2006vision,battaglia2013simulation,olshausen2014perceptionasinference}.

Overall, our contributions are threefold: 
\textbf{i)} We formulate 4D scene understanding as joint generative modeling over appearance and geometry, treating visual generation and geometric prediction as missing-modality denoising within a single conditional model. 
\textbf{ii)} We introduce a unified generative-predictive model that represents RGB, depth, and camera pose as spatially aligned modalities, enabling arbitrary conditioning and learning from heterogeneous datasets with incomplete supervision. 
\textbf{iii)} Finally, we leverage the generative formulation for inference-time search over multiple 4D scene hypotheses, using geometric consistency to improve geometric prediction.

\vspace{-3pt}
\section{Related Work}
\vspace{-3pt}

 \paragraph{Generative Modeling for 4D Scene Understanding.} Geometric prediction is largely dominated by feedforward discriminative models that regress depth, camera pose, and scene flow, or 3D structure from visual observations~\cite{wang2023dust3r, leroy2024groundingimagematching3d, zhang2024monst3r, wang2025vggt, shen2025fastvggt, yang2025fast3r, wang2025pi3, keetha2025mapanythinguniversalfeedforwardmetric}. While these methods achieve strong performance, they typically produce deterministic, single-pass estimates that struggle with inherent visual ambiguities, including sparse-view observations. To address these limitations, 
 recent work has explored generative priors for pose estimation~\cite{cai2025interpose, zhang2024raydiffusion}, depth prediction~\cite{ke2023marigold, hu2024depthcrafter, shao2025chronodepth}, and dynamic 4D reconstruction~\cite{jiang2025geo4d, zhu2025motioncrafter}.
 Generative models have also been used for novel view synthesis~\cite{yu2024viewcrafter, you2024nvs}.
 However, these methods typically treat geometric prediction as a unidirectional mapping from RGB to specific geometric outputs and rely on strictly paired datasets.

\paragraph{Geometry-aware Video and 4D generation.}
Recent generative world models have advanced from purely visual synthesis~\cite{blattmann2023SVD, wan2025wan, genie3, chen2025LVP} toward geometry-aware generation. 
One line of work injects 3D scene structure or persistent spatial memory into video generation to improve spatial consistency~\cite{ren2025gen3c, zhang2024worldWVD, huang2025voyager, wu2025video, zhou2025learning, lillemark2026flowequivariantworldmodels, yang2026neoverse}. 
Another line studies camera-conditioned generation, using camera trajectories to control synthesized videos~\cite{he2024cameractrl, zhou2025seva, wang2024motionctrl}. 
More closely related, One4D~\cite{mione4d2025} and WorldReel~\cite{fang2025worldreel} jointly generate RGB frames and 4D geometric representations from text or RGB prompts. 
However, these methods typically follow fixed conditioning and output pathways. 
In contrast, we treat RGB, depth, and camera rays as symmetric modalities, enabling an any-to-any conditional completion framework for both generation and prediction.

\paragraph{Unified Conditional Modeling.}
Our work is related to flexible conditional generation, where diverse tasks are cast as missing-modality completion. 
Mixture-of-Transformers (MoT) architectures~\cite{liang2024mot}, such as Motus~\cite{bi2025motus} and DreamZero~\cite{ye2026dreamzero}, show that a single model can process heterogeneous modalities. 
For flexible conditioning, we build on Diffusion Forcing~\cite{chen2024diffusionforcing} and DFoT~\cite{song2025dfot}, which assign different noise levels to different tokens or frames. 
Recent works extend this paradigm for long-horizon video generation or embodied planning~\cite{chen2025skyreelsv2infinitelengthfilmgenerative, wang2026eva}; in contrast, we apply it to multimodal 4D scene understanding, treating RGB, depth, and camera geometry as symmetric components.
This probabilistic formulation also enables inference-time search. 
Motivated by test-time scaling~\cite{zhang2025ttsdiffusion,liu2025videot1}, we use additional inference compute to compare multiple candidate scene completions. 
We then select the candidate with stronger photometric reprojection consistency to improve geometric prediction.

\vspace{-3pt}
\section{Method}
\vspace{-3pt}

\begin{figure}[t]
    \centering
    \includegraphics[width=0.98\textwidth]{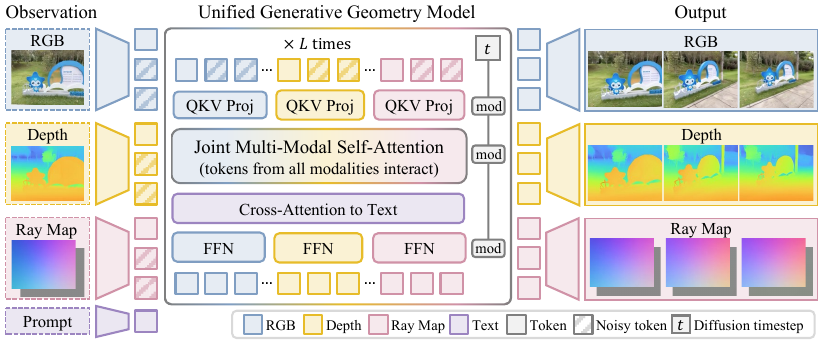}
    \caption{\textbf{Generative-predictive transformer.} 
    We model RGB, depth, and camera pose as spatially aligned tokens, with camera pose represented as dense ray maps.
    Partial observations are zero-noise context tokens, while missing modalities are noisy tokens to be denoised.
    A unified transformer processes all tokens with modality-specific projections and feed-forward layers together with shared multimodal self-attention, enabling information exchange across modalities. 
    The decoded RGB, depth, and ray maps form a complete 4D scene representation for cross-modal generation, geometric prediction, and inference-time consistency search.
    }
    \label{fig:method}
    \vspace{-17pt}
\end{figure}

We propose a unified generative-predictive framework for 4D scene understanding. Given partial observations of a scene, our goal is to infer a complete scene representation containing appearance, depth, and camera geometry. We formulate both visual generation and geometric prediction as conditional completion in a joint generative model over spatially aligned RGB, depth, and camera-ray modalities.
We specify each task by a noise pattern over multimodal tokens, with observed tokens as zero-noise context and missing tokens as noisy targets.
The framework supports heterogeneous training data with incomplete modality supervision, arbitrary partial conditioning, and test-time search over multiple 4D scene hypotheses.

\subsection{Unified 4D Scene Representation}

We represent a 4D scene as a multimodal sequence of $N$ frames. Each frame $i$ contains an RGB image $I_i$, the depth map $D_i$, camera parameters $C_i=(K_i,T_i)$, where $K_i$ denotes the camera intrinsics and $T_i$ denotes the camera extrinsics. Only a subset of this information may be observed. 
An example may contain one or more RGB frames, optionally with depth or camera parameters, or only a text prompt. 
Our model is designed to infer missing scene components from arbitrary partial observations.

To spatially align camera geometry with RGB and depth, we represent camera parameters $C_i$ as a dense map of 6D Pl\"ucker rays, following prior ray-based scene and camera representations~\citep{sitzmann2021light,zhang2024raydiffusion}. This representation turns camera geometry into an image-like modality.
Each pixel $(u,v)$ is associated with a ray moment $\mathbf{m}_i(u,v)$ in a common world coordinate system, constructed from the viewing direction $\mathbf{d}_i(u,v)$ and the camera center $\mathbf{o}_i$.
\begin{equation}
    R_i(u,v) = \left(\mathbf{d}_i(u,v), \mathbf{m}_i(u,v)\right),
    \qquad
    \mathbf{m}_i(u,v) = \mathbf{o}_i \times \mathbf{d}_i(u,v).
\end{equation}
The resulting scene representation $\mathcal{S}$ contains the appearance and local 3D geometry for each frame $i$:
\[
    \mathcal{S} = \{(I_i, D_i, R_i)\}_{i=1}^{N}.
\]

\subsection{Generative-Predictive Transformer}
Given the spatially aligned representation $\mathcal{S}$, we design a multimodal diffusion transformer that jointly processes appearance, depth, and camera-ray geometry (Fig.~\ref{fig:method}). 

\paragraph{Aligned Modality Tokenization.}
RGB frames and depth maps are encoded into latent tokens using a VAE encoder, producing $z^I$ and $z^D$.
Before encoding, we normalize depth and camera rays using a shared scale factor, making reprojection and cross-frame geometric consistency well-defined.
Ray maps are downsampled to the latent spatial resolution and projected directly into ray map tokens, producing $z^{R_d}$ for ray-direction tokens and $z^{R_m}$ for ray-moment tokens.
This yields the complete multimodal token sequence,
\[
z = [z^I; z^D; z^{R_d}; z^{R_m}].
\]
Each token is processed with modality-specific patch embedding layers and a spatiotemporal RoPE positional encoding over latent spatial coordinates and the frame index \citep{su2024roformer}. If the VAE encoder performs temporal downsampling, then the number of latents and frames could differ; we omit this from our notation for simplicity and continue to refer to the number of frames or latents as $N$. 

At the output stage, RGB and depth tokens are decoded back to pixel space using the VAE decoder.
Ray map tokens are directly mapped back to dense ray maps, which can be converted back to camera parameters $C_i$ including rotation, translation, and intrinsics.

\paragraph{Diffusion Mixture-of-Transformers.} The backbone uses a mixture-of-transformers~\cite{liang2024mot} design. Each modality has its own input projections, allowing for modality-specific representation structure. After modality-specific projection, all modalities are concatenated and processed by shared self-attention, enabling cross-modal information exchange. Modalities are separated to modality-specific feed-forward layers and output heads. Optional text prompts are injected through cross-attention.

Each token has its own diffusion timestep. These timesteps are embedded and injected into transformer blocks through adaptive layer normalization modulation~\citep{peebles2023scalable}, allowing the model to process clean context tokens and noisy target tokens at different noise levels.

\begin{figure}[t]
    \centering
    \includegraphics[width=0.98\textwidth]{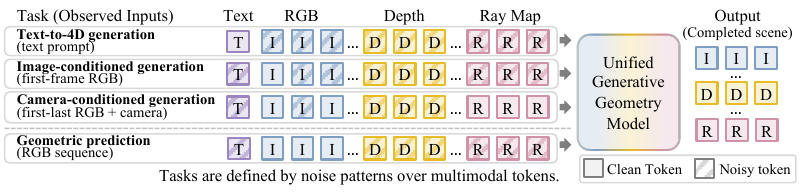}
    \vspace{-8pt}
    \caption{\small \textbf{Conditional completion with noise as masking.}
We specify each task by the clean/noisy pattern over RGB, depth, camera-ray, and text tokens. 
Clean tokens provide observed context, while noisy tokens are denoised as missing targets, enabling generation and prediction with the same model architecture.}
\label{fig:mask}
    \vspace{-15pt}
\end{figure}

\subsection{Conditional Completion with Noise as Masking}

We formulate generation and geometric prediction as conditional denoising over a shared multimodal representation. Let $\mathcal{M}=\{I,D,R_d,R_m\}$ denote the set of modalities, and let $z_{i,0}^{m}$ denote the clean latent token at temporal position $i$ for modality $m$. Which modalities are revealed or corrupted specify different tasks without changing the architecture. For example, RGB-conditioned geometric prediction reveals RGB tokens while treating depth and camera-ray tokens as prediction targets.

\paragraph{Task Specification by Noise Pattern.} Diffusion forcing enables flexible conditional generation by assigning independently sampled noise levels across temporal positions~\citep{chen2024diffusionforcing}. We extend this formulation to multimodal sequences, allowing noise levels $t_i^m \in [0,1]$ to vary across both modalities ($m$) and temporal positions ($i$). We construct the noisy input $\tilde z_{i,t}^{m}$ through linear interpolation of the clean token $z_{i,0}^m$ and Gaussian noise:
\begin{equation}
\tilde z_{i,t}^{m}
=
(1-t_i^{m})z_{i,0}^{m}
+
t_i^{m}\epsilon_i^{m},
\qquad \epsilon_i^m \sim \mathcal{N}(0,I).
\end{equation}
At $t_i^m=0$, the token is fully clean, while at $t_i^m=1$, it is pure noise. By varying the noise levels across both modality and temporal position, we realize different conditional generation and prediction tasks without modifying the model architecture.

\paragraph{Training with Heterogeneous Data.}
During training, we sample conditioning patterns that designate subsets of modality tokens as context, and others as prediction targets. Context tokens share a single noise level, while target tokens share a separately sampled noise level. With probability $p_{\mathrm{clean}}$, context noise levels are set to $t_i^m=0$ to allow for fully clean context conditioning. 

The model is trained using conditional Flow Matching~\citep{lipman2023flow}, with target velocity $v_i^m=\partial \tilde z_{i,t}^m/\partial t_i^m = \epsilon_i^m - z_{i,0}^m$. Given the corrupted multimodal tokens $\tilde z$, noise levels $t$, and optional text conditioning $c$, the model predicts $\hat{v} = f_\theta(\tilde{z}_t,t,c)$, with the loss:
\begin{equation}
\mathcal L_{\mathrm{FM}}
=
\mathbb E_{z,t,\epsilon}\left[
\sum_{i,\;m \in \mathcal{M}}
a_i^m \lambda^m
\left\|
\hat v_i^m-v_i^m
\right\|_2^2
\right].
\end{equation}
$\lambda^m$ is a weight that balances the losses for each modality. Modality availability is denoted as $a_i^m\in\{0,1\}$. The availability mask allows a single model to be trained using heterogeneous mixes of RGB-only, RGB-camera, RGB-depth, and RGB-depth-camera data.

\paragraph{Guided Inference Using Available Modalities.}
At inference time, the same noise pattern is used, but conditioning is stricter. 
Observed tokens are clamped to their clean values at every denoising step and assigned $t_i^m=0$, while unknown tokens are initialized from Gaussian noise and progressively denoised. For example, in a first-last-frame camera-controlled setting, the first and last RGB tokens and all camera ray tokens are clamped as context, while intermediate RGB and depth tokens are generated. We also use History Guidance~\cite{song2025dfot} to strengthen consistency with the observed context.

\subsection{Inference-time Search}
\label{sec:inference_time_search}

The generative formulation also enables inference-time search, in contrast with a deterministic predictor that commits to one scene. Given the same observed context, we sample $L$ candidate completions from different random initializations. Each sample yields a complete 4D scene hypothesis:
\[
\hat{\mathcal{S}}^{(\ell)}
=
\{(\hat{I}^{(\ell)}_i,\hat{D}^{(\ell)}_i,\hat{R}^{(\ell)}_i)\}_{i=1}^{N},
\qquad
\ell=1,\ldots,L.
\]
To select among these hypotheses without ground-truth depth or camera pose, we evaluate how well each hypothesis explains geometric evidence extracted from the observed RGB frames.
Full equations and hyperparameters are given in App.~\ref{app:tts}.

\paragraph{Geometric Evidence Derived from RGB.}
We extract pixel correspondences between observed RGB frames using RAFT~\citep{teed2020raft}. To remove issues with occlusions and moving objects, we keep correspondences only if forward mapping followed by backward mapping returns a pixel to its original point. For dense observations, we select frame pairs with informative pixel motion, adapting the temporal separation according to the observed optical flow magnitude. For sparse observations, both generation and verification use only the first and last observed RGB frames. These correspondences provide geometric constraints against which candidate depth and camera predictions can be evaluated.

\paragraph{Depth Evaluation.}
We evaluate whether a candidate's predicted depth is consistent with the observed pixel motion. For each frame pair, we first estimate the relative camera pose from image correspondences using essential matrix estimation with RANSAC, shared across depth candidates. Each candidate's inverse depth is aligned to the observed parallax by fitting an affine transformation. The aligned depth is used to reproject pixels between views, which measures agreement with the observed correspondences. For sparse observations, evaluation uses correspondence residuals after epipolar filtering. For dense observations, we additionally incorporate photometric and DINOv2 feature consistency~\citep{oquab2023dinov2}. 

\paragraph{Pose Evaluation.}
For camera pose prediction, we compare each candidate's relative translation direction against the direction estimated from image correspondences using the candidate's predicted intrinsics. We compute the angular discrepancy over selected frame pairs, measuring whether the predicted camera motion agrees with the observed images.

\paragraph{Final Candidate Selection.}
Depth and pose are selected independently: there is no trade-off between the two scores, and the depth and pose results may come from different candidates. For depth prediction, we standardize and combine the verification residuals, rank candidates within each frame pair, and then aggregate these rankings across pairs. For pose prediction, we use the mean angular discrepancy between predicted and image-estimated relative translation directions. Finally, the model selects the candidate $\ell^*$ with the lowest task-specific verification score.

\vspace{-3pt}
\section{Experiments}
\vspace{-3pt}
\begin{figure*}[t]
    \centering
    \footnotesize
    \renewcommand{\arraystretch}{0.5}
    \setlength{\tabcolsep}{0.5pt}

    \begin{tabularx}{\textwidth}{
        >{\centering\arraybackslash}m{1cm}
        @{\hspace{2pt}}
        *{3}{C} @{\hspace{2pt}}
        *{3}{C} @{\hspace{2pt}}
        *{3}{C}
    }
        & \multicolumn{3}{c}{\textbf{DL3DV-10K}} &
        \multicolumn{3}{c}{\textbf{SpatialVID}} &
        \multicolumn{3}{c}{\textbf{RealEstate10K}} \\[0pt]

        \textbf{Input} &
        \includegraphics[width=\linewidth]{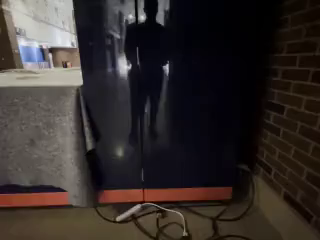} &
        \includegraphics[width=\linewidth]{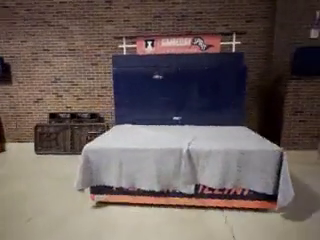} &
        \includegraphics[width=\linewidth]{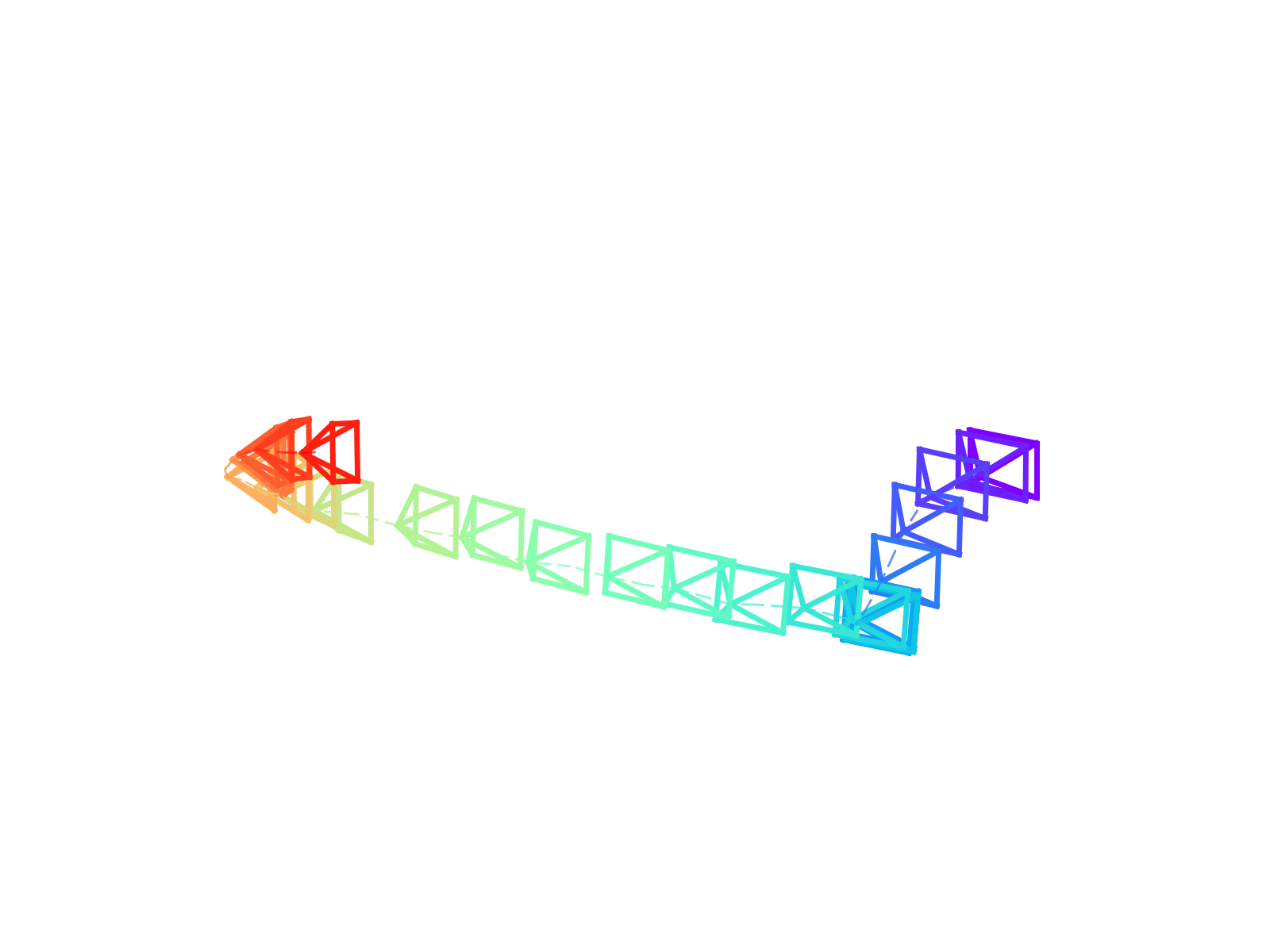} &
        \includegraphics[width=\linewidth]{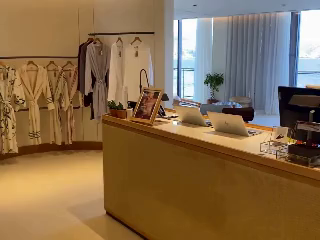} &
        \includegraphics[width=\linewidth]{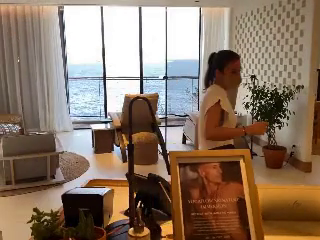} &
        \includegraphics[width=\linewidth]{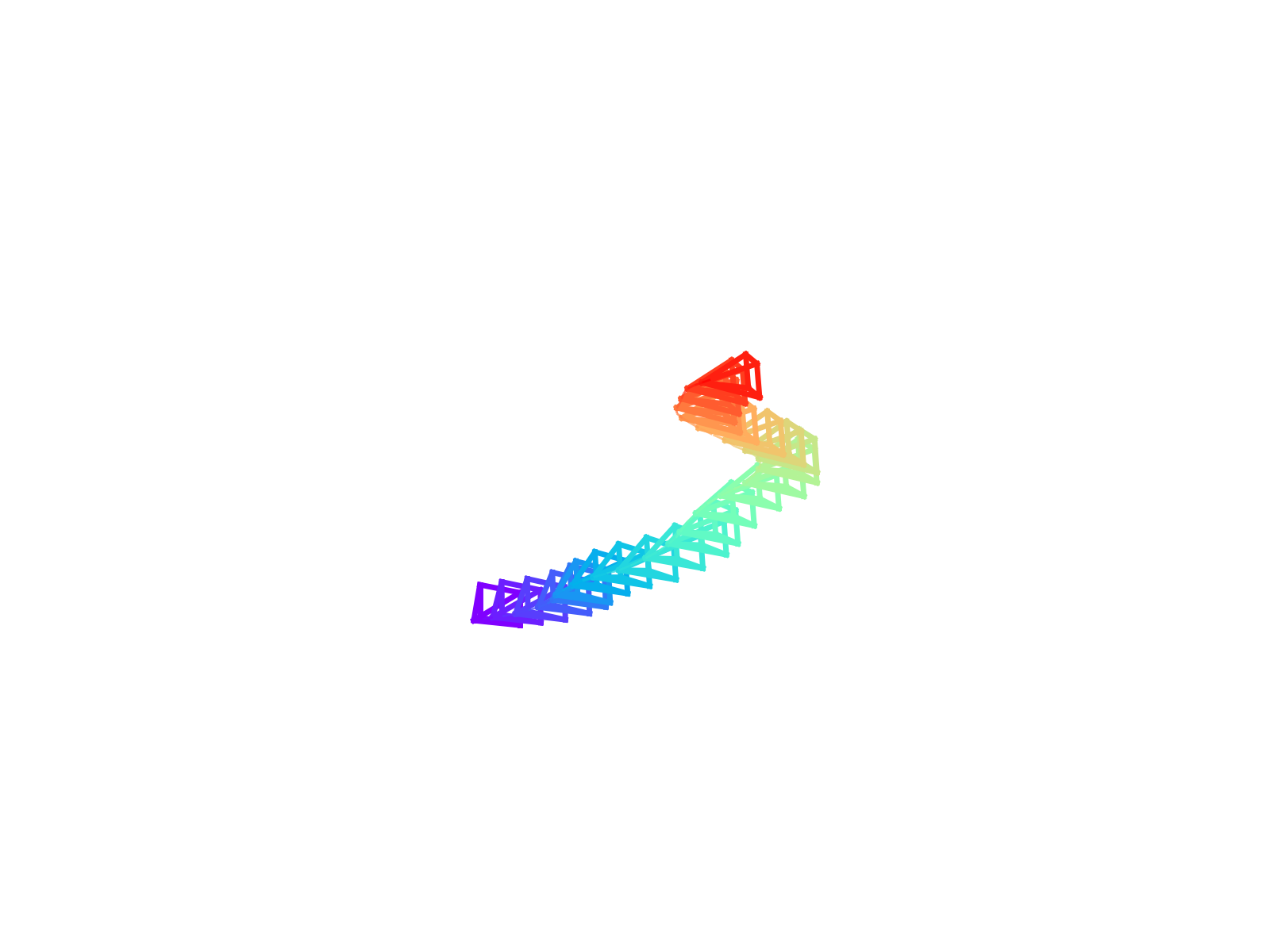} &
        \includegraphics[width=\linewidth]{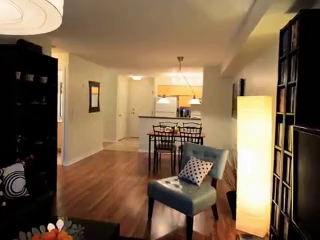} &
        \includegraphics[width=\linewidth]{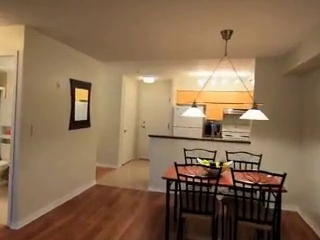} &
        \includegraphics[width=\linewidth]{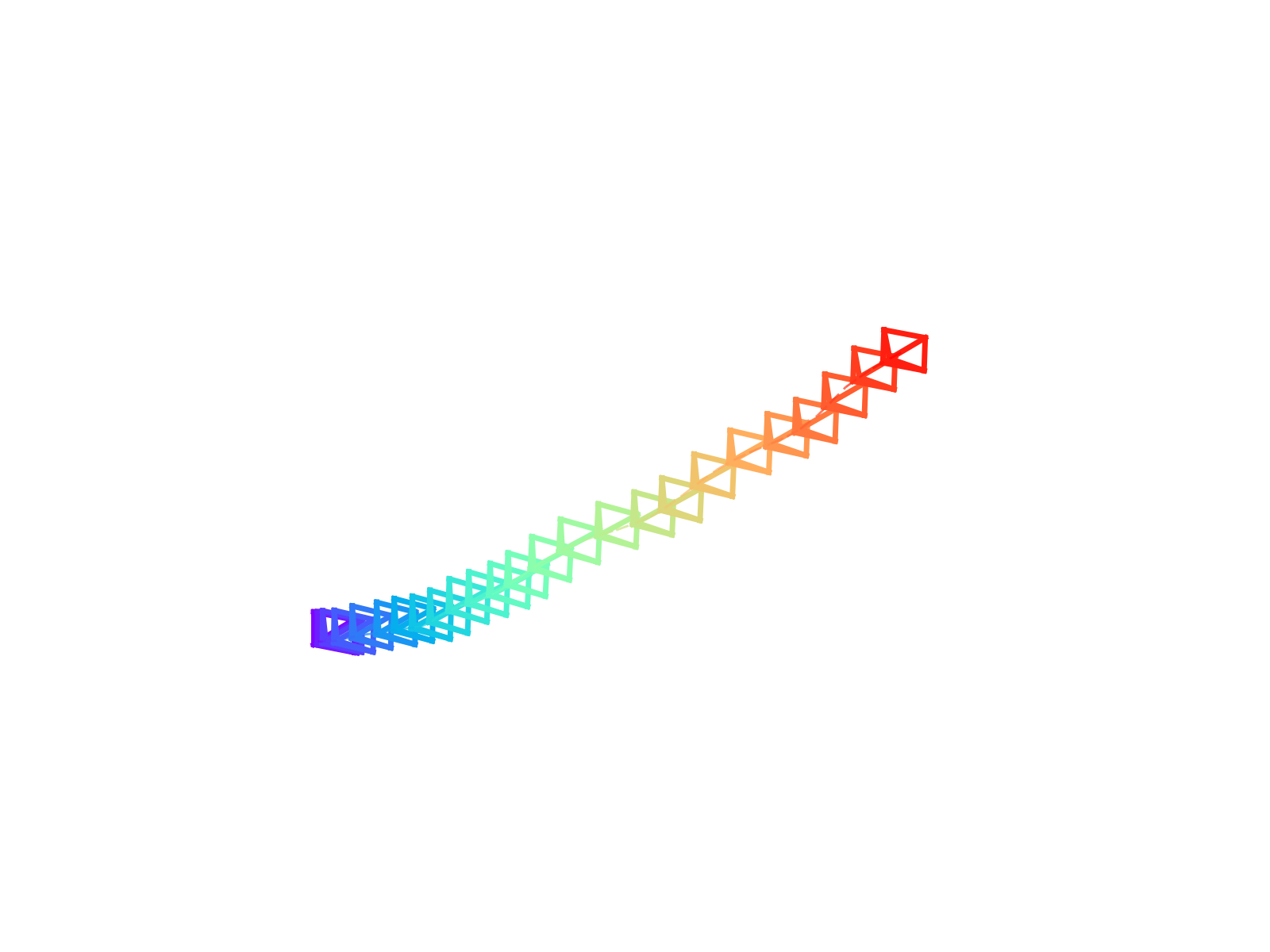} \\ [0pt]

        \textbf{GT} &
        \includegraphics[width=\linewidth]{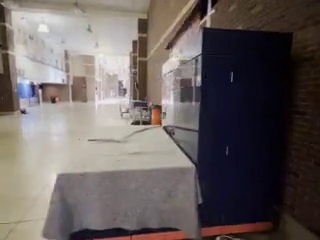} &
        \includegraphics[width=\linewidth]{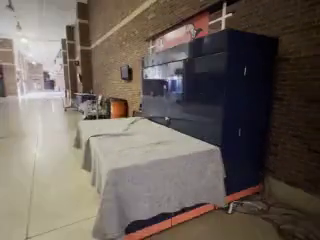} &
        \includegraphics[width=\linewidth]{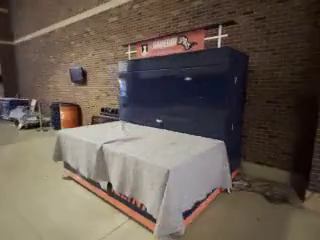} &
        \includegraphics[width=\linewidth]{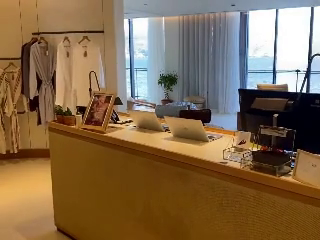} &
        \includegraphics[width=\linewidth]{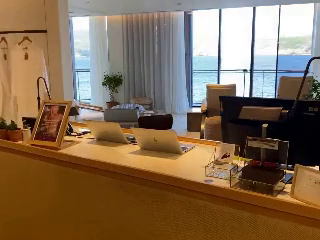} &
        \includegraphics[width=\linewidth]{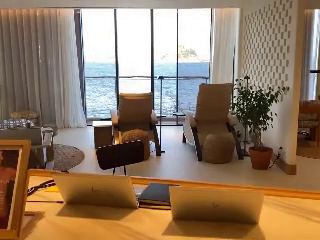} &
        \includegraphics[width=\linewidth]{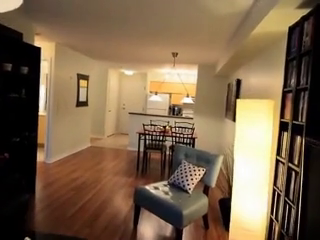} &
        \includegraphics[width=\linewidth]{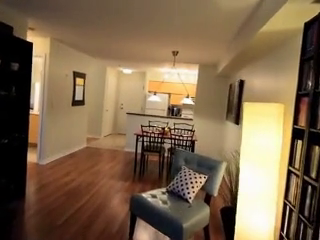} &
        \includegraphics[width=\linewidth]{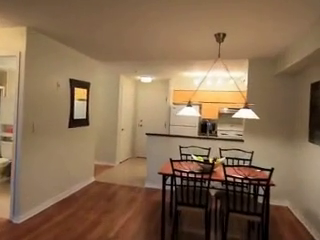} \\ [0pt]

        \textbf{\textcolor{myPurple}{Ours}} &
        \includegraphics[width=\linewidth]{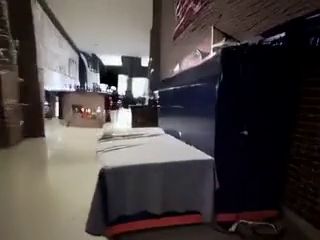} &
        \includegraphics[width=\linewidth]{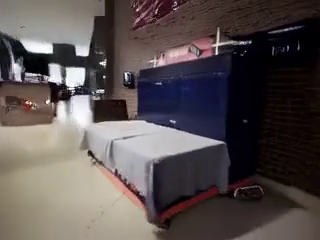} &
        \includegraphics[width=\linewidth]{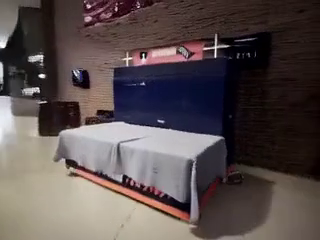} &
        \includegraphics[width=\linewidth]{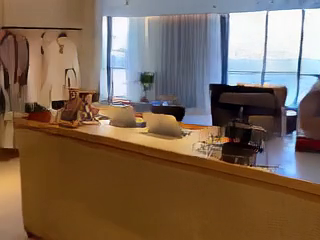} &
        \includegraphics[width=\linewidth]{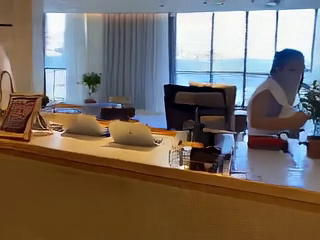} &
        \includegraphics[width=\linewidth]{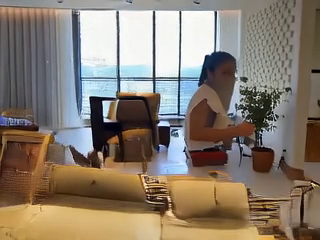} &
        \includegraphics[width=\linewidth]{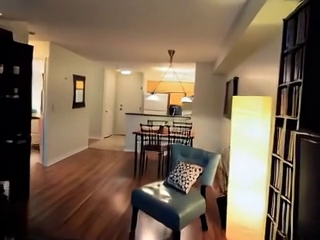} &
        \includegraphics[width=\linewidth]{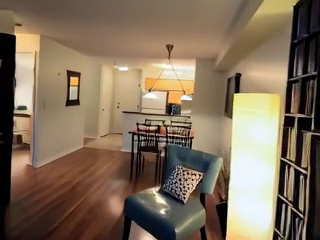} &
        \includegraphics[width=\linewidth]{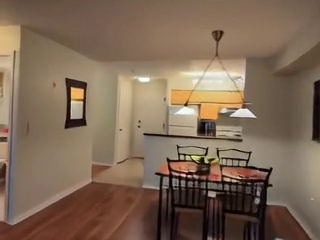} \\[0pt]

        \textbf{SEVA} &
        \includegraphics[width=\linewidth]{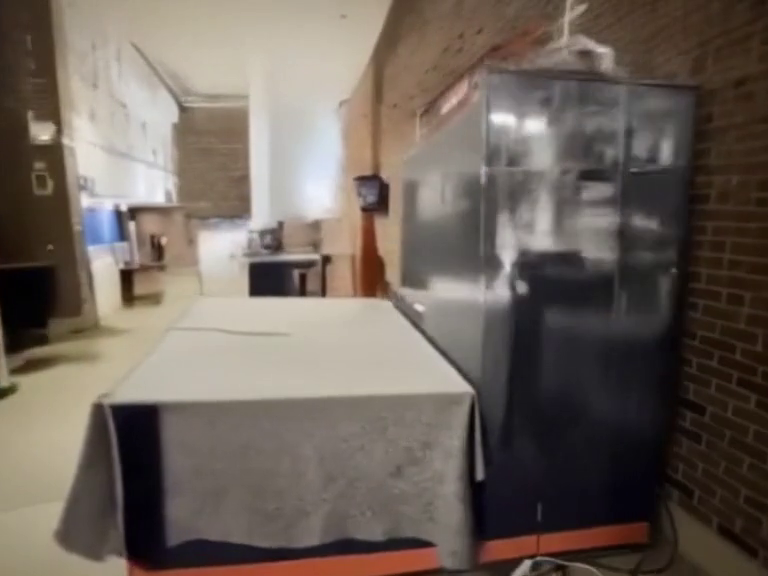} &
        \includegraphics[width=\linewidth]{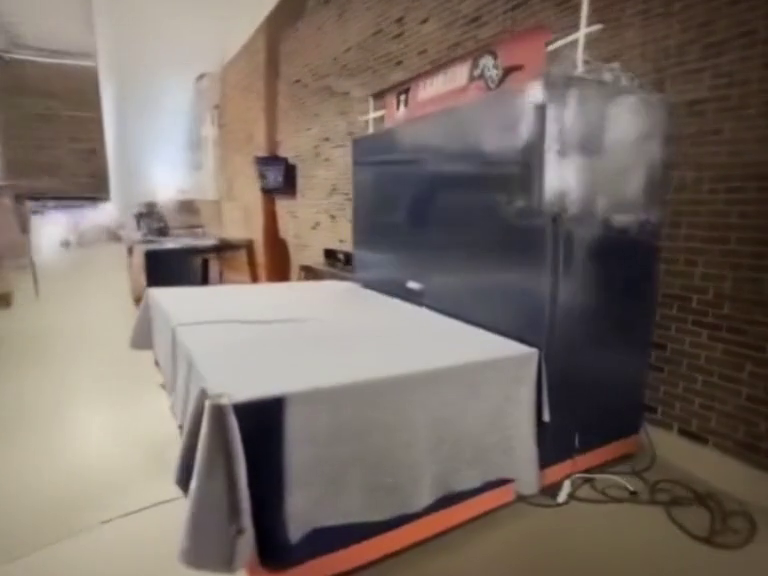} &
        \includegraphics[width=\linewidth]{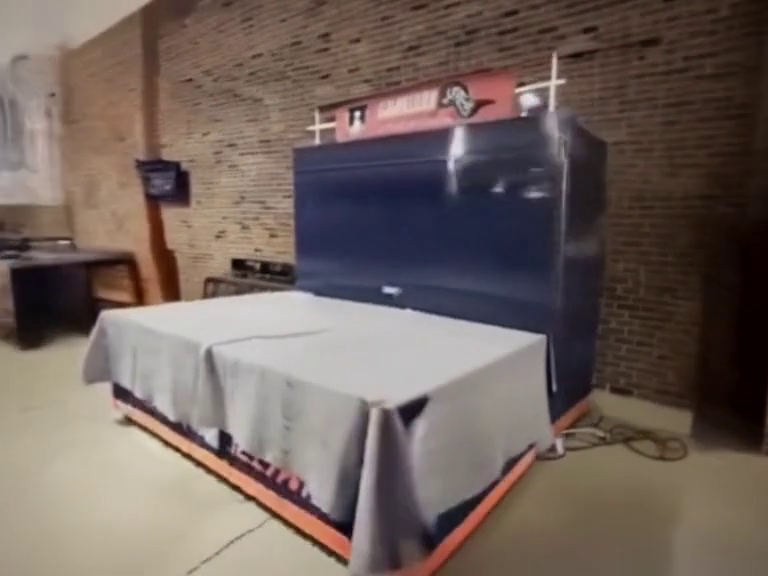} &
        \includegraphics[width=\linewidth]{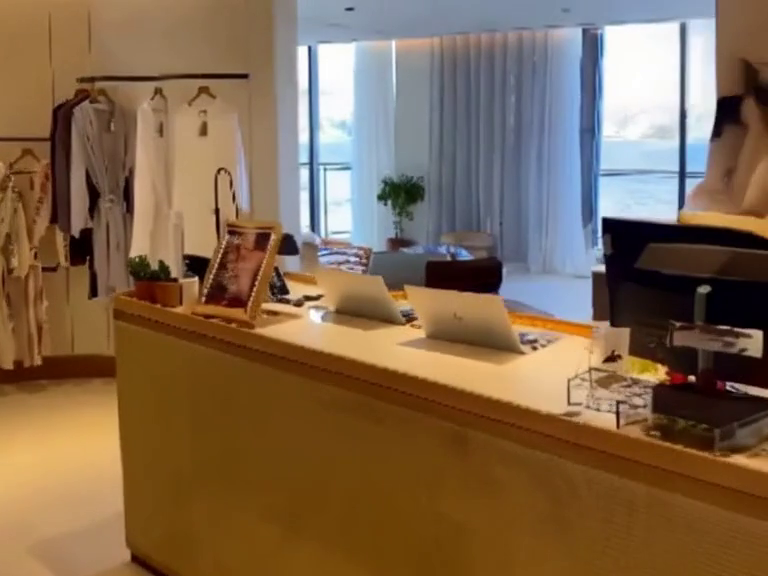} &
        \includegraphics[width=\linewidth]{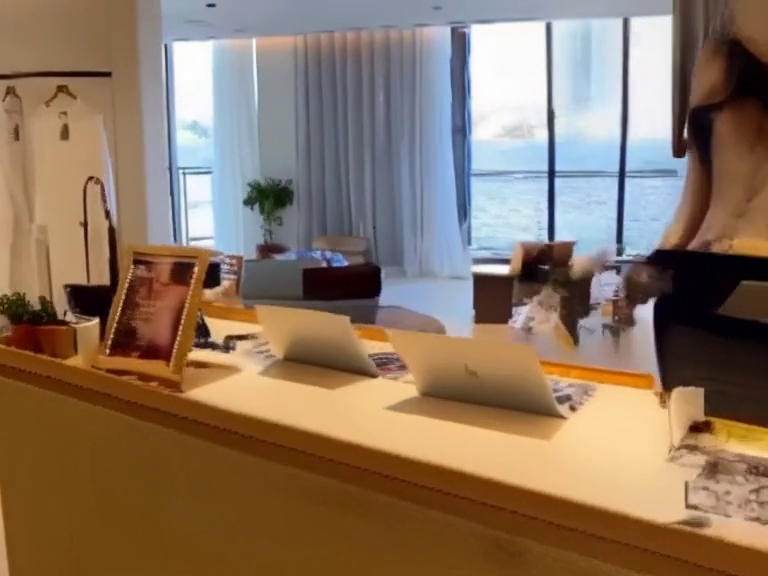} &
        \includegraphics[width=\linewidth]{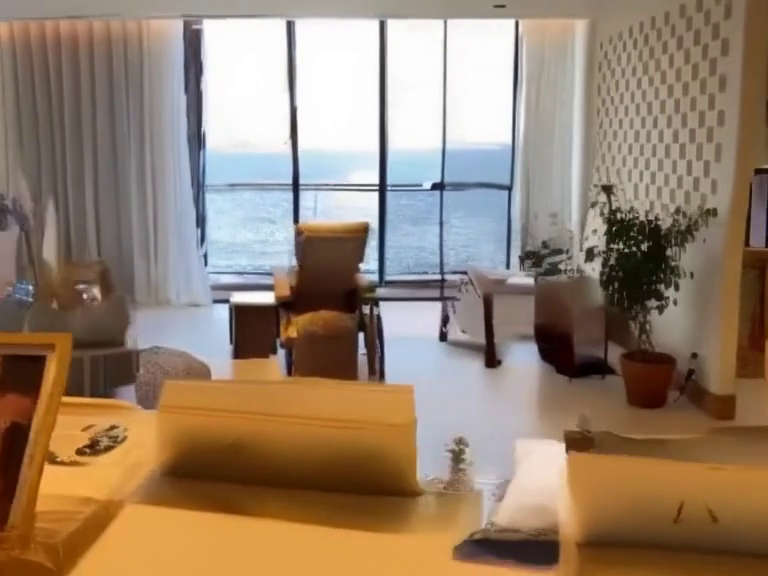} &
        \includegraphics[width=\linewidth]{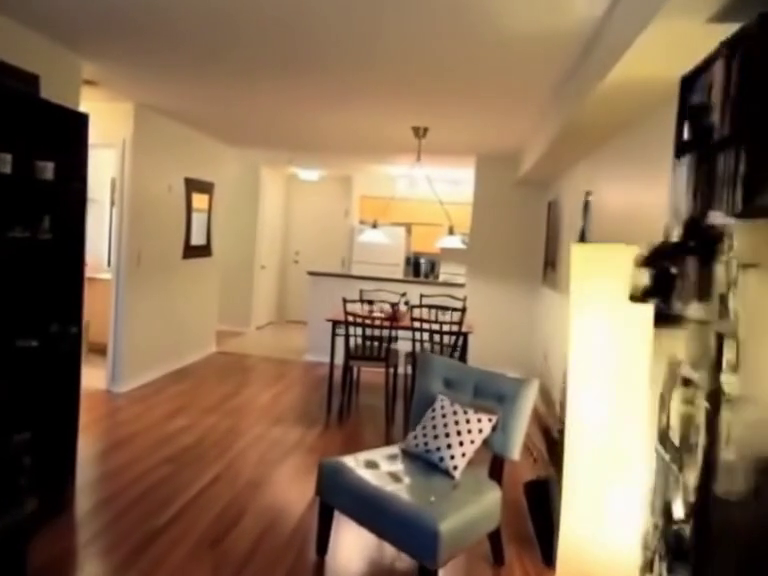} &
        \includegraphics[width=\linewidth]{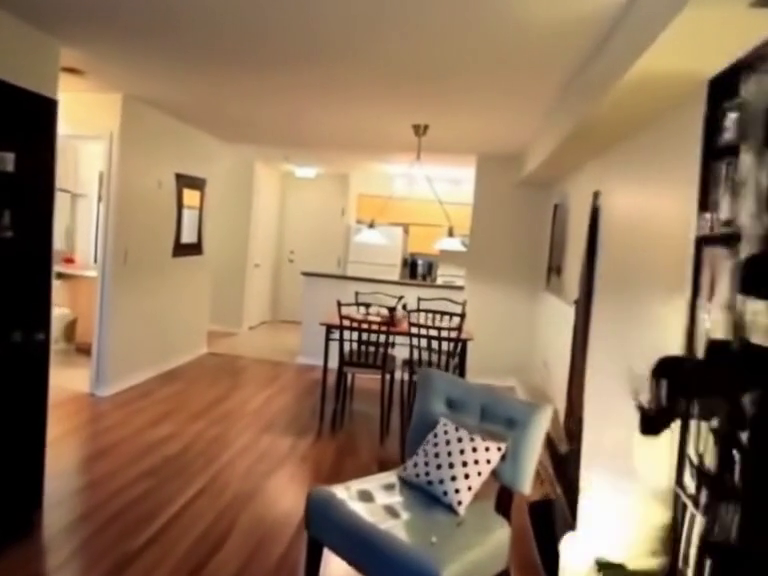} &
        \includegraphics[width=\linewidth]{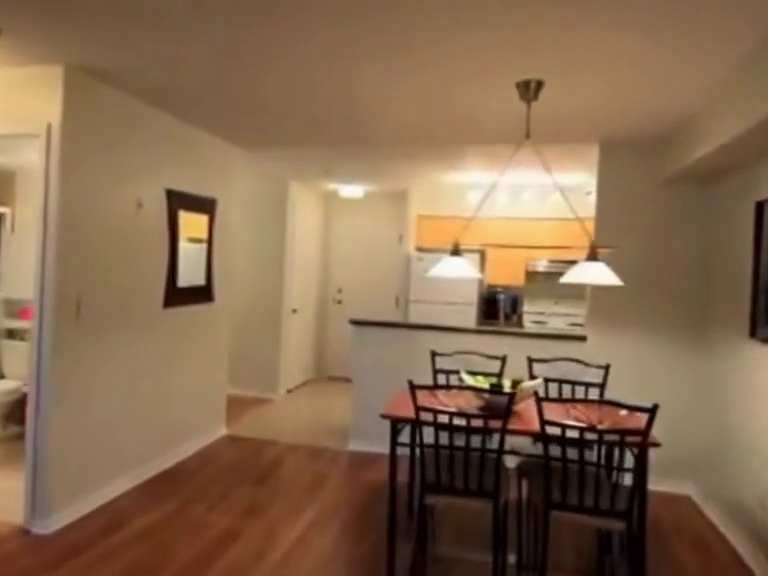} \\ [0pt]

        \textbf{WAN} &
        \includegraphics[width=\linewidth]{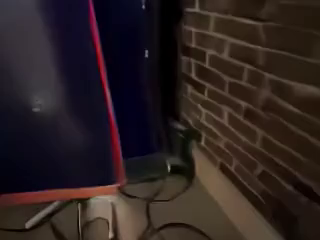} &
        \includegraphics[width=\linewidth]{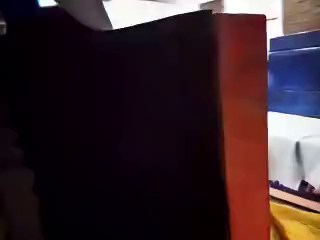} &
        \includegraphics[width=\linewidth]{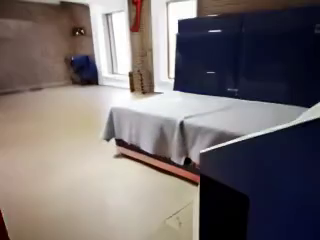} &
        \includegraphics[width=\linewidth]{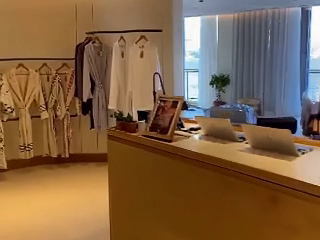} &
        \includegraphics[width=\linewidth]{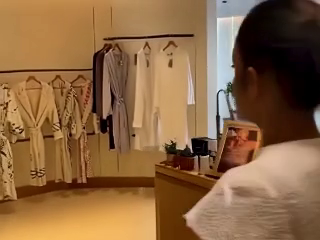} &
        \includegraphics[width=\linewidth]{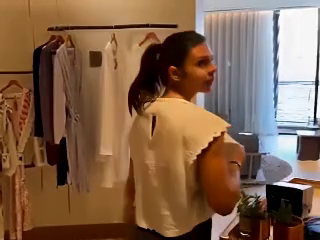} &
        \includegraphics[width=\linewidth]{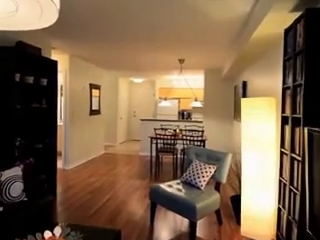} &
        \includegraphics[width=\linewidth]{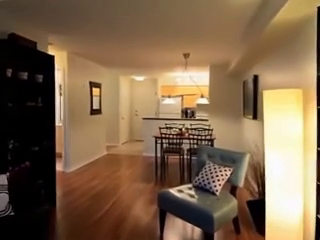} &
        \includegraphics[width=\linewidth]{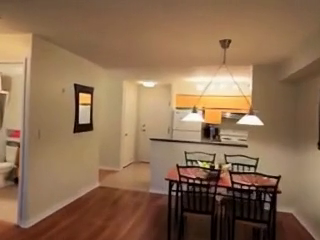} \\ [0pt]

        \textbf{Dfot} &
        \includegraphics[width=\linewidth]{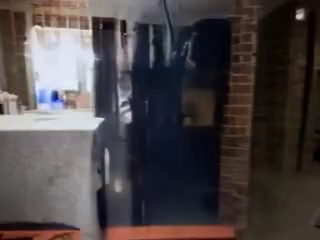} &
        \includegraphics[width=\linewidth]{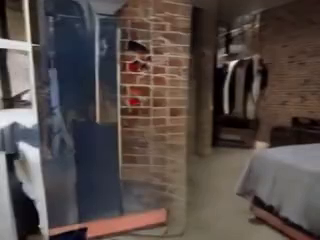} &
        \includegraphics[width=\linewidth]{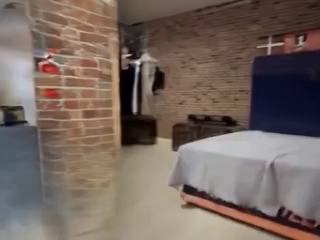} &
        \includegraphics[width=\linewidth]{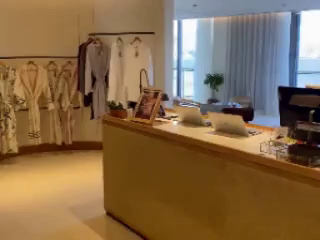} &
        \includegraphics[width=\linewidth]{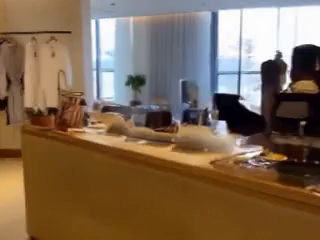} &
        \includegraphics[width=\linewidth]{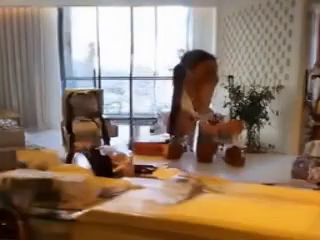} &
        \includegraphics[width=\linewidth]{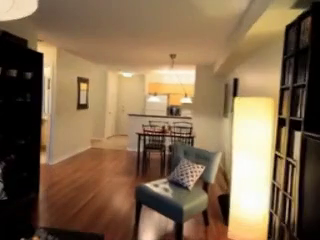} &
        \includegraphics[width=\linewidth]{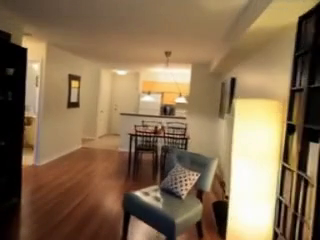} &
        \includegraphics[width=\linewidth]{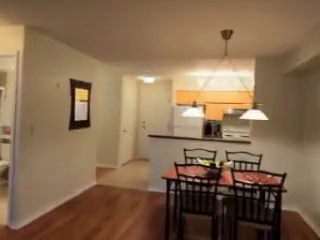} \\

    \end{tabularx}
    \caption{\textbf{Qualitative comparison of camera-controlled generation.} 
    Given the first and last RGB frames and a target camera trajectory, we compare generated intermediate frames against ground truth and baseline video generation methods. UGG better preserves scene identity and dynamic content compared with the generic video completion baselines.}
    \label{fig:nvs_results}
    \vspace{-18pt}
\end{figure*}

\label{sec:experiments}
We evaluate UGG on a suite of 4D scene understanding tasks, including camera-controlled generation (Sec.~\ref{sec:camera_generation}) and geometric prediction. Geometric prediction tasks under both dense and sparse observation settings include camera pose estimation (Sec.~\ref{sec:pose_estimation}) and video depth estimation (Sec.~\ref{sec:depth_estimation}). Our single unified model handles all tasks by changing only the clean/noisy masking pattern, without requiring task-specific architectures. We further show that UGG's generative formulation enables inference-time search, improving predicted scene consistency with additional samples (Sec.~\ref{sec:test_time_search}). Full code and additional visualizations are provided on the \href{https://ugg-model.github.io/}{project page}.

\vspace{-3pt}
\subsection{Model Training}
\vspace{-3pt}
We train on a weighted mixture of real and synthetic video datasets, including DL3DV-10K~\cite{ling2024dl3dv}, RealEstate10K~\cite{zhou2018realestate}, TartanAir~\cite{wang2020tartanair}, BEDLAM~\cite{black2023bedlam}, Aria Synthetic Environments~\cite{avetisyan2024scenescript}, OmniWorld-Game~\cite{zhou2025omniworld}, Virtual-KITTI2~\cite{cabon2020vkitti}, and Mid-Air~\cite{fonder2019midair}.
These datasets provide heterogeneous supervision over RGB video, depth, camera pose, and text; missing modalities are omitted from the loss. 
UGG is trained with multiple clean/noisy mask patterns over modality tokens to cover generation and geometric prediction tasks. 
UGG weights are initialized from Wan2.2-TI2V-5B, then finetuned on 50-frame clips at $[288\times384]$ resolution. The VAE and text encoder are frozen. Training and preprocessing details, including the full dataset list and mixture weights, are provided in the appendix.

\subsection{Camera-Controlled Generation}
\label{sec:camera_generation}

\providecommand{\best}[1]{\cellcolor{blue!12}#1}
\providecommand{\secondbest}[1]{\cellcolor{blue!5}#1}

\begin{table}[t]
\centering
\caption{\textbf{Camera-controlled generation results.}
Generation is conditioned on the first RGB frame (1-view) or the
first and last frames (2-view), together with the camera trajectory.
We report PSNR~($\uparrow$), LPIPS~($\downarrow$), and SSIM~($\uparrow$).
Darker and lighter blue shading indicate the best and second-best
results within each setting.}
\label{tab:final_nvs_table}

\begingroup

\fontsize{8.5}{9.5}\selectfont       %
\renewcommand{\arraystretch}{0.9}   %
\setlength{\aboverulesep}{0.2ex}    %
\setlength{\belowrulesep}{0.3ex}    %

\begin{tabular*}{\linewidth}{
  @{\extracolsep{\fill}} c l *{9}{r} @{}
}
\toprule
& \multirow{2}{*}{Method}
& \multicolumn{3}{c}{RealEstate10K}
& \multicolumn{3}{c}{SpatialVID}
& \multicolumn{3}{c}{DL3DV} \\
\cmidrule(lr){3-5}
\cmidrule(lr){6-8}
\cmidrule(lr){9-11}
& & PSNR & LPIPS & SSIM
  & PSNR & LPIPS & SSIM
  & PSNR & LPIPS & SSIM \\
\midrule

\multirow{3}{*}{\rotatebox[origin=c]{90}{2-view}}
& Aether
& 20.02 & 0.196 & 0.624
& 13.84 & 0.489 & 0.334
& 13.14 & 0.399 & 0.243 \\
& SEVA
& \secondbest{25.25} & \secondbest{0.081} & \secondbest{0.838}
& \secondbest{15.44} & \secondbest{0.344} & \secondbest{0.466}
& \best{17.61} & \best{0.211} & \best{0.552} \\
& UGG
& \best{25.73} & \best{0.061} & \best{0.849}
& \best{16.99} & \best{0.288} & \best{0.535}
& \secondbest{16.61} & \secondbest{0.239} & \secondbest{0.495} \\

\midrule

\multirow{3}{*}{\rotatebox[origin=c]{90}{1-view}}
& Aether
& 15.79 & 0.335 & 0.469
& 11.81 & 0.632 & 0.283
& 11.96 & 0.506 & 0.240 \\
& SEVA
& \secondbest{19.58} & \secondbest{0.194} & \secondbest{0.703}
& \secondbest{13.35} & \secondbest{0.472} & \secondbest{0.343}
& \secondbest{13.50} & \secondbest{0.393} & \secondbest{0.335} \\
& UGG
& \best{21.79} & \best{0.150} & \best{0.779}
& \best{14.86} & \best{0.428} & \best{0.405}
& \best{13.92} & \best{0.385} & \best{0.346} \\

\bottomrule
\end{tabular*}
\endgroup
\vspace{-8pt}

\end{table}

We evaluate the unified model on camera-controlled video generation, testing whether camera-ray
conditioning can guide visual synthesis while preserving appearance and scene layout across
viewpoint changes.

\paragraph{Benchmarks and Metrics.}
We evaluate on RealEstate10K, SpatialVID, and DL3DV, covering diverse indoor and
outdoor scenes. 
For each sequence, the context contains the target camera trajectory and either the first RGB frame or the first and last RGB frames. The model then generates the remaining frames from this context, which are evaluated against ground-truth videos using PSNR, SSIM, and LPIPS \citep{wang2004image,Zhang_2018_CVPR}. 

\textbf{Baselines.}
We compare quantitatively against Aether~\cite{zhu2025iccv-aether} and SEVA~\cite{zhou2025seva}, and qualitatively against DFoT~\cite{song2025dfot} and Wan2.1-FLF-14B in a first-last-frame conditioning setting~\cite{wan2025wan}. 
These baselines represent unified generation and geometry prediction, diffusion-based novel view synthesis, diffusion-forcing-based video completion, and endpoint-conditioned video generation, respectively. 
In contrast to the task-specific baselines,
UGG has the same multimodal architecture for both camera-controlled generation and geometry prediction.

\textbf{Evaluation results.}
Camera-controlled generation results are shown in Table~\ref{tab:final_nvs_table}. 
Although not trained solely for novel-view synthesis, UGG remains competitive across camera-conditioned generation benchmarks. 
It achieves the best PSNR, LPIPS, and SSIM on all three datasets in the one-view setting and on RealEstate10K and SpatialVID in the two-view setting; SEVA remains slightly better on two-view DL3DV. 
This shows that ray-map conditioning supports accurate camera control, even within a broader RGB-depth-camera completion model.
We visualize generated examples in Figure~\ref{fig:nvs_results}. 
Wan2.1-FLF does not take camera trajectories as input, and therefore struggles with complex camera motion. 
We find that UGG also preserves dynamic content better; SEVA is primarily designed for static scenes, and DFoT distorts or removes moving humans.

\begin{figure*}[tb]
    \centering
    \small
    \renewcommand{\arraystretch}{1.0}
    \setlength{\tabcolsep}{8pt}
    \setstackgap{S}{3pt}
    \begin{tabularx}{\textwidth}{CCCCCCCC}
         \textbf{Input Video} & \textbf{UGG} & \textbf{RayDiffusion} & \textbf{Geo4D} & \textbf{Input Video} & \textbf{UGG} & \textbf{RayDiffusion} & \textbf{Geo4D}\\[0pt]
        \includegraphics[width=\linewidth]{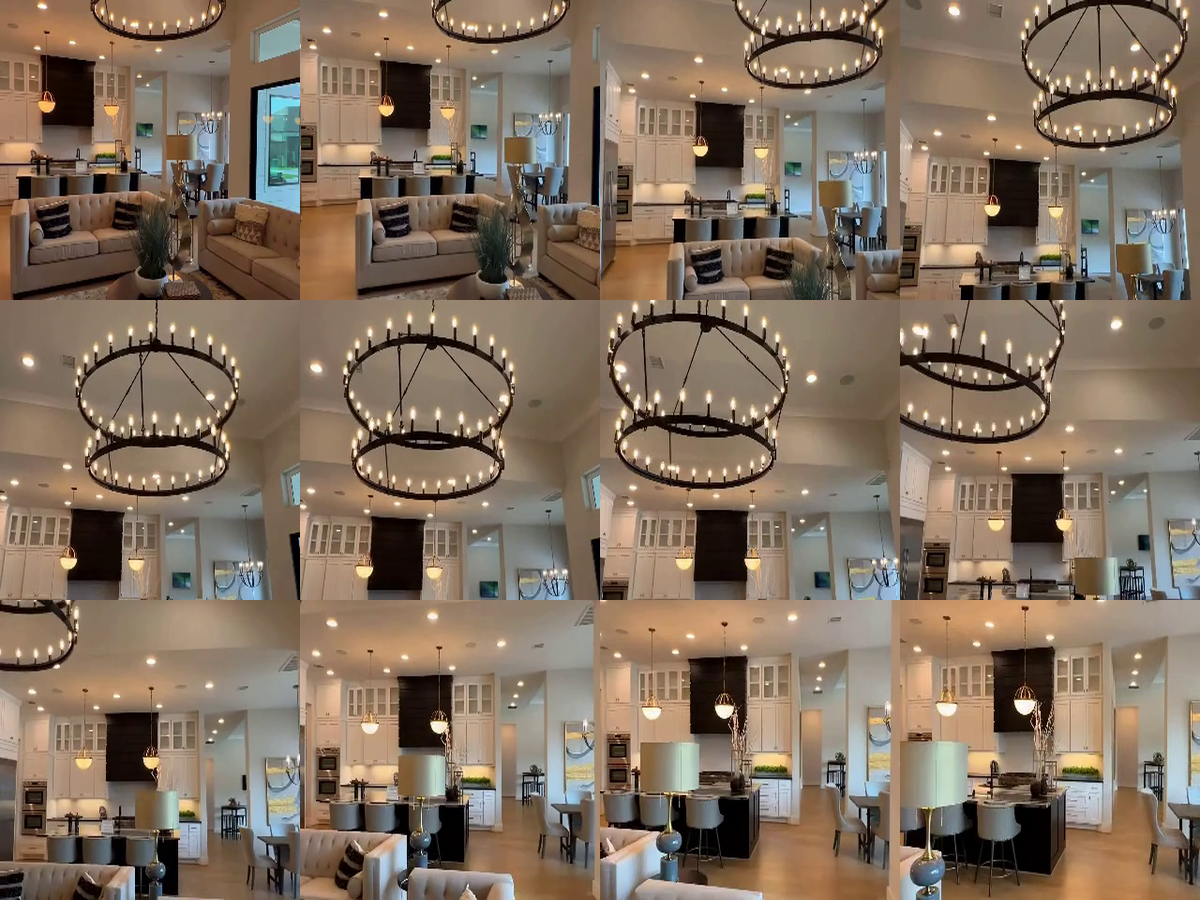} &
        \includegraphics[width=\linewidth]{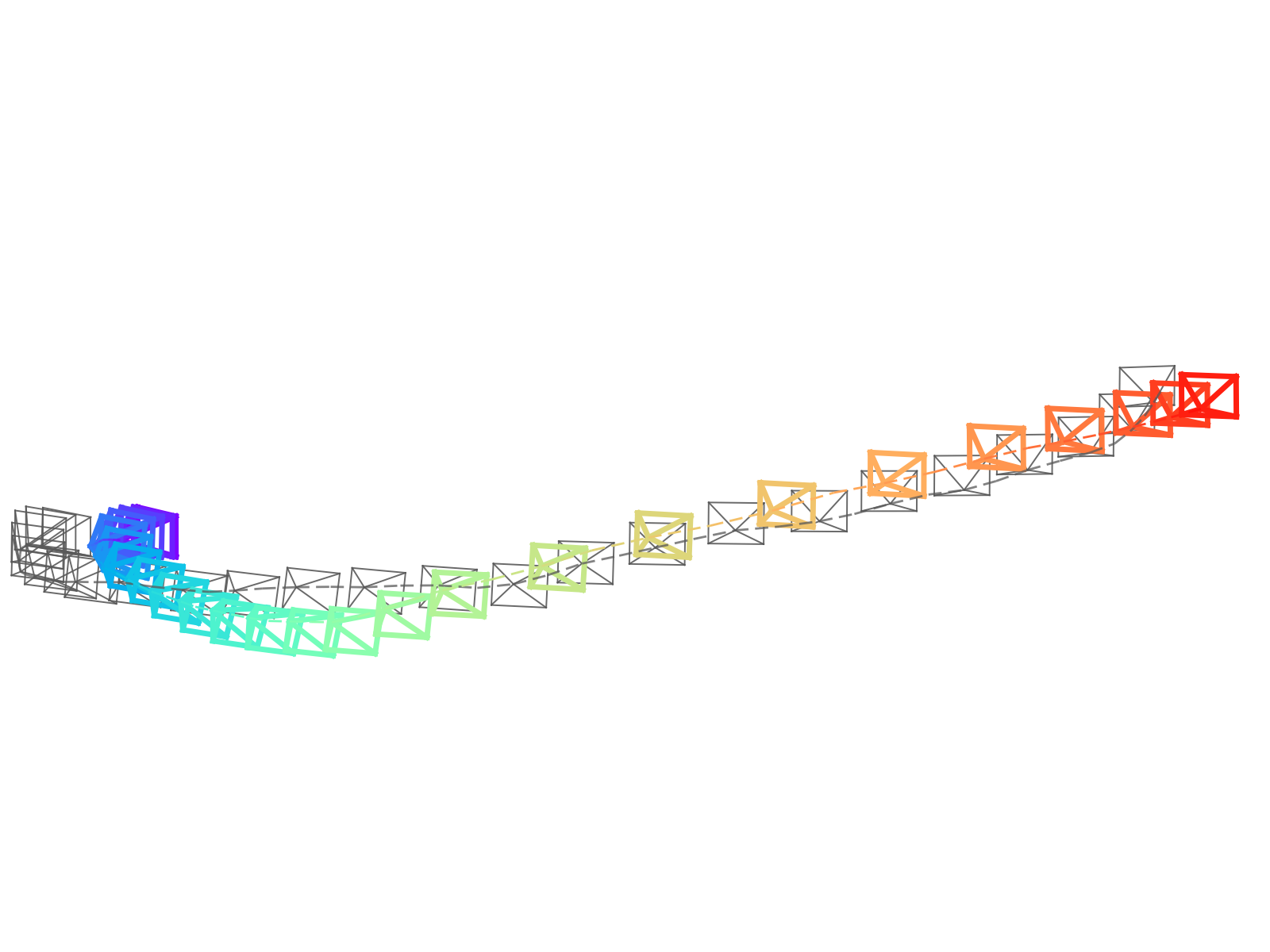} &
        \includegraphics[width=\linewidth]{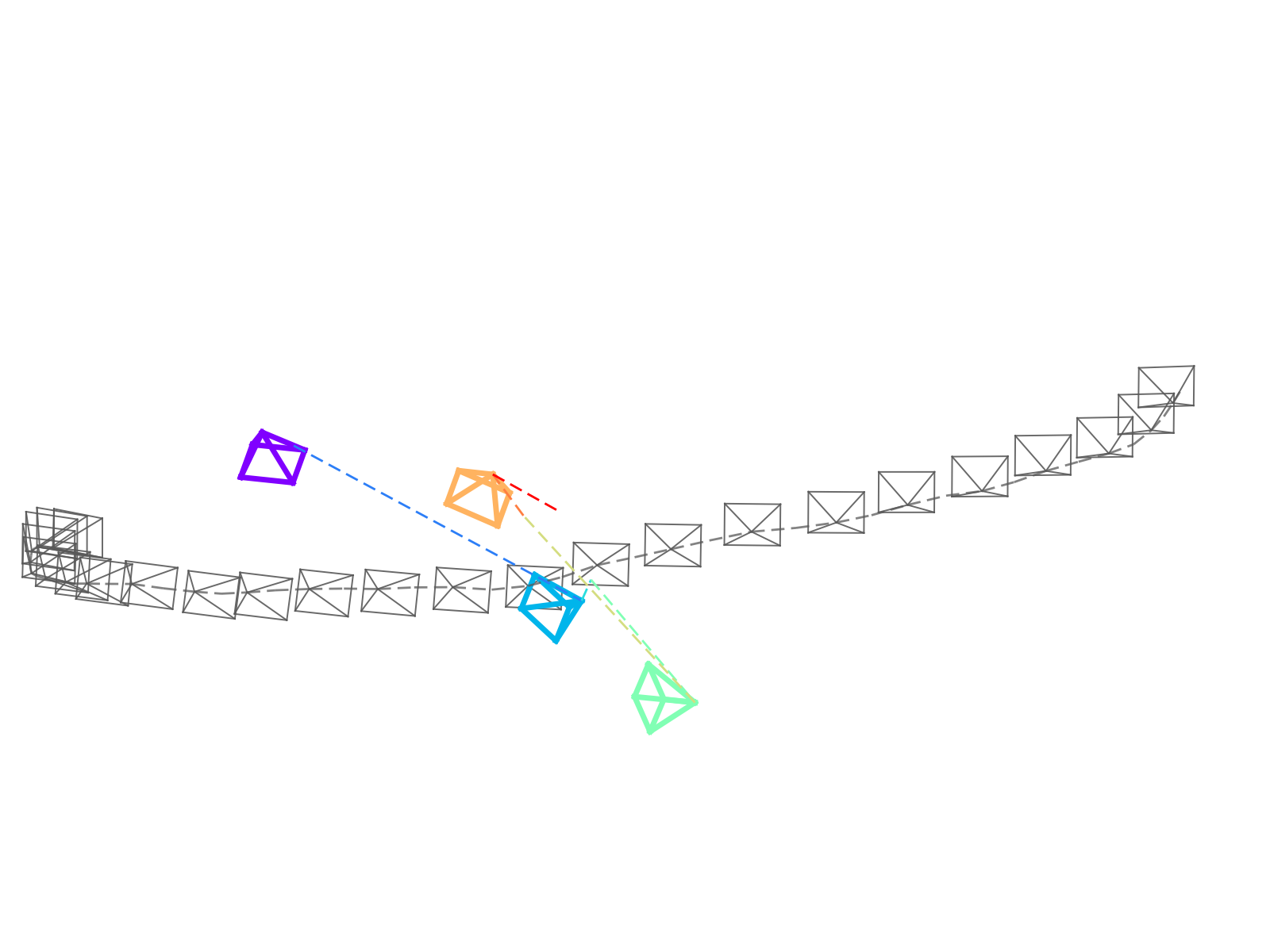} &
        \includegraphics[width=\linewidth]{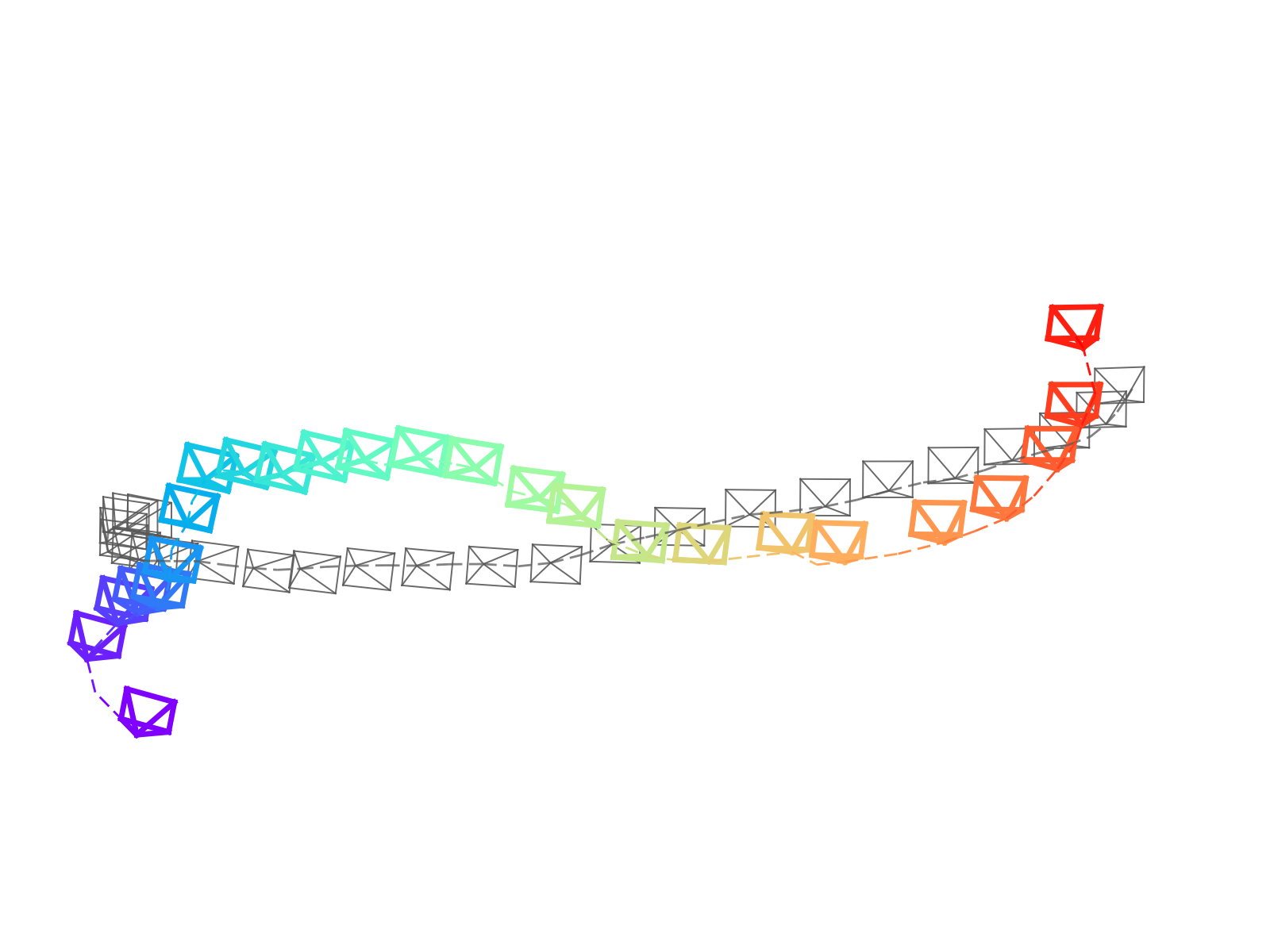}&
        \includegraphics[width=\linewidth]{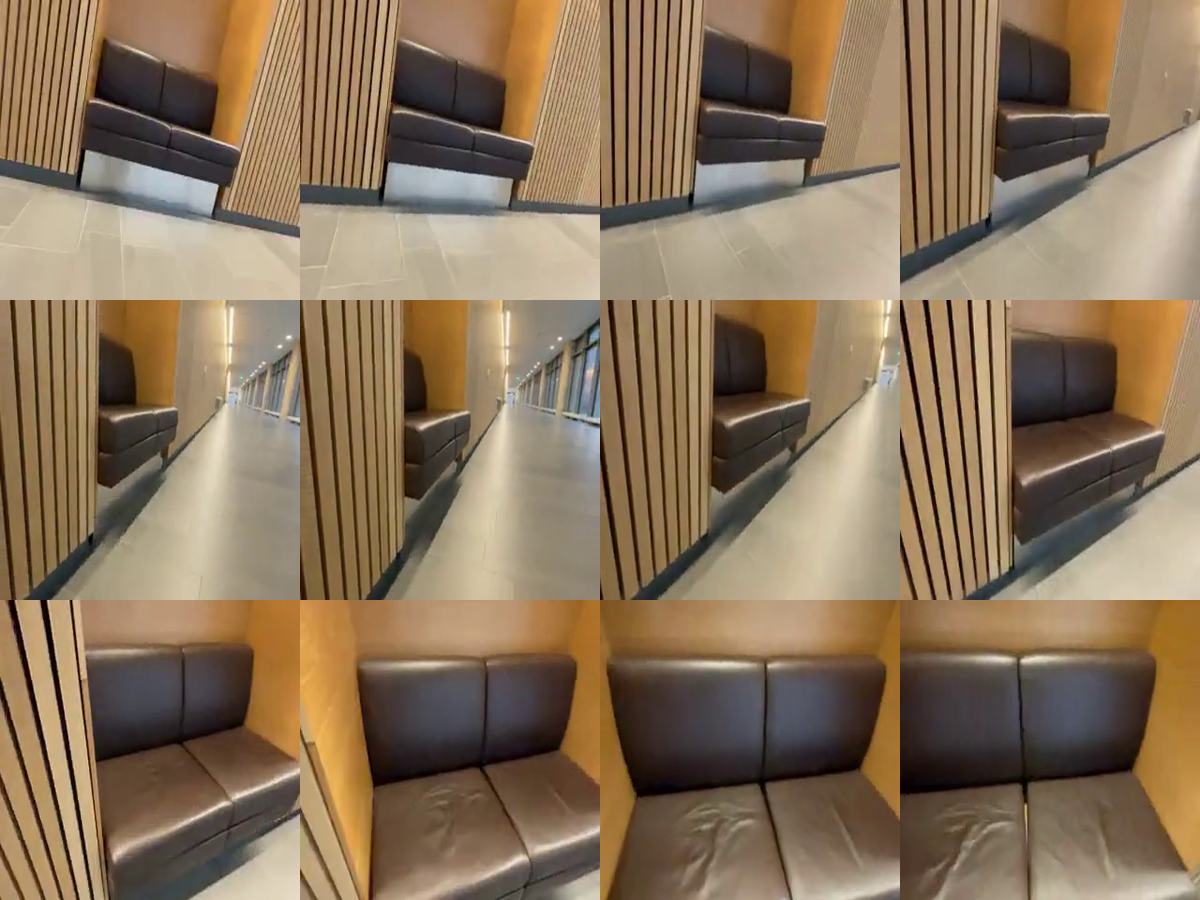} &
        \includegraphics[width=\linewidth]{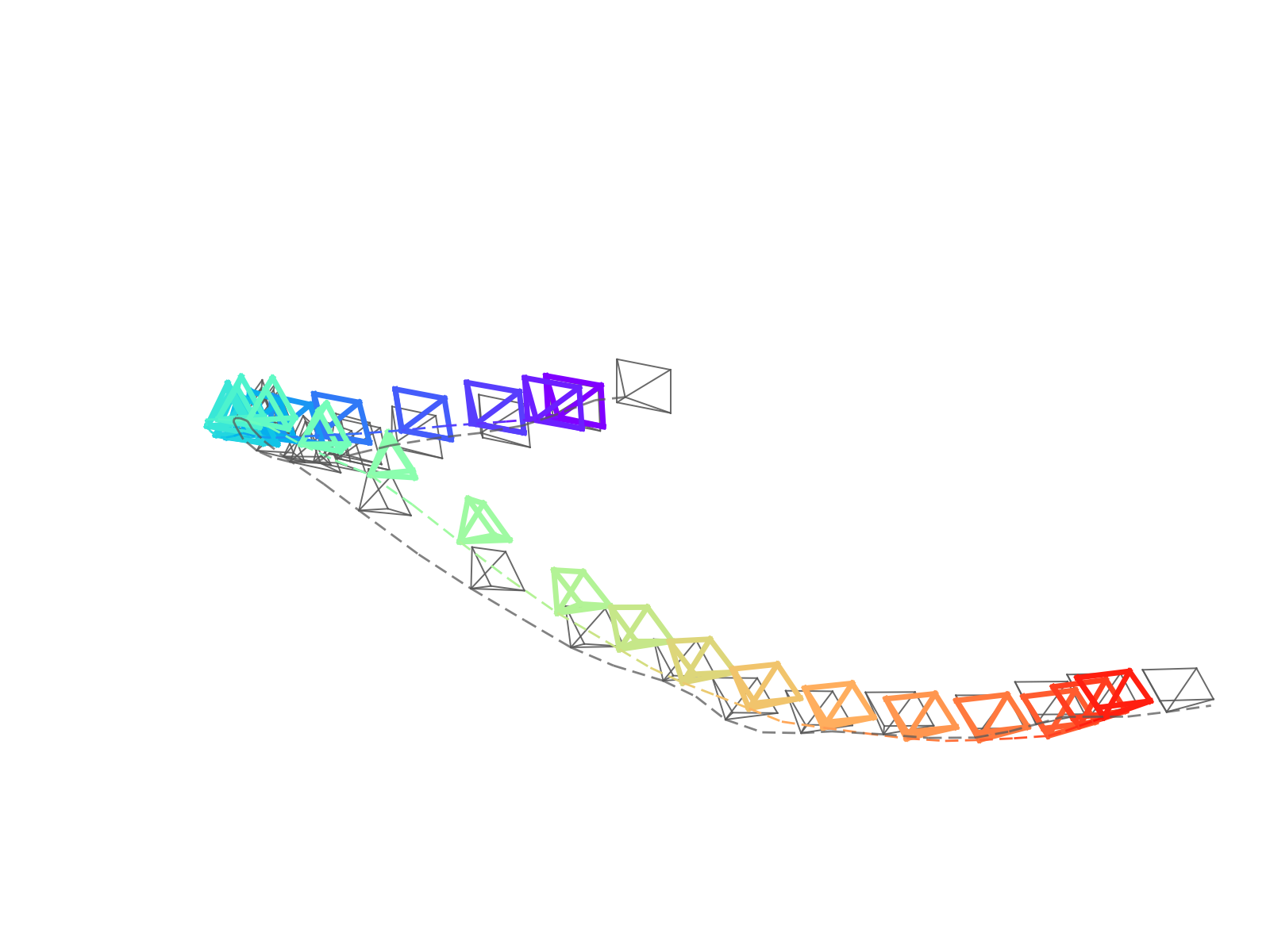} &
        \includegraphics[width=\linewidth]{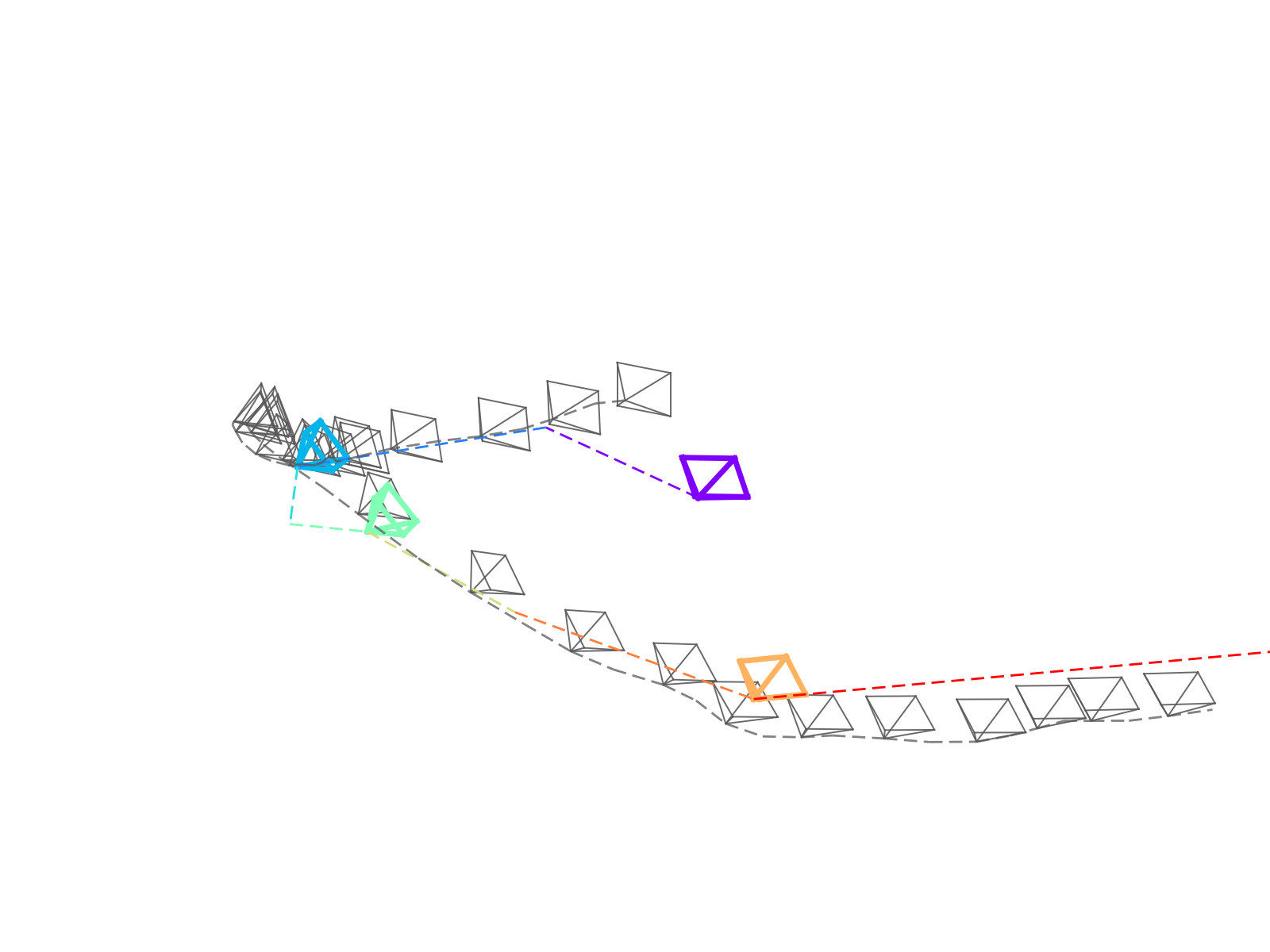} &
        \includegraphics[width=\linewidth]{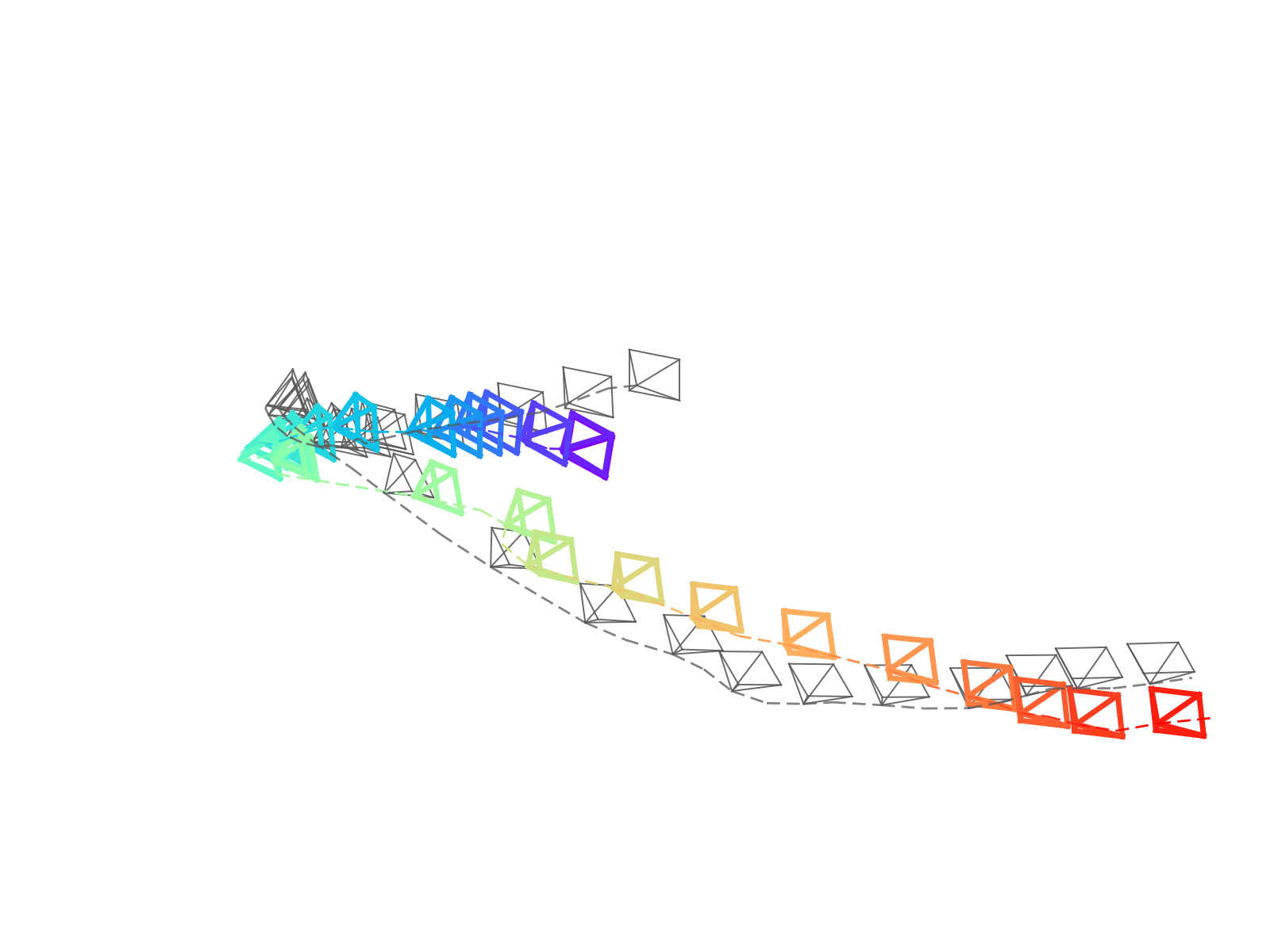} \\[-1pt] %
        
        \includegraphics[width=\linewidth]{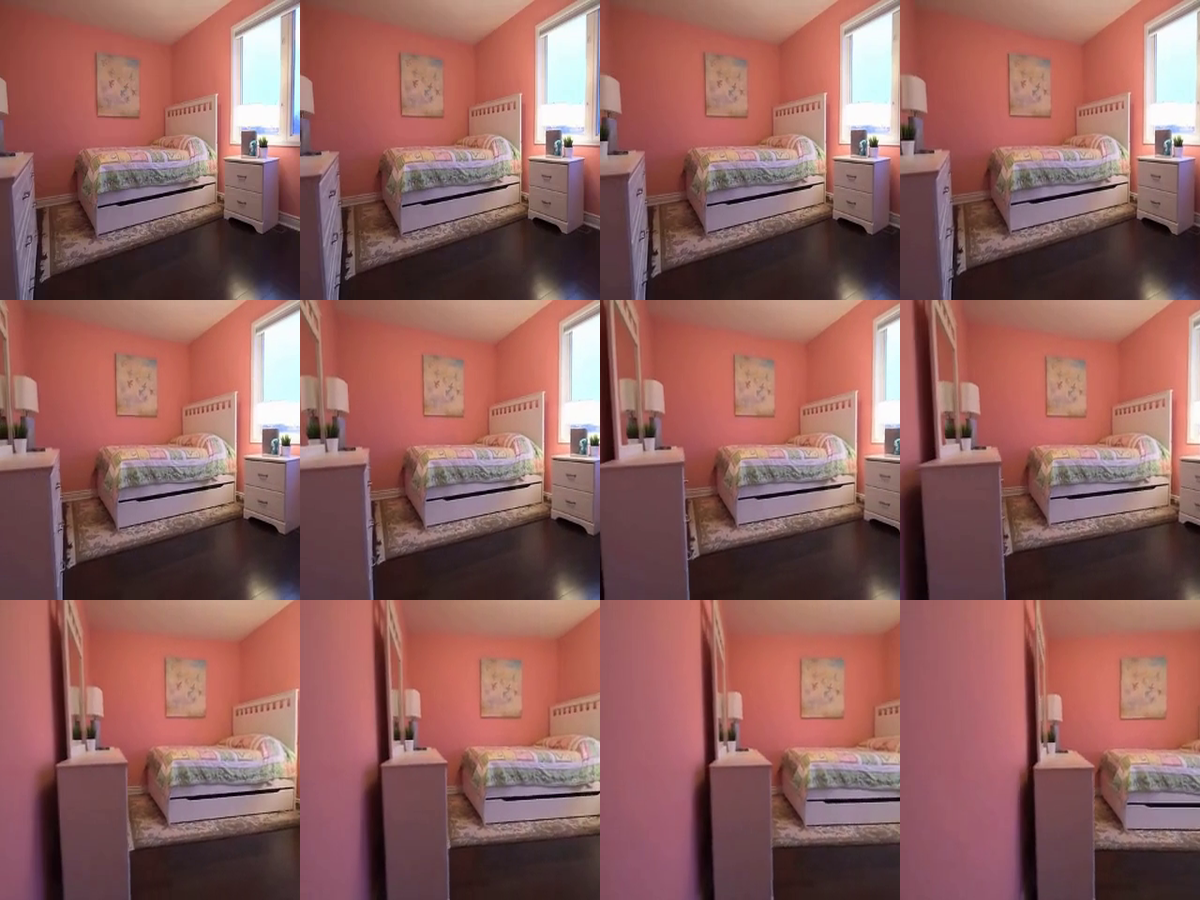} &
        \includegraphics[width=\linewidth]{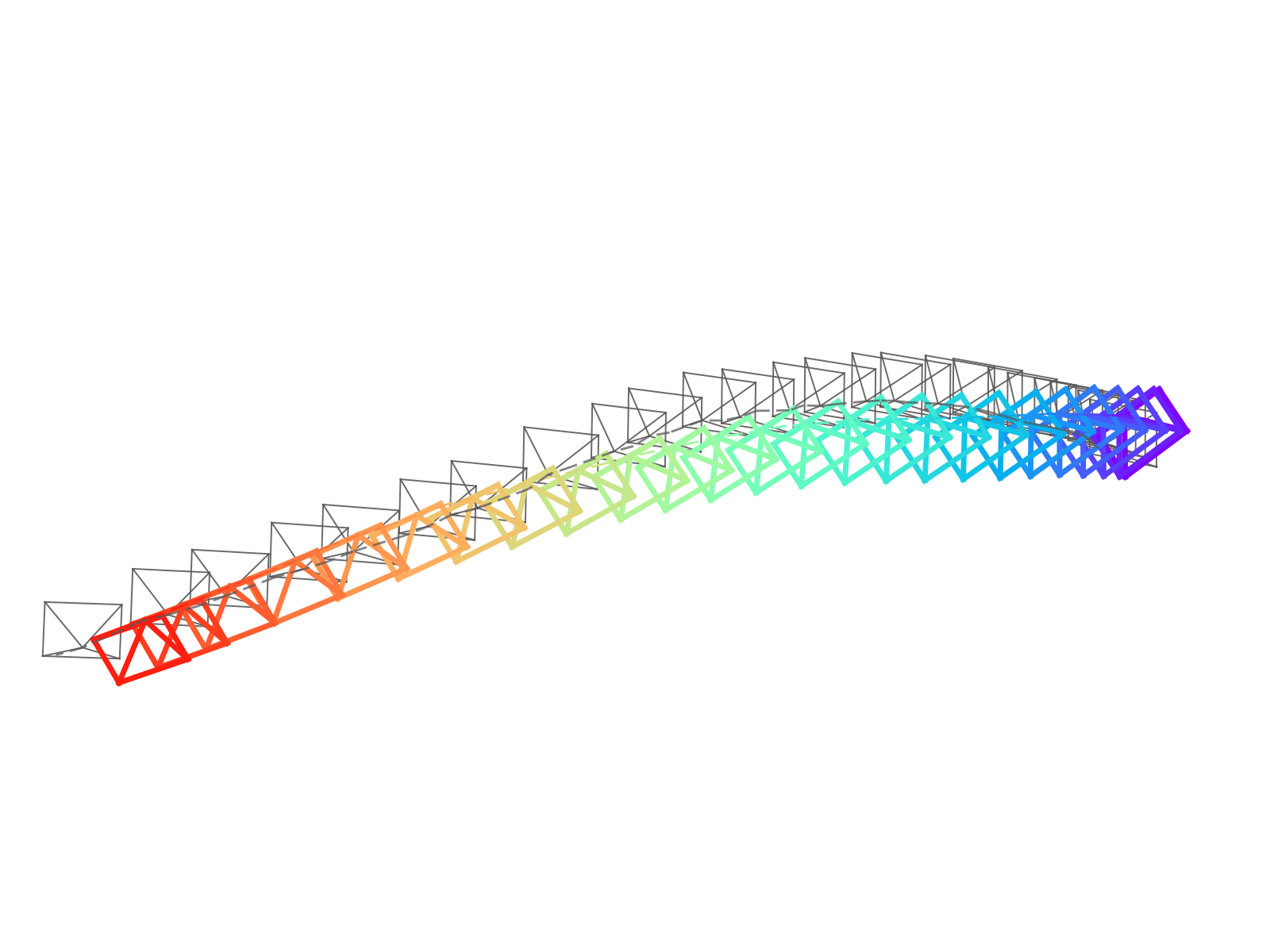} &
        \includegraphics[width=\linewidth]{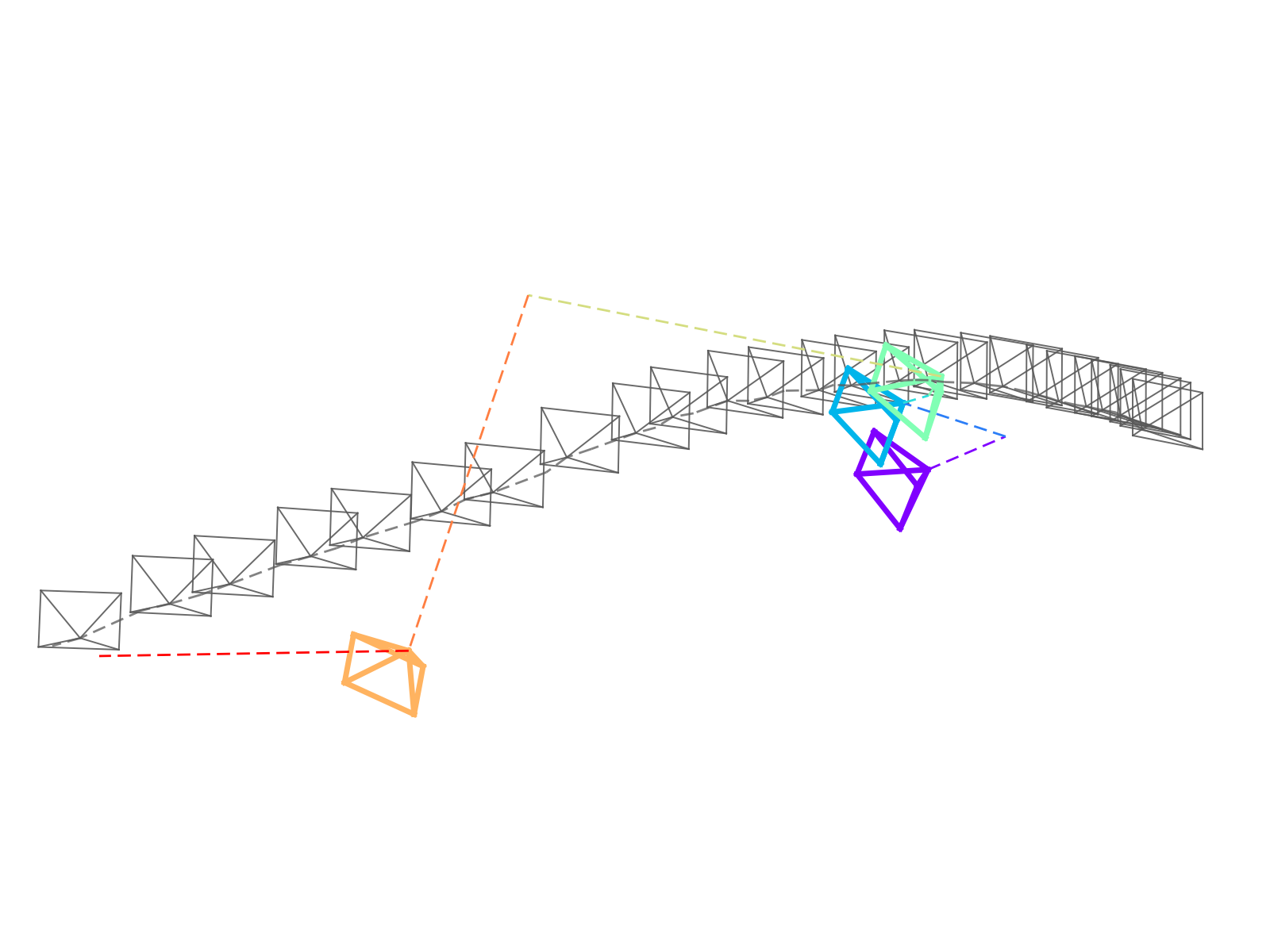} &
        \includegraphics[width=\linewidth]{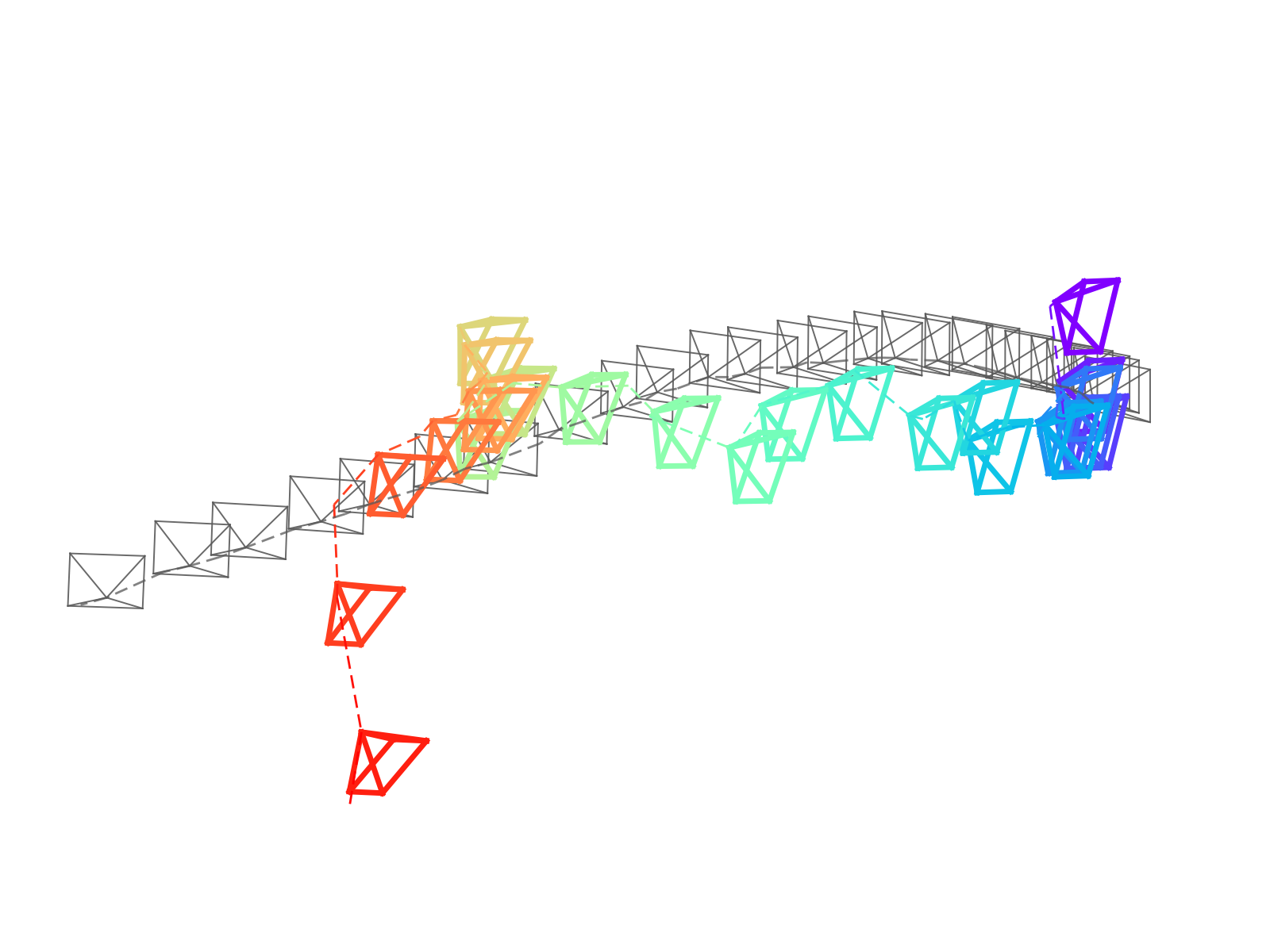}&
         \includegraphics[width=\linewidth]{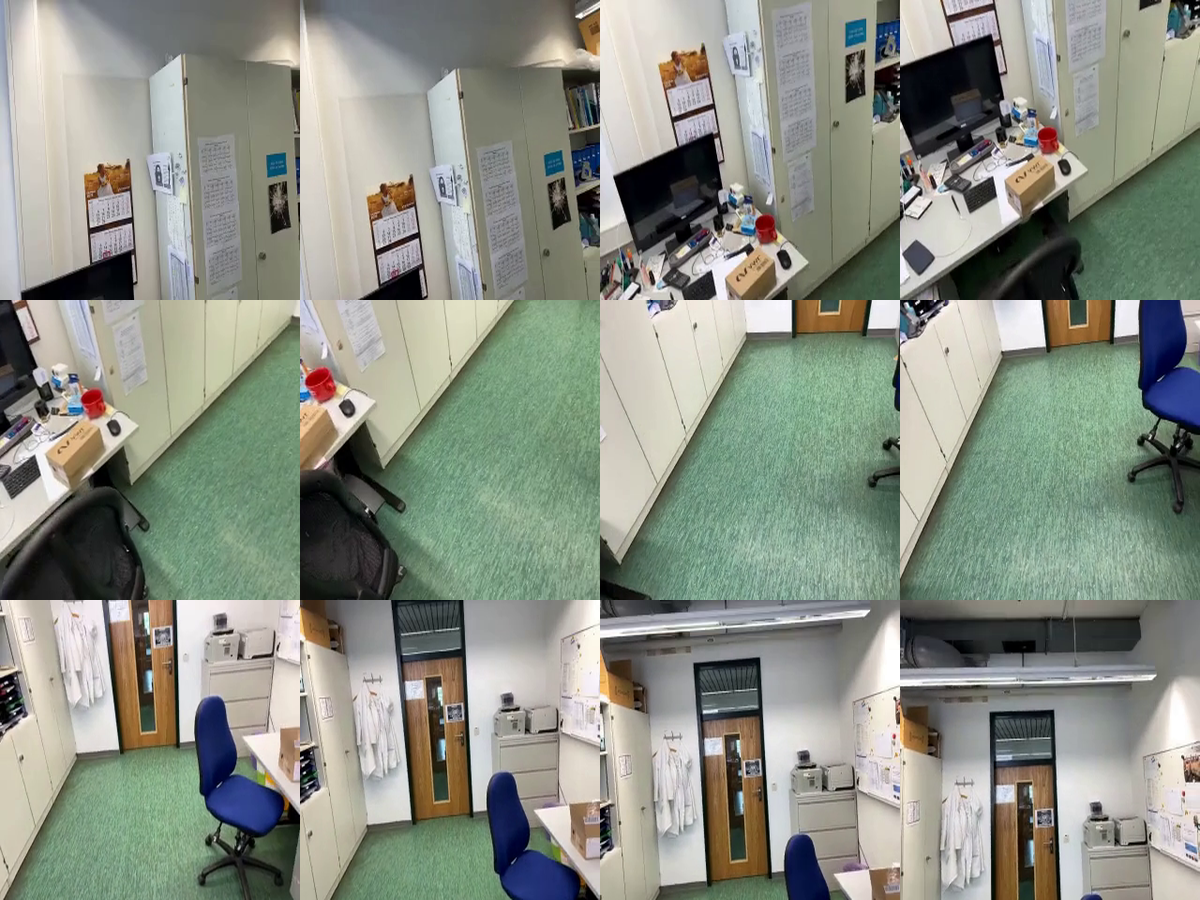} &
        \includegraphics[width=\linewidth]{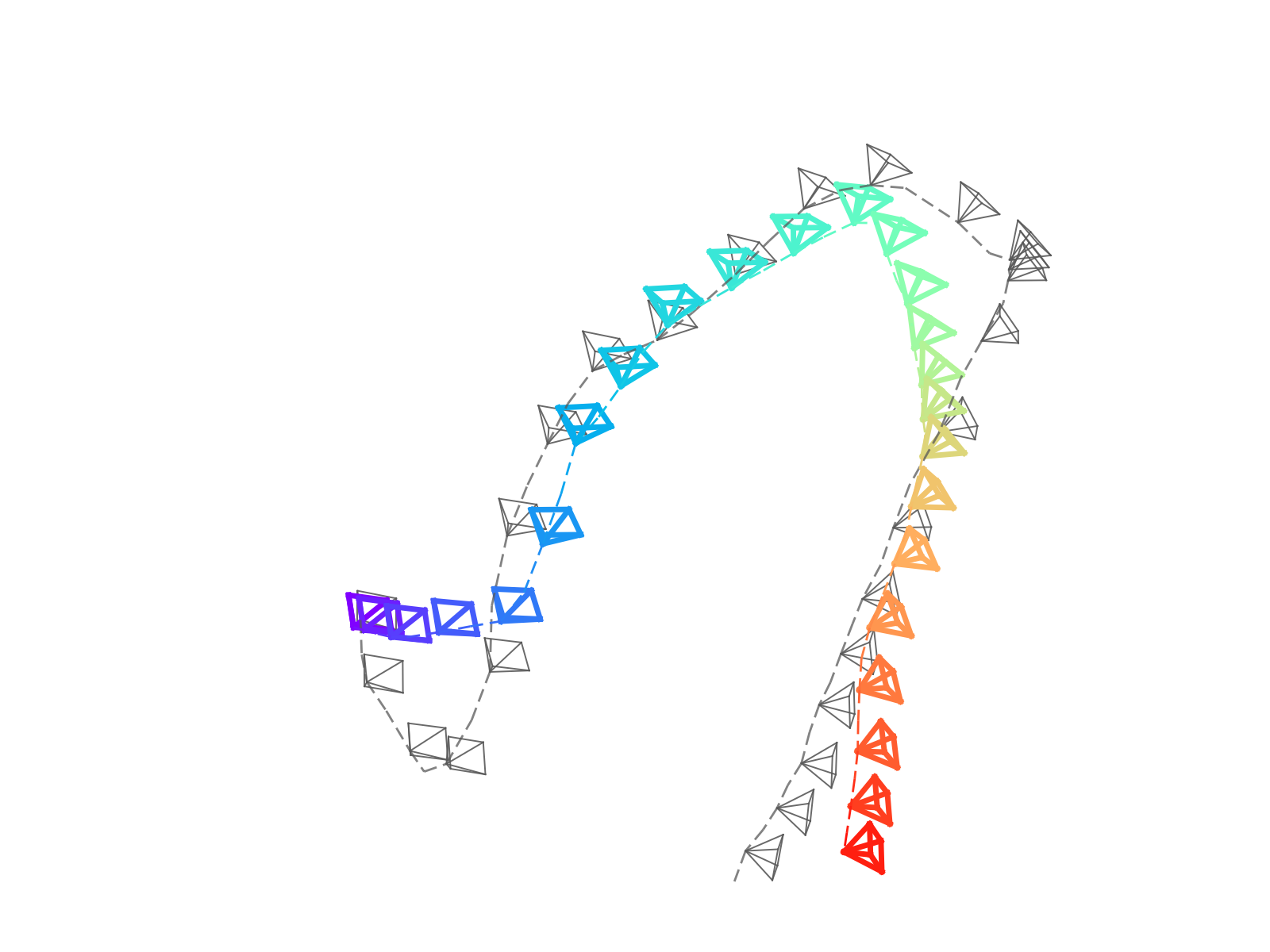} &
        \includegraphics[width=\linewidth]{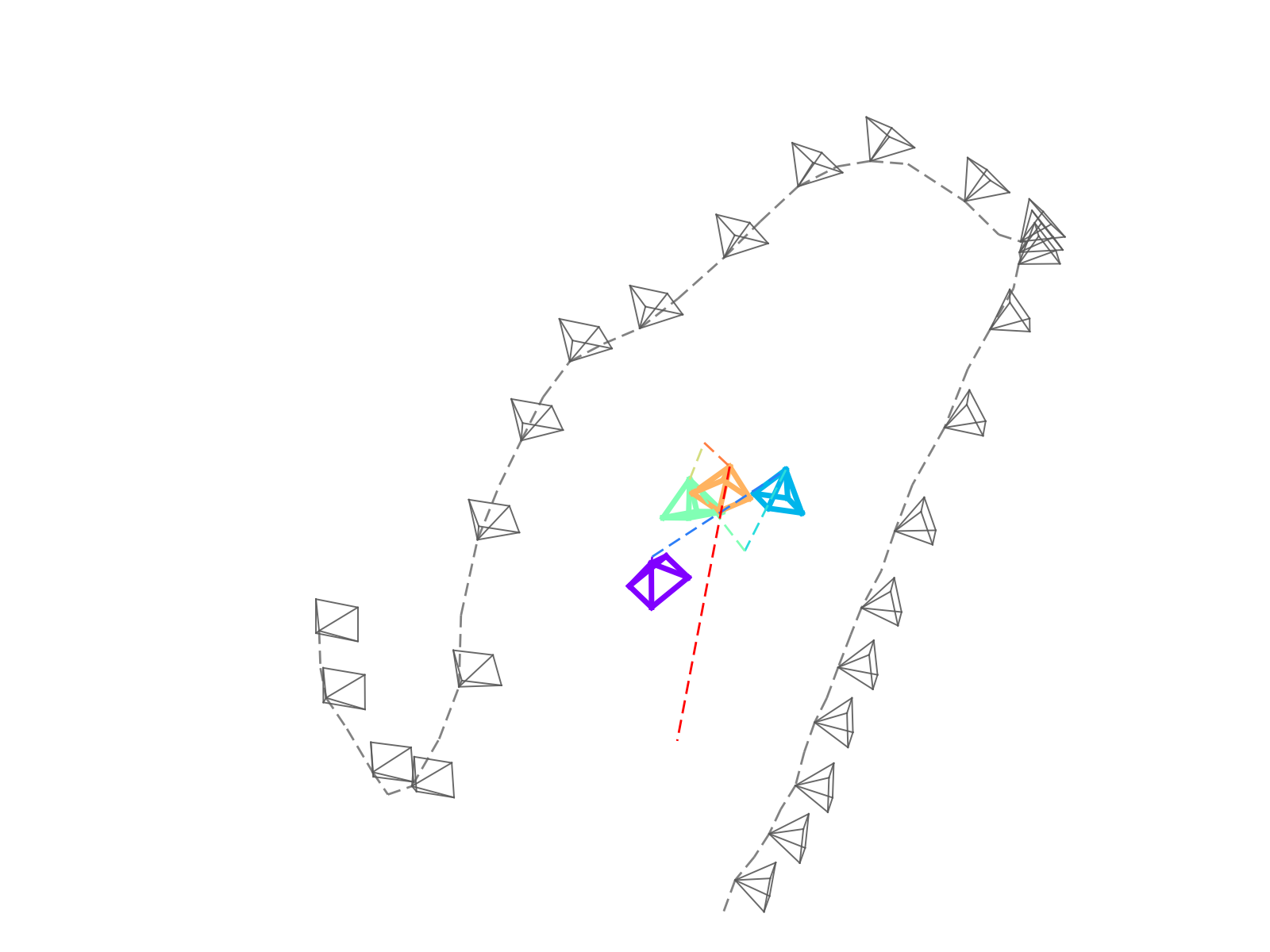} &
        \includegraphics[width=\linewidth]{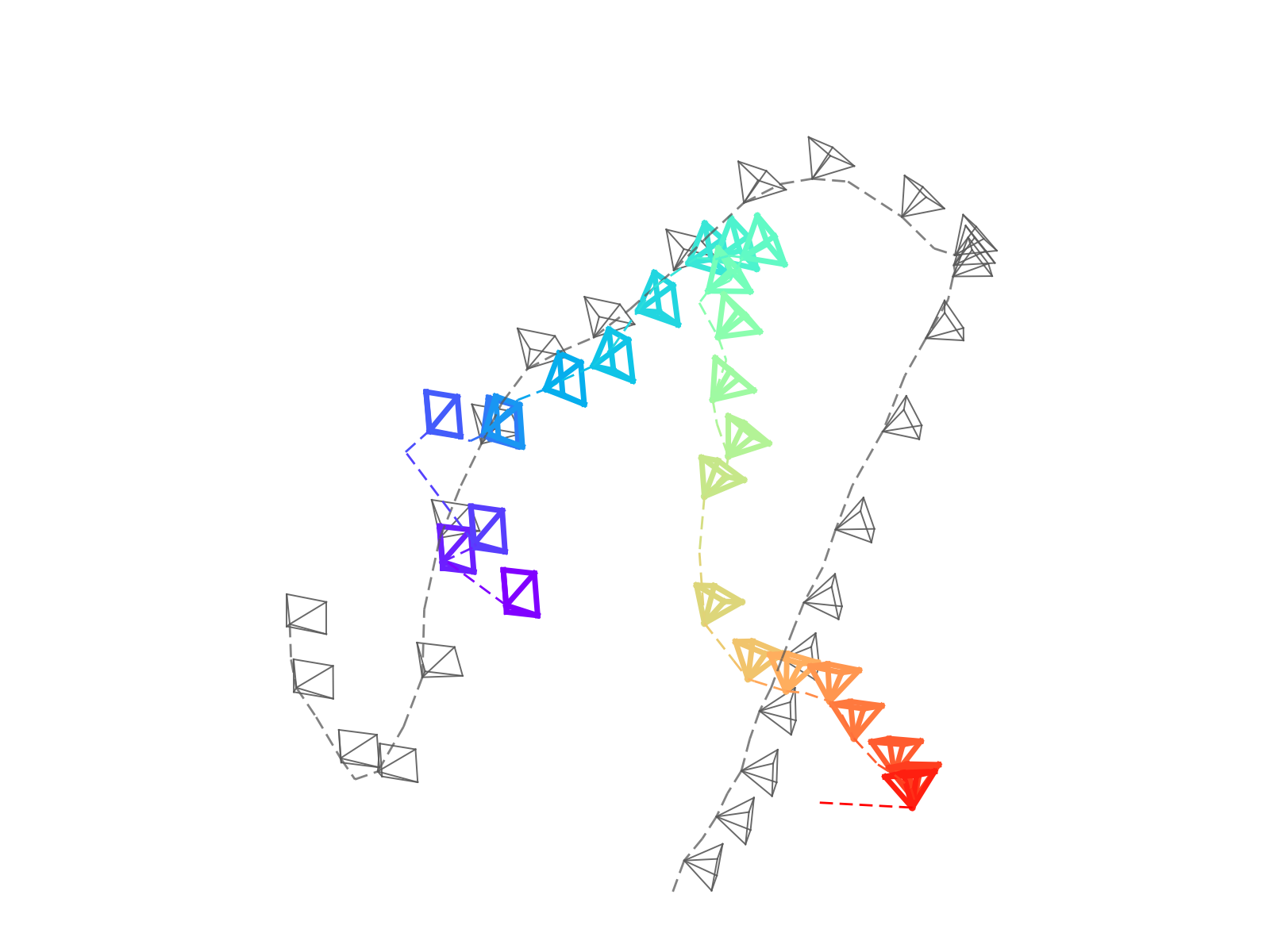}\\[-3pt]
    \end{tabularx}
    
    \caption{\textbf{Qualitative comparison of estimated camera trajectories.} 
The temporal evolution of each trajectory is encoded by a color gradient, from \textcolor{myPurple}{purple} (start) to \textcolor{red}{red} (end). Ground-truth trajectory is shown in black.
The leftmost column shows representative RGB frames from the input video. 
Compared with Geo4D and RayDiffusion, our predicted trajectories more closely follow the ground-truth trajectory, with better overall structure in challenging cases.}
    \label{fig:pose_comparison}
    \vspace{-17pt}
\end{figure*}

\providecommand{\best}[1]{\cellcolor{blue!12}#1}
\providecommand{\secondbest}[1]{\cellcolor{blue!5}#1}

\begin{table}[t]
\centering
\caption{\textbf{Camera pose estimation results.}
Dense: ATE, RPE$_\mathrm{t}$, and RPE$_\mathrm{r}$.
Sparse: tAE and RPE$_\mathrm{r}$,
evaluated only on the two observed frames.
Angular errors are in degrees; lower is better for all metrics.
Darker and lighter blue shading mark the best and
second-best distinct results in each column, including oracle results.}

\label{tab:final_pose_table}

\begingroup

\fontsize{8}{9}\selectfont
\setlength{\tabcolsep}{0.75pt}
\renewcommand{\arraystretch}{0.9}
\setlength{\aboverulesep}{0.2ex}
\setlength{\belowrulesep}{0.3ex}

\begin{tabular*}{\linewidth}{@{\extracolsep{\fill}}l@{\hspace{5pt}}*{3}{r}@{\hspace{4pt}}*{3}{r}@{\hspace{4pt}}*{3}{r}@{\hspace{4pt}}*{3}{r}@{\hspace{4pt}}*{3}{r}@{}}
\multicolumn{16}{@{}l}{\textbf{(a) Dense observations}} \\[2pt]
\toprule
Method & \multicolumn{3}{c}{RealEstate10K} & \multicolumn{3}{c}{Sintel} & \multicolumn{3}{c}{TUM} & \multicolumn{3}{c}{ScanNetV2} & \multicolumn{3}{c}{ScanNet++} \\
\cmidrule(lr){2-4} \cmidrule(lr){5-7} \cmidrule(lr){8-10} \cmidrule(lr){11-13} \cmidrule(lr){14-16}
 & ATE & $\mathrm{RPE}_{\mathrm{t}}$ & $\mathrm{RPE}_{\mathrm{r}}$ & ATE & $\mathrm{RPE}_{\mathrm{t}}$ & $\mathrm{RPE}_{\mathrm{r}}$ & ATE & $\mathrm{RPE}_{\mathrm{t}}$ & $\mathrm{RPE}_{\mathrm{r}}$ & ATE & $\mathrm{RPE}_{\mathrm{t}}$ & $\mathrm{RPE}_{\mathrm{r}}$ & ATE & $\mathrm{RPE}_{\mathrm{t}}$ & $\mathrm{RPE}_{\mathrm{r}}$ \\
\midrule
Geo4D
 & 0.026 & 0.028 & 0.147
 & 0.160 & 0.097 & 0.911
 & 0.052 & 0.029 & 1.414
 & 0.074 & 0.044 & 1.762
 & 0.190 & 0.090 & 5.591 \\
RayDiffusion
 & 0.139 & 0.081 & 1.809
 & 0.471 & 0.293 & 14.495
 & 0.088 & 0.058 & 2.942
 & 0.170 & 0.090 & 8.909
 & 0.344 & 0.148 & 25.941 \\
Gen3R
 & 0.030 & \secondbest{0.011} & \secondbest{0.141}
 & 0.233 & 0.066 & 1.538
 & 0.046 & 0.016 & \best{0.943}
 & 0.075 & 0.024 & 1.018
 & 0.232 & 0.066 & 5.760 \\
Aether
 & 0.018 & 0.014 & 0.851
 & 0.180 & 0.867 & 1.354
 & 0.041 & 0.016 & 1.462
 & 0.075 & 0.030 & 1.617
 & 0.225 & 0.068 & 5.944 \\
\midrule
UGG
 & 0.010 & \best{0.004} & \best{0.076}
 & 0.098 & 0.034 & 0.708
 & 0.018 & \secondbest{0.007} & \secondbest{1.052}
 & \secondbest{0.038} & \secondbest{0.016} & \secondbest{0.938}
 & 0.087 & \secondbest{0.036} & \best{3.861} \\
UGG (TTS)
 & \secondbest{0.009} & \best{0.004} & \best{0.076}
 & \secondbest{0.091} & \secondbest{0.033} & \best{0.684}
 & \secondbest{0.017} & \secondbest{0.007} & 1.054
 & \secondbest{0.038} & \secondbest{0.016} & \best{0.937}
 & \secondbest{0.082} & \best{0.035} & 4.105 \\
UGG (TTS Oracle)
 & \best{0.008} & \best{0.004} & \best{0.076}
 & \best{0.072} & \best{0.030} & \secondbest{0.695}
 & \best{0.015} & \best{0.006} & 1.053
 & \best{0.033} & \best{0.015} & 0.939
 & \best{0.072} & \best{0.035} & \secondbest{4.078} \\
\bottomrule
\end{tabular*}
\par\vspace{2pt}

\begin{tabular*}{\linewidth}{@{\extracolsep{\fill}}l*{10}{r}@{}}
\multicolumn{11}{@{}l}{\textbf{(b) Sparse observations}} \\[2pt]
\toprule
Method & \multicolumn{2}{c}{RealEstate10K} & \multicolumn{2}{c}{Sintel} & \multicolumn{2}{c}{TUM} & \multicolumn{2}{c}{ScanNetV2} & \multicolumn{2}{c}{ScanNet++} \\
\cmidrule(lr){2-3} \cmidrule(lr){4-5} \cmidrule(lr){6-7} \cmidrule(lr){8-9} \cmidrule(lr){10-11}
 & tAE & $\mathrm{RPE}_{\mathrm{r}}$ & tAE & $\mathrm{RPE}_{\mathrm{r}}$ & tAE & $\mathrm{RPE}_{\mathrm{r}}$ & tAE & $\mathrm{RPE}_{\mathrm{r}}$ & tAE & $\mathrm{RPE}_{\mathrm{r}}$ \\
\midrule
Geo4D
 & 117.079 & 3.407
 & 108.628 & 24.160
 & 100.224 & 11.936
 & 90.812 & 43.458
 & 82.747 & 83.019 \\
 RayDiffusion
 & 75.135 & 11.206
 & 60.805 & 45.403
 & 57.922 & 5.155
 & 67.727 & 41.695
 & 72.027 & 66.447 \\
Gen3R
 & 44.004 & 4.197
 & 75.984 & 24.984
 & 82.875 & 12.374
 & 79.871 & 43.481
 & 92.324 & 82.540 \\
Aether
 & 39.834 & 14.969
 & 78.012 & 33.640
 & 48.436 & 6.234
 & 70.925 & 36.720
 & 77.234 & 81.681 \\
\midrule
UGG
 & 8.260 & \best{1.287}
 & 35.479 & \secondbest{22.065}
 & 9.938 & \secondbest{1.371}
 & 28.240 & 21.417
 & 40.194 & \secondbest{43.078} \\
UGG (TTS)
 & \secondbest{3.475} & \secondbest{1.393}
 & \secondbest{18.733} & \best{20.646}
 & \secondbest{8.520} & 1.444
 & \secondbest{20.057} & \best{18.911}
 & \secondbest{31.835} & \best{37.875} \\
UGG (TTS Oracle)
 & \best{1.550} & 1.399
 & \best{8.521} & 26.135
 & \best{3.401} & \best{1.154}
 & \best{9.683} & \secondbest{20.688}
 & \best{7.892} & 43.836 \\
\bottomrule
\end{tabular*}
\endgroup
\vspace{-10pt}
\end{table}

\subsection{Camera Pose Estimation}
\label{sec:pose_estimation}
We evaluate camera pose estimation from RGB observations in dense and sparse settings, conditioning on either the full video or only the first and last frames. 
Camera-ray tokens are denoised and
converted back into camera parameters, testing whether camera geometry can be predicted as a
missing modality within the same conditional-completion framework.

\paragraph{Benchmarks and Metrics.}
We report results on RealEstate10K, Sintel~\cite{Butler:ECCV:2012},
TUM~\cite{sturm2012tumdynamics}, ScanNetV2~\cite{dai2017scannet},
and ScanNet++. Following Geo4D~\cite{jiang2025geo4d}, we align
predicted trajectories to ground truth using Sim(3) Umeyama
alignment in the dense setting, and report absolute trajectory
error (ATE), relative pose error in translation (RPE$_\mathrm{t}$),
and in rotation (RPE$_\mathrm{r}$). In the sparse setting, we
evaluate the relative pose between the first and last frames
using translation angular error (tAE) and RPE$_\mathrm{r}$.
Detailed metric definitions and evaluation protocols are provided
in Appendix~\ref{app:pose_evaluation}.

\paragraph{Baselines.}
We compare quantitatively against Geo4D~\cite{jiang2025geo4d}, Gen3R~\cite{Huang_2026_CVPR}, Aether~\cite{zhu2025iccv-aether}, and RayDiffusion~\cite{zhang2024raydiffusion}. 
Geo4D estimates 4D geometry using a generative video prior, while RayDiffusion directly models camera rays through a diffusion denoising process. 
The geometry-focused baselines are more specialized for pose estimation, whereas UGG predicts camera rays as part of a broader RGB-depth-camera scene representation.

\textbf{Evaluation results.}
Table~\ref{tab:final_pose_table} reports camera pose estimation
results. In the dense setting, UGG achieves lower ATE and
RPE$_\mathrm{t}$ than the external baselines across all five
datasets. For example, ATE is reduced from 0.190 to 0.087 on
ScanNet++ compared with Geo4D. Figure~\ref{fig:pose_comparison}
visualizes representative trajectories. Our predictions better
follow the ground-truth motion, while RayDiffusion and Geo4D
produce collapsed paths in challenging examples.
In the sparse setting, UGG obtains the lowest tAE and
RPE$_\mathrm{r}$ across all five datasets. These results suggest
that the model can infer reasonable relative camera poses from
only the first and last RGB frames, despite the ambiguity of
sparse observations.

\begin{figure*}[tb]
    \centering
    \small
    \setlength{\tabcolsep}{1pt} %
    \begin{tabularx}{\textwidth}{MMMMM @{\hspace{72pt}} Z @{\hspace{1pt}} Z @{\hspace{1pt}} Z @{\hspace{1pt}} Z @{}}
    
    \textbf{Input RGB} & \textbf{GT} & \textbf{\textcolor{myPurple} {Ours}} & \textbf{S1:Chrono} & \textbf{S2:Geo4D} & 
    \makebox[0pt][c]{\hspace{-42pt} \textbf{Zoom-in (GT / Ours / S1 / S2)}} \\

        \includegraphics[width=\linewidth]{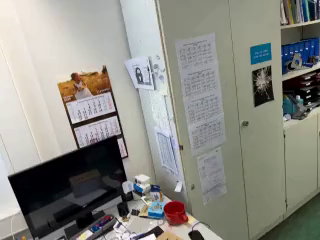} &

        \begin{tikzpicture}[nvsspy]
            \node[anchor=south west, inner sep=0] (imGT) at (0,0) {\includegraphics[width=\linewidth]{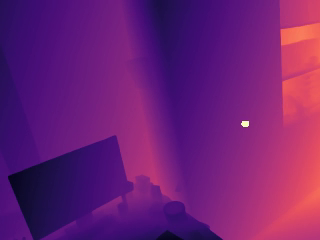}};
            
            \spy [every spy on node/.append style={draw=white,line width=0.6pt}, 
                  every spy in node/.append style={draw=none,line width=0.6pt,fill=white}] 
                  on ($(imGT.south west)!0.35!(imGT.south east)!0.24!(imGT.north east)$) 
                  in node at (7.7, 0.7); 
        \end{tikzpicture} &
        
        \begin{tikzpicture}[nvsspy]
            \node[anchor=south west, inner sep=0] (imOurs) at (0,0) {\includegraphics[width=\linewidth]{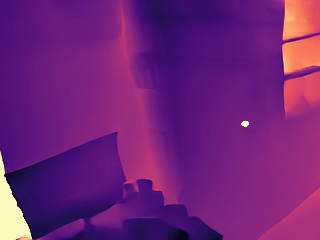}};
            
            \spy [every spy on node/.append style={draw=white,line width=0.6pt}, 
                  every spy in node/.append style={draw=none,line width=0.6pt,fill=white}] 
                  on ($(imOurs.south west)!0.35!(imOurs.south east)!0.24!(imOurs.north east)$) 
                  in node at (7, 0.7); 
        \end{tikzpicture} &

        \begin{tikzpicture}[nvsspy]
            \node[anchor=south west, inner sep=0] (imS1) at (0,0) {\includegraphics[width=\linewidth]{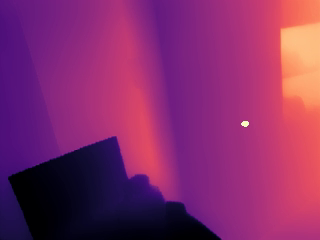}};
            
            \spy [every spy on node/.append style={draw=white,line width=0.6pt}, 
                  every spy in node/.append style={draw=none,line width=0.6pt,fill=white}] 
                  on ($(imS1.south west)!0.35!(imS1.south east)!0.24!(imS1.north east)$) 
                  in node at (6.3, 0.7); %
        \end{tikzpicture} &
        
        \begin{tikzpicture}[nvsspy]
            \node[anchor=south west, inner sep=0] (imS2) at (0,0) {\includegraphics[width=\linewidth]{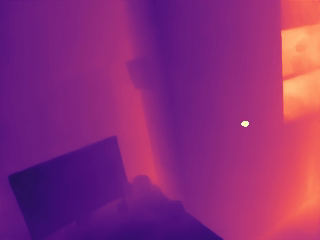}};
            
            \spy [every spy on node/.append style={draw=white,line width=0.6pt}, 
                  every spy in node/.append style={draw=none,line width=0.6pt,fill=white}] 
                  on ($(imS2.south west)!0.35!(imS2.south east)!0.24!(imS2.north east)$) 
                  in node at (5.6, 0.7); %
        \end{tikzpicture} & 
        & &\\[0pt] %

        \includegraphics[width=\linewidth]{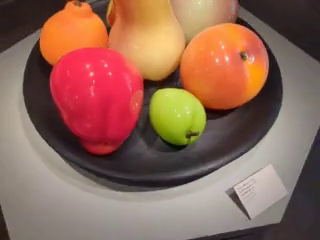} &
        \begin{tikzpicture}[nvsspy]
            \node[anchor=south west, inner sep=0] (imGT) at (0,0) {\includegraphics[width=\linewidth]{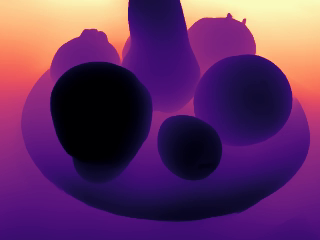}};
            
            \spy [every spy on node/.append style={draw=white,line width=0.6pt}, 
                  every spy in node/.append style={draw=none,line width=0.6pt,fill=white}] 
                  on ($(imGT.south west)!-0.9!(imGT.south east)!0.8!(imGT.north east)$) 
                  in node at (7.7, 0.7); 
        \end{tikzpicture} &
        
        \begin{tikzpicture}[nvsspy]
            \node[anchor=south west, inner sep=0] (imOurs) at (0,0) {\includegraphics[width=\linewidth]{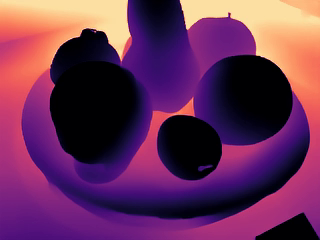}};
            
            \spy [every spy on node/.append style={draw=white,line width=0.6pt}, 
                  every spy in node/.append style={draw=none,line width=0.6pt,fill=white}] 
                  on ($(imOurs.south west)!-0.9!(imOurs.south east)!0.8!(imOurs.north east)$) 
                  in node at (7, 0.7); 
        \end{tikzpicture} &

        \begin{tikzpicture}[nvsspy]
            \node[anchor=south west, inner sep=0] (imS1) at (0,0) {\includegraphics[width=\linewidth]{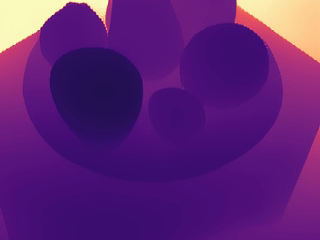}};
            
            \spy [every spy on node/.append style={draw=white,line width=0.6pt}, 
                  every spy in node/.append style={draw=none,line width=0.6pt,fill=white}] 
                  on ($(imS1.south west)!-0.9!(imS1.south east)!0.8!(imS1.north east)$) 
                  in node at (6.3, 0.7); %
        \end{tikzpicture} &
        
        \begin{tikzpicture}[nvsspy]
            \node[anchor=south west, inner sep=0] (imS2) at (0,0) {\includegraphics[width=\linewidth]{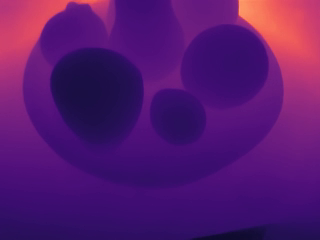}};
            
            \spy [every spy on node/.append style={draw=white,line width=0.6pt}, 
                  every spy in node/.append style={draw=none,line width=0.6pt,fill=white}] 
                  on ($(imS2.south west)!-0.9!(imS2.south east)!0.8!(imS2.north east)$) 
                  in node at (5.6, 0.7); %
        \end{tikzpicture} & 
        & \\[0pt] %

        \includegraphics[width=\linewidth]{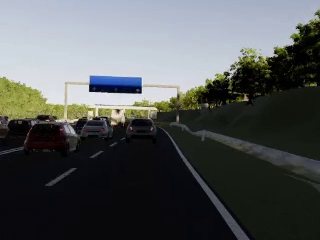} &
        \begin{tikzpicture}[nvsspy]
            \node[anchor=south west, inner sep=0] (imGT) at (0,0) {\includegraphics[width=\linewidth]{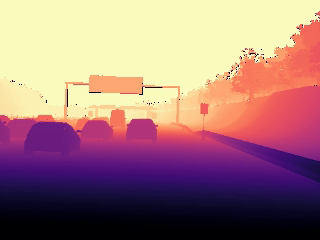}};
            
            \spy [every spy on node/.append style={draw=white,line width=0.6pt}, 
                  every spy in node/.append style={draw=none,line width=0.6pt,fill=white}] 
                  on ($(imGT.south west)!-0.2!(imGT.south east)!0.5!(imGT.north east)$) 
                  in node at (7.7, 0.7); 
        \end{tikzpicture} &
        
        \begin{tikzpicture}[nvsspy]
            \node[anchor=south west, inner sep=0] (imOurs) at (0,0) {\includegraphics[width=\linewidth]{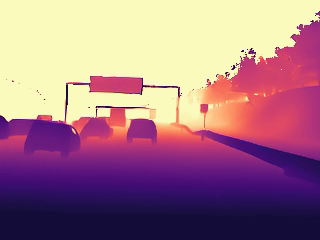}};
            
            \spy [every spy on node/.append style={draw=white,line width=0.6pt}, 
                  every spy in node/.append style={draw=none,line width=0.6pt,fill=white}] 
                  on ($(imOurs.south west)!-0.2!(imOurs.south east)!0.5!(imOurs.north east)$) 
                  in node at (7, 0.7); 
        \end{tikzpicture} &

        \begin{tikzpicture}[nvsspy]
            \node[anchor=south west, inner sep=0] (imS1) at (0,0) {\includegraphics[width=\linewidth]{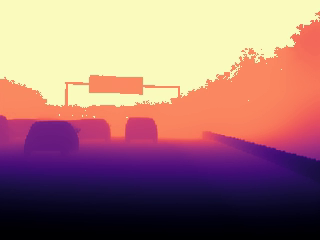}};
            
            \spy [every spy on node/.append style={draw=white,line width=0.6pt}, 
                  every spy in node/.append style={draw=none,line width=0.6pt,fill=white}] 
                  on ($(imS1.south west)!-0.2!(imS1.south east)!0.5!(imS1.north east)$) 
                  in node at (6.3, 0.7); %
        \end{tikzpicture} &
        
        \begin{tikzpicture}[nvsspy]
            \node[anchor=south west, inner sep=0] (imS2) at (0,0) {\includegraphics[width=\linewidth]{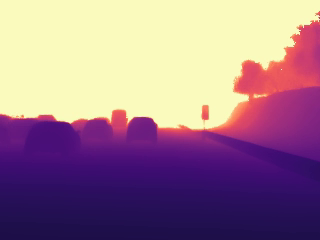}};
            
            \spy [every spy on node/.append style={draw=white,line width=0.6pt}, 
                  every spy in node/.append style={draw=none,line width=0.6pt,fill=white}] 
                  on ($(imS2.south west)!-0.2!(imS2.south east)!0.5!(imS2.north east)$) 
                  in node at (5.6, 0.7); %
        \end{tikzpicture} & 
        & \\[-8pt] %
    \end{tabularx}
    \caption{\textbf{Depth prediction qualitative comparison.} 
The white box marks the region shown in the zoom-in columns, ordered as GT, Ours, ChronoDepth, and Geo4D from left to right. 
Our predictions preserve sharper local structures and object boundaries in the highlighted regions.}
    \label{fig:depth_results}
    \vspace{-24pt}
\end{figure*}
\providecommand{\best}[1]{\cellcolor{blue!12}#1}
\providecommand{\secondbest}[1]{\cellcolor{blue!5}#1}

\begin{table}[t]
\centering
\caption{\textbf{Depth prediction results.}
We report Abs Rel~($\downarrow$) and $\delta < 1.25$~($\uparrow$)
on ScanNet++, 7-Scenes, KITTI, Sintel, and Bonn in dense and sparse
settings. Darker and lighter blue shading indicate the best and
second-best distinct results within each setting, including oracle results.}
\label{tab:final_depth_table}

\begingroup

\fontsize{8.5}{9.2}\selectfont
\setlength{\tabcolsep}{2pt}
\renewcommand{\arraystretch}{0.85}
\setlength{\aboverulesep}{0.15ex}
\setlength{\belowrulesep}{0.2ex}

\begin{tabular*}{\linewidth}{
  @{\extracolsep{\fill}} c l *{10}{r} @{}
}
\toprule
& \multirow{2}{*}{Method}
& \multicolumn{2}{c}{ScanNet++}
& \multicolumn{2}{c}{7-Scenes}
& \multicolumn{2}{c}{KITTI}
& \multicolumn{2}{c}{Sintel}
& \multicolumn{2}{c}{Bonn} \\
\cmidrule(lr){3-4}
\cmidrule(lr){5-6}
\cmidrule(lr){7-8}
\cmidrule(lr){9-10}
\cmidrule(lr){11-12}
& & Abs Rel & $\delta_{1.25}$
  & Abs Rel & $\delta_{1.25}$
  & Abs Rel & $\delta_{1.25}$
  & Abs Rel & $\delta_{1.25}$
  & Abs Rel & $\delta_{1.25}$ \\
\midrule

\multirow{8}{*}{\rotatebox[origin=c]{90}{Dense}}
& ChronoDepth
& 0.177 & 0.724
& 0.116 & 0.871
& 0.120 & 0.857
& 0.394 & 0.530
& 0.092 & 0.920 \\
& Geo4D
& 0.106 & 0.894
& 0.080 & \secondbest{0.923}
& 0.097 & 0.923
& \best{0.247} & \secondbest{0.676}
& \secondbest{0.057} & \secondbest{0.972} \\
& Gen3R
& 0.157 & 0.772
& \secondbest{0.079} & \best{0.937}
& 0.185 & 0.777
& 0.414 & 0.520
& 0.168 & 0.805 \\
& Aether
& 0.166 & 0.769
& 0.109 & 0.898
& \secondbest{0.056} & \best{0.975}
& 0.351 & 0.607
& 0.095 & 0.911 \\
& Marigold
& 0.175 & 0.730
& 0.094 & 0.910
& 0.098 & 0.910
& 0.388 & 0.517
& 0.082 & 0.933 \\
\cmidrule(lr){2-12}
& UGG
& 0.073 & 0.952
& 0.083 & 0.921
& 0.064 & 0.960
& 0.303 & 0.654
& 0.060 & 0.963 \\
& UGG (TTS)
& \secondbest{0.072} & \secondbest{0.954}
& 0.084 & 0.920
& 0.063 & 0.962
& 0.293 & 0.661
& 0.059 & 0.969 \\
& UGG (TTS Oracle)
& \best{0.063} & \best{0.963}
& \best{0.074} & \secondbest{0.923}
& \best{0.055} & \secondbest{0.967}
& \secondbest{0.263} & \best{0.692}
& \best{0.056} & \best{0.973} \\

\midrule

\multirow{8}{*}{\rotatebox[origin=c]{90}{Sparse}}
& ChronoDepth
& 0.199 & 0.700
& 0.130 & 0.844
& 0.089 & 0.905
& 0.424 & 0.487
& 0.072 & 0.960 \\
& Geo4D
& 0.177 & 0.737
& 0.098 & 0.912
& 0.100 & 0.897
& 0.310 & 0.554
& 0.067 & 0.966 \\
& Gen3R
& 0.238 & 0.620
& 0.124 & 0.867
& 0.104 & 0.888
& 0.347 & 0.510
& 0.127 & 0.847 \\
& Aether
& 0.269 & 0.567
& 0.167 & 0.772
& 0.149 & 0.786
& 0.487 & 0.424
& 0.138 & 0.849 \\
& Marigold
& 0.169 & 0.733
& 0.106 & 0.890
& \secondbest{0.087} & 0.921
& 0.345 & 0.527
& 0.071 & 0.970 \\
\cmidrule(lr){2-12}
& UGG
& 0.101 & 0.884
& 0.087 & \secondbest{0.916}
& 0.099 & 0.917
& 0.258 & 0.660
& 0.049 & \secondbest{0.977} \\
& UGG (TTS)
& \secondbest{0.082} & \secondbest{0.917}
& \secondbest{0.086} & \secondbest{0.916}
& \secondbest{0.087} & \secondbest{0.934}
& \secondbest{0.219} & \secondbest{0.678}
& \secondbest{0.047} & \best{0.979} \\
& UGG (TTS Oracle)
& \best{0.055} & \best{0.966}
& \best{0.075} & \best{0.923}
& \best{0.068} & \best{0.946}
& \best{0.177} & \best{0.781}
& \best{0.043} & \best{0.979} \\

\bottomrule
\end{tabular*}
\endgroup
\vspace{-10pt}
\end{table}
\subsection{Depth Prediction}
\label{sec:depth_estimation}

We evaluate depth prediction from RGB observations in the dense and sparse settings described in the previous section.
In both cases, depth tokens are initialized from noise and denoised, testing whether the unified conditional-completion model learns a useful geometric
posterior over scene depth.

\textbf{Benchmarks and Metrics.}
We evaluate on ScanNet++, 7-Scenes~\cite{shotton2013scene}, KITTI, Sintel, and Bonn. These datasets
cover driving scenes, synthetic video, and indoor RGB-D scans. Following Geo4D~\cite{jiang2025geo4d}, we apply video-level scale and shift alignment before
evaluation, and report absolute relative error between the true and predicted depth (Abs Rel) and threshold accuracy, whether the predicted depth differs from the ground truth by less than a factor of $1.25$ ($\delta < 1.25$).

\textbf{Baselines.}
We compare with Geo4D, ChronoDepth~\cite{shao2025chronodepth}, Gen3R, Aether, and Marigold~\cite{ke2023marigold}.
These baselines include specialized depth and 4D geometry models, as well as a unified model, while ours is trained for RGB-depth-camera completion.

\textbf{Evaluation results.}
We report depth prediction results in Table~\ref{tab:final_depth_table}. 
Overall, our model is competitive with specialized depth and 4D geometry baselines across both dense and sparse settings. 
In the dense setting, it obtains the lowest Abs Rel and highest $\delta < 1.25$ on ScanNet++ among the external baselines. 
In the sparse setting, where only an image pair is available, our model improves over the baselines on both metrics on 4 out of 5 datasets. 
At the same time, Geo4D remains stronger on Sintel and 7-Scenes metrics for the dense setting, indicating that specialized geometry models may still have advantages in some settings. 
Figure~\ref{fig:depth_results} provides qualitative comparisons. 
The zoom-in crops show that our predictions preserve sharper object boundaries and more coherent local structures in highlighted regions. 
This suggests that joint RGB-depth training, together with the generative completion objective, helps the model recover fine-scale depth discontinuities while maintaining globally coherent depth maps.

\begin{figure}[t]
    \centering
    \includegraphics[width=0.95\textwidth]{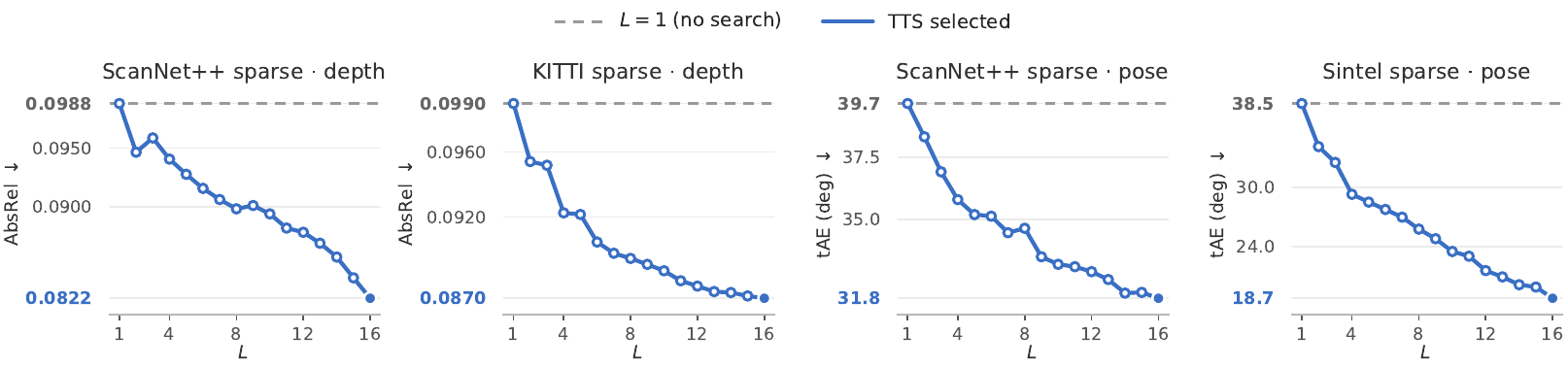}
    \caption{\textbf{Test-time search with increasing sample budget.}
We sample up to 16 candidate completions and select one using the image-based depth or pose verifier.
The dashed line shows the single-sample result, and the solid curve shows the selected result as $L$ increases.
Search improves several pose and depth settings, showing that the generative model produces useful candidate diversity.}
\label{fig:tts}
\vspace{-17pt}
\end{figure}
\subsection{Test-Time Search on Geometric Prediction}
\label{sec:test_time_search}

We evaluate whether test-time search (TTS) improves geometric prediction by sampling multiple RGB-depth-camera completions at inference time. For each input, we generate 16 candidates from different noise initializations and select depth and pose independently using their respective verification scores, as described in Section~\ref{sec:inference_time_search}. 

\looseness=-1
Tables~\ref{tab:final_depth_table} and~\ref{tab:final_pose_table} show that \emph{UGG (TTS)} improves sparse depth Abs Rel and pose translation angular error on all five benchmarks. It also improves dense depth Abs Rel and pose ATE on four of five benchmarks. These results demonstrate that the verifier selects lower-error depth and camera translation estimates from the model's sampled hypotheses using image evidence alone. Figure~\ref{fig:tts} studies how pose and depth accuracy vary with the number of sampled candidates. Larger search budgets often improve the selected result, but the gains are not uniform because the verification score is only a proxy for ground-truth geometric accuracy.

\emph{UGG (TTS Oracle)} selects one of the same 16 candidates per input by minimizing ground-truth scale-and-shift-aligned Abs Rel for depth, ATE for dense pose, or tAE for sparse pose. All metrics are reported for the selected candidate. Overall, these results show that inference-time search is a promising benefit of the generative formulation, while leaving room for stronger search criteria and verification.

\subsection{Ablation Studies}
\label{sec:ablations}
\providecommand{\best}[1]{\cellcolor{blue!12}#1}
\providecommand{\secondbest}[1]{\cellcolor{blue!5}#1}
\providecommand{\abltitle}[1]{\textbf{\strut #1\strut}\\[0pt]}
\newcolumntype{Q}{>{\centering\arraybackslash}X}

\begin{table}[t]
\centering
\caption{\textbf{Ablations.}
Each row changes one training task, dataset, architecture, or camera representation.
We report Abs Rel~($\downarrow$) and $\delta_{1.25}$~($\uparrow$) for depth,
ATE~($\downarrow$) and RPE$_\mathrm{r}$~($\downarrow$, degrees) for pose,
and PSNR~($\uparrow$) for generation.
RE10K: RealEstate10K; Spatial: SpatialVID; gen.: RGB generation; tr.: transformer;
Dense rays: dense ray maps; Global cam.: global camera tokens.
Shading marks the best result within each block.}
\label{tab:ablations}
\begingroup
\fontsize{8}{9}\selectfont
\setlength{\tabcolsep}{0.4pt}
\renewcommand{\arraystretch}{0.9}
\setlength{\aboverulesep}{0.15ex}
\setlength{\belowrulesep}{0.2ex}

\begin{minipage}[t]{0.325\linewidth}
\centering
\abltitle{(a) Depth prediction}
\begin{tabularx}{\linewidth}{@{}l*{4}{Q}@{}}
\toprule
& \multicolumn{2}{c}{Bonn} & \multicolumn{2}{c}{KITTI} \\
\cmidrule(lr){2-3} \cmidrule(l){4-5}
Variant & Rel & $\delta_{1.25}$ & Rel & $\delta_{1.25}$ \\
\midrule
Full model & 0.135 & 0.843 & \best{0.152} & \best{0.791} \\
w/o gen. & 0.178 & 0.782 & 0.322 & 0.486 \\
w/o RE10K & 0.239 & 0.651 & 0.196 & 0.732 \\
Single tr. & \best{0.129} & \best{0.870} & 0.248 & 0.640 \\
\bottomrule
\end{tabularx}
\end{minipage}%
\hfill
\begin{minipage}[t]{0.33\linewidth}
\centering
\abltitle{(b) Camera pose estimation}
\begin{tabularx}{\linewidth}{@{}l*{4}{Q}@{}}
\toprule
& \multicolumn{2}{c}{ScanNetV2} & \multicolumn{2}{c}{ScanNet++} \\
\cmidrule(lr){2-3} \cmidrule(l){4-5}
Variant & ATE & RPE$_\mathrm{r}$ & ATE & RPE$_\mathrm{r}$ \\
\midrule
Full model & \best{0.074} & \best{2.91} & \best{0.193} & \best{6.18} \\
w/o depth & 0.081 & 4.30 & 0.234 & 7.57 \\
Single tr. & 0.083 & 3.02 & 0.204 & 6.60 \\
\midrule
Dense rays & \best{0.071} & \best{2.80} & \best{0.206} & \best{6.09} \\
Global cam. & 0.103 & 3.44 & 0.233 & 6.61 \\
\bottomrule
\end{tabularx}
\end{minipage}%
\hfill
\begin{minipage}[t]{0.315\linewidth}
\centering
\abltitle{(c) Camera-controlled generation}
\begin{tabularx}{\linewidth}{@{}l*{3}{Q}@{}}
\toprule
& \multicolumn{3}{c}{PSNR} \\
\cmidrule(l){2-4}
Variant & RE10K & Spatial & DL3DV \\
\midrule
Full model & \best{20.18} & \best{13.67} & \best{12.60} \\
Global cam. & 20.03 & 12.89 & 11.64 \\
Single tr. & 19.68 & 12.75 & 11.61 \\
w/o pose & 13.26 & 11.87 & 10.95 \\
\bottomrule
\end{tabularx}
\end{minipage}\par
\endgroup
\vspace{-18pt}
\end{table}

\textbf{Setup.}
We study the effects of joint training, training data, architecture,
and camera representation using the smaller Wan2.1-1.3B instead of
the final Wan2.2-5B model. Unless otherwise specified, we train on
TartanAir and RE10K for 7,500 optimizer steps at batch size 8, with
the same optimization settings. Additional details on the ablation
setup are provided in Appendix~\ref{app:ablation_setup}.

\textbf{Joint training.}
Learning to generate RGB video also helps the model recover
geometry. On KITTI, depth Abs Rel more than doubles when
RGB generation is removed from training, rising from 0.152
to 0.322. Geometry supervision also benefits the other tasks:
depth prediction improves camera estimation, while pose
estimation improves camera-controlled generation by
6.92\,dB on RE10K (Table~\ref{tab:ablations}a--c).
Joint training therefore improves what the model learns
about both scene appearance and geometry.

\textbf{Heterogeneous training data.}
To test whether these benefits extend to data with incomplete
annotations, we add RE10K, which provides RGB and camera
annotations but no depth supervision. Adding it to TartanAir
training improves depth prediction on both Bonn and KITTI,
reducing Abs Rel from 0.239 to 0.135 and from 0.196 to 0.152,
respectively (Table~\ref{tab:ablations}a). The model can
therefore use RGB and camera supervision from additional
datasets to improve depth prediction without requiring
depth labels from those datasets.

\textbf{Architecture.}
MoT allows the modalities to share information through
attention while retaining separate processing branches.
Compared with a single shared transformer, this design
improves camera estimation on both benchmarks and generation
on all three. Its effect on depth is less uniform:
MoT performs better on KITTI, while the shared transformer
is slightly better on Bonn (Table~\ref{tab:ablations}a--c).

\textbf{Camera representation.}
We compare dense ray maps with global camera tokens in
models trained only for pose estimation
(Table~\ref{tab:ablations}b). Dense ray maps improve both
trajectory and rotation accuracy on ScanNetV2 and ScanNet++.
They also improve camera-controlled generation, yielding
higher PSNR than global camera tokens on all three benchmarks
(Table~\ref{tab:ablations}c). Aligning the camera representation
with image locations is thus useful both for recovering
cameras from images and for generating images from cameras. Additional results on history guidance are provided in
Appendix~\ref{app:history_guidance}.

\vspace{-3pt}
\section{Conclusion}
\vspace{-3pt}
\label{sec:conclusion}
We present a unified generative geometry model for 4D scene understanding, formulating visual generation and geometric prediction as conditional completion over RGB, depth, and camera rays. 
Through noise patterns over modality tokens, a single model supports camera-controlled generation and geometric prediction under dense and sparse observations, while learning from heterogeneous datasets with varying modality availability.
Across diverse datasets, UGG achieves strong video generation, depth prediction, and pose estimation performance, and further improves geometric consistency through test-time search. 
These results suggest that jointly modeling appearance and geometry with generative models is a promising direction toward more general 4D scene understanding.

\paragraph{Limitations.}
Our current model is not uniformly stronger than specialized methods: task-specific NVS models still perform better on pixel-aligned metrics, and specialized geometry models remain stronger on several indoor depth metrics. 
Future work could address these gaps with better dataset balancing, stronger geometric supervision, and improved depth-camera consistency during training. 
Our test-time search also relies on RGB reprojection error, which is useful but not uniformly reliable; learned or multimodal consistency scores may provide stronger search criteria. 
Finally, training at larger scale, higher resolution, and with more diverse data could further improve both visual detail and geometric accuracy.
\begin{ack}
We thank the Kempner Institute at Harvard University for
providing computational resources that supported this work. Ayush Tewari acknowledges the NVIDIA Academic Grant Program.
\end{ack}

\bibliographystyle{plain}
\bibliography{references}

\begin{thebibliography}{10}

\bibitem{avetisyan2024scenescript}
Armen Avetisyan, Christopher Xie, Henry Howard-Jenkins, Tsun-Yi Yang, Samir Aroudj, Suvam Patra, Fuyang Zhang, Duncan Frost, Luke Holland, Campbell Orme, Jakob Engel, Edward Miller, Richard Newcombe, and Vasileios Balntas.
\newblock {SceneScript}: Reconstructing scenes with an autoregressive structured language model.
\newblock In {\em European Conference on Computer Vision (ECCV)}, 2024.

\bibitem{genie3}
Philip~J. Ball, Jakob Bauer, Frank Belletti, Bethanie Brownfield, Ariel Ephrat, Shlomi Fruchter, Agrim Gupta, Kristian Holsheimer, Aleksander Holynski, Jiri Hron, Christos Kaplanis, Marjorie Limont, Matt McGill, Yanko Oliveira, Jack Parker-Holder, Frank Perbet, Guy Scully, Jeremy Shar, Stephen Spencer, Omer Tov, Ruben Villegas, Emma Wang, Jessica Yung, Cip Baetu, Jordi Berbel, David Bridson, Jake Bruce, Gavin Buttimore, Sarah Chakera, Bilva Chandra, Paul Collins, Alex Cullum, Bogdan Damoc, Vibha Dasagi, Maxime Gazeau, Charles Gbadamosi, Woohyun Han, Ed~Hirst, Ashyana Kachra, Lucie Kerley, Kristian Kjems, Eva Knoepfel, Vika Koriakin, Jessica Lo, Cong Lu, Zeb Mehring, Alex Moufarek, Henna Nandwani, Valeria Oliveira, Fabio Pardo, Jane Park, Andrew Pierson, Ben Poole, Helen Ran, Tim Salimans, Manuel Sanchez, Igor Saprykin, Amy Shen, Sailesh Sidhwani, Duncan Smith, Joe Stanton, Hamish Tomlinson, Dimple Vijaykumar, Luyu Wang, Piers Wingfield, Nat Wong, Keyang Xu, Christopher Yew, Nick Young, Vadim Zubov, Douglas Eck, Dumitru Erhan, Koray Kavukcuoglu, Demis Hassabis, Zoubin Gharamani, Raia Hadsell, A{\"a}ron van~den Oord, Inbar Mosseri, Adrian Bolton, Satinder Singh, and Tim Rockt{\"a}schel.
\newblock Genie 3: A new frontier for world models.
\newblock 2025.

\bibitem{battaglia2013simulation}
Peter~W Battaglia, Jessica~B Hamrick, and Joshua~B Tenenbaum.
\newblock Simulation as an engine of physical scene understanding.
\newblock {\em Proceedings of the national academy of sciences}, 110(45):18327--18332, 2013.

\bibitem{bi2025motus}
Hongzhe Bi, Hengkai Tan, Shenghao Xie, Zeyuan Wang, Shuhe Huang, Haitian Liu, Ruowen Zhao, Yao Feng, Chendong Xiang, Yinze Rong, et~al.
\newblock Motus: A unified latent action world model.
\newblock {\em arXiv preprint arXiv:2512.13030}, 2025.

\bibitem{black2023bedlam}
Michael~J. Black, Priyanka Patel, Joachim Tesch, and Jinlong Yang.
\newblock {BEDLAM}: A synthetic dataset of bodies exhibiting detailed lifelike animated motion.
\newblock In {\em Proceedings of the IEEE/CVF Conference on Computer Vision and Pattern Recognition (CVPR)}, 2023.

\bibitem{blattmann2023SVD}
Andreas Blattmann, Tim Dockhorn, Sumith Kulal, Daniel Mendelevitch, Maciej Kilian, Dominik Lorenz, Yam Levi, Zion English, Vikram Voleti, Adam Letts, et~al.
\newblock Stable video diffusion: Scaling latent video diffusion models to large datasets.
\newblock {\em arXiv preprint arXiv:2311.15127}, 2023.

\bibitem{Butler:ECCV:2012}
D.~J. Butler, J.~Wulff, G.~B. Stanley, and M.~J. Black.
\newblock A naturalistic open source movie for optical flow evaluation.
\newblock In {A. Fitzgibbon et al. (Eds.)}, editor, {\em European Conf. on Computer Vision (ECCV)}, Part IV, LNCS 7577, pages 611--625. Springer-Verlag, October 2012.

\bibitem{cabon2020vkitti}
Yohann Cabon, Naila Murray, and Martin Humenberger.
\newblock Virtual kitti 2.
\newblock {\em arXiv preprint arXiv:2001.10773}, 2020.

\bibitem{cai2025interpose}
Ruojin Cai, Jason~Y Zhang, Philipp Henzler, Zhengqi Li, Noah Snavely, and Ricardo Martin-Brualla.
\newblock Can generative video models help pose estimation?
\newblock In {\em Proceedings of the IEEE/CVF Conference on Computer Vision and Pattern Recognition}, pages 16764--16773, 2025.

\bibitem{chen2024diffusionforcing}
Boyuan Chen, Diego Mart{\'\i}~Mons{\'o}, Yilun Du, Max Simchowitz, Russ Tedrake, and Vincent Sitzmann.
\newblock Diffusion forcing: Next-token prediction meets full-sequence diffusion.
\newblock {\em Advances in Neural Information Processing Systems}, 37:24081--24125, 2024.

\bibitem{chen2025LVP}
Boyuan Chen, Tianyuan Zhang, Haoran Geng, Kiwhan Song, Caiyi Zhang, Peihao Li, William~T Freeman, Jitendra Malik, Pieter Abbeel, Russ Tedrake, et~al.
\newblock Large video planner enables generalizable robot control.
\newblock {\em arXiv preprint arXiv:2512.15840}, 2025.

\bibitem{chen2025skyreelsv2infinitelengthfilmgenerative}
Guibin Chen, Dixuan Lin, Jiangping Yang, Chunze Lin, Junchen Zhu, Mingyuan Fan, Hao Zhang, Sheng Chen, Zheng Chen, Chengcheng Ma, Weiming Xiong, Wei Wang, Nuo Pang, Kang Kang, Zhiheng Xu, Yuzhe Jin, Yupeng Liang, Yubing Song, Peng Zhao, Boyuan Xu, Di~Qiu, Debang Li, Zhengcong Fei, Yang Li, and Yahui Zhou.
\newblock Skyreels-v2: Infinite-length film generative model, 2025.

\bibitem{dai2017scannet}
Angela Dai, Angel~X. Chang, Manolis Savva, Maciej Halber, Thomas Funkhouser, and Matthias Nie{\ss}ner.
\newblock Scannet: Richly-annotated 3d reconstructions of indoor scenes.
\newblock In {\em Proc. Computer Vision and Pattern Recognition (CVPR), IEEE}, 2017.

\bibitem{fang2025worldreel}
Shaoheng Fang, Hanwen Jiang, Yunpeng Bai, Niloy~J Mitra, and Qixing Huang.
\newblock Worldreel: 4d video generation with consistent geometry and motion modeling.
\newblock {\em arXiv preprint arXiv:2512.07821}, 2025.

\bibitem{fonder2019midair}
Micha{\"e}l Fonder and Marc Van~Droogenbroeck.
\newblock {Mid-Air}: A multi-modal dataset for extremely low altitude drone flights.
\newblock In {\em Proceedings of the IEEE/CVF Conference on Computer Vision and Pattern Recognition Workshops (CVPRW)}, 2019.

\bibitem{he2024cameractrl}
Hao He, Yinghao Xu, Yuwei Guo, Gordon Wetzstein, Bo~Dai, Hongsheng Li, and Ceyuan Yang.
\newblock Cameractrl: Enabling camera control for video diffusion models.
\newblock In {\em International Conference on Learning Representations}, 2025.

\bibitem{hu2024depthcrafter}
Wenbo Hu, Xiangjun Gao, Xiaoyu Li, Sijie Zhao, Xiaodong Cun, Yong Zhang, Long Quan, and Ying Shan.
\newblock Depthcrafter: Generating consistent long depth sequences for open-world videos.
\newblock In {\em CVPR}, 2025.

\bibitem{Huang_2026_CVPR}
Jiaxin Huang, Yuanbo Yang, Bangbang Yang, Lin Ma, Yuewen Ma, and Yiyi Liao.
\newblock Gen3r: 3d scene generation meets feed-forward reconstruction.
\newblock In {\em Proceedings of the IEEE/CVF Conference on Computer Vision and Pattern Recognition (CVPR)}, pages 25358--25369, June 2026.

\bibitem{huang2018deepmvs}
Po-Han Huang, Kevin Matzen, Johannes Kopf, Narendra Ahuja, and Jia-Bin Huang.
\newblock {DeepMVS}: Learning multi-view stereopsis.
\newblock In {\em Proceedings of the IEEE Conference on Computer Vision and Pattern Recognition (CVPR)}, 2018.

\bibitem{huang2025voyager}
Tianyu Huang, Wangguandong Zheng, Tengfei Wang, Yuhao Liu, Zhenwei Wang, Junta Wu, Jie Jiang, Hui Li, Rynson Lau, Wangmeng Zuo, et~al.
\newblock Voyager: Long-range and world-consistent video diffusion for explorable 3d scene generation.
\newblock {\em ACM Transactions on Graphics (TOG)}, 44(6):1--15, 2025.

\bibitem{jiang2025geo4d}
Zeren Jiang, Chuanxia Zheng, Iro Laina, Diane Larlus, and Andrea Vedaldi.
\newblock Geo4d: Leveraging video generators for geometric 4d scene reconstruction.
\newblock In {\em Proceedings of the IEEE/CVF International Conference on Computer Vision}, pages 20658--20671, 2025.

\bibitem{ke2023marigold}
Bingxin Ke, Anton Obukhov, Shengyu Huang, Nando Metzger, Rodrigo~Caye Daudt, and Konrad Schindler.
\newblock Repurposing diffusion-based image generators for monocular depth estimation.
\newblock In {\em Proceedings of the IEEE/CVF Conference on Computer Vision and Pattern Recognition (CVPR)}, 2024.

\bibitem{keetha2025mapanythinguniversalfeedforwardmetric}
Nikhil Keetha, Norman Müller, Johannes Schönberger, Lorenzo Porzi, Yuchen Zhang, Tobias Fischer, Arno Knapitsch, Duncan Zauss, Ethan Weber, Nelson Antunes, Jonathon Luiten, Manuel Lopez-Antequera, Samuel~Rota Bulò, Christian Richardt, Deva Ramanan, Sebastian Scherer, and Peter Kontschieder.
\newblock Mapanything: Universal feed-forward metric 3d reconstruction, 2025.

\bibitem{kersten2004object}
Daniel Kersten, Pascal Mamassian, and Alan Yuille.
\newblock Object perception as bayesian inference.
\newblock {\em Annu. Rev. Psychol.}, 55(1):271--304, 2004.

\bibitem{leroy2024groundingimagematching3d}
Vincent Leroy, Yohann Cabon, and Jérôme Revaud.
\newblock Grounding image matching in 3d with mast3r, 2024.

\bibitem{li2023matrixcity}
Yixuan Li, Lihan Jiang, Linning Xu, Yuanbo Xiangli, Zhenzhi Wang, Dahua Lin, and Bo~Dai.
\newblock {MatrixCity}: A large-scale city dataset for city-scale neural rendering and beyond.
\newblock In {\em Proceedings of the IEEE/CVF International Conference on Computer Vision (ICCV)}, 2023.

\bibitem{liang2024mot}
Weixin Liang, Lili Yu, Liang Luo, Srinivasan Iyer, Ning Dong, Chunting Zhou, Gargi Ghosh, Mike Lewis, Wen-tau Yih, Luke Zettlemoyer, et~al.
\newblock Mixture-of-transformers: A sparse and scalable architecture for multi-modal foundation models.
\newblock {\em arXiv preprint arXiv:2411.04996}, 2024.

\bibitem{lillemark2026flowequivariantworldmodels}
Hansen~Jin Lillemark, Benhao Huang, Fangneng Zhan, Yilun Du, and Thomas~Anderson Keller.
\newblock Flow equivariant world models: Structured memory for dynamic environments.
\newblock In {\em International Conference on Machine Learning (ICML)}, 2026.

\bibitem{ling2024dl3dv}
Lu~Ling, Yicheng Sheng, Zhi Tu, Wentian Zhao, Cheng Xin, Kun Wan, Lantao Yu, Qianyu Guo, Zixun Yu, Yawen Lu, et~al.
\newblock Dl3dv-10k: A large-scale scene dataset for deep learning-based 3d vision.
\newblock In {\em Proceedings of the IEEE/CVF Conference on Computer Vision and Pattern Recognition (CVPR)}, 2024.

\bibitem{lipman2023flow}
Yaron Lipman, Ricky T.~Q. Chen, Heli Ben-Hamu, Maximilian Nickel, and Matthew Le.
\newblock {Flow Matching for Generative Modeling}.
\newblock In {\em International Conference on Learning Representations}, 2023.

\bibitem{liu2025videot1}
Fangfu Liu, Hanyang Wang, Yimo Cai, Kaiyan Zhang, Xiaohang Zhan, and Yueqi Duan.
\newblock Video-t1: Test-time scaling for video generation.
\newblock In {\em Proceedings of the IEEE/CVF international conference on computer vision}, pages 18671--18681, 2025.

\bibitem{mccormac2017scenenet}
John McCormac, Ankur Handa, Stefan Leutenegger, and Andrew~J. Davison.
\newblock {SceneNet RGB-D}: Can 5m synthetic images beat generic imagenet pre-training on indoor segmentation?
\newblock In {\em Proceedings of the IEEE International Conference on Computer Vision (ICCV)}, 2017.

\bibitem{mehl2023spring}
Lukas Mehl, Jenny Schmalfuss, Azin Jahedi, Yaroslava Nalivayko, and Andr{\'e}s Bruhn.
\newblock Spring: A high-resolution high-detail dataset and benchmark for scene flow, optical flow and stereo.
\newblock In {\em Proceedings of the IEEE/CVF Conference on Computer Vision and Pattern Recognition (CVPR)}, 2023.

\bibitem{mione4d2025}
Zhenxing Mi, Yuxin Wang, and Dan Xu.
\newblock One4d: Unified 4d generation and reconstruction via decoupled lora control.
\newblock {\em arXiv preprint arXiv:2511.18922}, 2025.

\bibitem{olshausen2014perceptionasinference}
Bruno~A. Olshausen.
\newblock Perception as an inference problem.
\newblock In Michael~S. Gazzaniga and George~R. Mangun, editors, {\em The Cognitive Neurosciences}. The MIT Press, 10 2014.

\bibitem{oquab2023dinov2}
Maxime Oquab, Timoth{\'e}e Darcet, Th{\'e}o Moutakanni, Huy~V. Vo, Marc Szafraniec, Vasil Khalidov, Pierre Fernandez, Daniel Haziza, Francisco Massa, Alaaeldin El-Nouby, Mido Assran, Nicolas Ballas, Wojciech Galuba, Russell Howes, Po-Yao Huang, Shang-Wen Li, Ishan Misra, Michael Rabbat, Vasu Sharma, Gabriel Synnaeve, Hu~Xu, Herve Jegou, Julien Mairal, Patrick Labatut, Armand Joulin, and Piotr Bojanowski.
\newblock {DINO}v2: Learning robust visual features without supervision.
\newblock {\em Transactions on Machine Learning Research}, 2024.

\bibitem{peebles2023scalable}
William Peebles and Saining Xie.
\newblock Scalable diffusion models with transformers.
\newblock In {\em 2023 IEEE/CVF International Conference on Computer Vision (ICCV)}, pages 4172--4182. IEEE, 2023.

\bibitem{ren2025gen3c}
Xuanchi Ren, Tianchang Shen, Jiahui Huang, Huan Ling, Yifan Lu, Merlin Nimier-David, Thomas M{\"u}ller, Alexander Keller, Sanja Fidler, and Jun Gao.
\newblock Gen3c: 3d-informed world-consistent video generation with precise camera control.
\newblock In {\em Proceedings of the IEEE/CVF Conference on Computer Vision and Pattern Recognition}, pages 6121--6132, 2025.

\bibitem{shao2025chronodepth}
Jiahao Shao, Yuanbo Yang, Hongyu Zhou, Youmin Zhang, Yujun Shen, Vitor Guizilini, Yue Wang, Matteo Poggi, and Yiyi Liao.
\newblock Learning temporally consistent video depth from video diffusion priors.
\newblock In {\em Proceedings of the Computer Vision and Pattern Recognition Conference}, pages 22841--22852, 2025.

\bibitem{shen2025fastvggt}
You Shen, Zhipeng Zhang, Yansong Qu, and Liujuan Cao.
\newblock Fastvggt: Training-free acceleration of visual geometry transformer.
\newblock {\em arXiv preprint arXiv:2509.02560}, 2025.

\bibitem{shotton2013scene}
Jamie Shotton, Ben Glocker, Christopher Zach, Shahram Izadi, Antonio Criminisi, and Andrew Fitzgibbon.
\newblock Scene coordinate regression forests for camera relocalization in rgb-d images.
\newblock In {\em Proceedings of the IEEE conference on computer vision and pattern recognition}, pages 2930--2937, 2013.

\bibitem{sitzmann2021light}
Vincent Sitzmann, Semon Rezchikov, Bill Freeman, Josh Tenenbaum, and Fredo Durand.
\newblock Light field networks: Neural scene representations with single-evaluation rendering.
\newblock {\em Advances in neural information processing systems}, 34:19313--19325, 2021.

\bibitem{song2025dfot}
Kiwhan Song, Boyuan Chen, Max Simchowitz, Yilun Du, Russ Tedrake, and Vincent Sitzmann.
\newblock History-guided video diffusion, 2025.

\bibitem{sturm2012tumdynamics}
Jürgen Sturm, Nikolas Engelhard, Felix Endres, Wolfram Burgard, and Daniel Cremers.
\newblock A benchmark for the evaluation of rgb-d slam systems.
\newblock In {\em 2012 IEEE/RSJ International Conference on Intelligent Robots and Systems}, pages 573--580, 2012.

\bibitem{su2024roformer}
Jianlin Su, Murtadha Ahmed, Yu~Lu, Shengfeng Pan, Wen Bo, and Yunfeng Liu.
\newblock Roformer: Enhanced transformer with rotary position embedding.
\newblock {\em Neurocomputing}, 568:127063, 2024.

\bibitem{sundaram2010dense}
Narayanan Sundaram, Thomas Brox, and Kurt Keutzer.
\newblock Dense point trajectories by {GPU}-accelerated large displacement optical flow.
\newblock In {\em European Conference on Computer Vision (ECCV)}, 2010.

\bibitem{teed2020raft}
Zachary Teed and Jia Deng.
\newblock {RAFT}: Recurrent all-pairs field transforms for optical flow.
\newblock In {\em Computer Vision -- ECCV 2020}, pages 402--419. Springer, 2020.

\bibitem{tesch2025bedlam2}
Joachim Tesch, Giorgio Becherini, Prerana Achar, Anastasios Yiannakidis, Muhammed Kocabas, Priyanka Patel, and Michael~J. Black.
\newblock {BEDLAM2.0}: Synthetic humans and cameras in motion.
\newblock In {\em Advances in Neural Information Processing Systems (NeurIPS), Datasets and Benchmarks Track}, 2025.

\bibitem{wan2025wan}
Team Wan, Ang Wang, Baole Ai, Bin Wen, Chaojie Mao, Chen-Wei Xie, Di~Chen, Feiwu Yu, Haiming Zhao, Jianxiao Yang, et~al.
\newblock Wan: Open and advanced large-scale video generative models.
\newblock {\em arXiv preprint arXiv:2503.20314}, 2025.

\bibitem{wang2025spatialvid}
Jiahao Wang, Yufeng Yuan, Rujie Zheng, Youtian Lin, Jian Gao, Lin-Zhuo Chen, Yajie Bao, Yi~Zhang, Chang Zeng, Yanxi Zhou, et~al.
\newblock Spatialvid: A large-scale video dataset with spatial annotations.
\newblock {\em arXiv preprint arXiv:2509.09676}, 2025.

\bibitem{wang2025vggt}
Jianyuan Wang, Minghao Chen, Nikita Karaev, Andrea Vedaldi, Christian Rupprecht, and David Novotny.
\newblock Vggt: Visual geometry grounded transformer.
\newblock In {\em Proceedings of the Computer Vision and Pattern Recognition Conference}, pages 5294--5306, 2025.

\bibitem{wang2026eva}
Ruixiang Wang, Qingming Liu, Yueci Deng, Guiliang Liu, Zhen Liu, and Kui Jia.
\newblock Eva: Aligning video world models with executable robot actions via inverse dynamics rewards.
\newblock {\em arXiv preprint arXiv:2603.17808}, 2026.

\bibitem{wang2023dust3r}
Shuzhe Wang, Vincent Leroy, Yohann Cabon, Boris Chidlovskii, and Jerome Revaud.
\newblock Dust3r: Geometric 3d vision made easy.
\newblock {\em arXiv preprint arXiv:2312.14132}, 2023.

\bibitem{wang2020tartanair}
Wenshan Wang, Delong Zhu, Xiangwei Wang, Yaoyu Hu, Yuheng Qiu, Chen Wang, Yafei Hu, Ashish Kapoor, and Sebastian Scherer.
\newblock Tartanair: A dataset to push the limits of visual slam.
\newblock In {\em 2020 IEEE/RSJ International Conference on Intelligent Robots and Systems (IROS)}, pages 4909--4916. IEEE, 2020.

\bibitem{wang2025pi3}
Yifan Wang, Jianjun Zhou, Haoyi Zhu, Wenzheng Chang, Yang Zhou, Zizun Li, Junyi Chen, Jiangmiao Pang, Chunhua Shen, and Tong He.
\newblock $\pi^3$: Scalable permutation-equivariant visual geometry learning, 2025.

\bibitem{wang2004image}
Zhou Wang, A.~C. Bovik, H.~R. Sheikh, and E.~P. Simoncelli.
\newblock Image quality assessment: from error visibility to structural similarity.
\newblock {\em Trans. Img. Proc.}, 13(4):600–612, April 2004.

\bibitem{wang2024motionctrl}
Zhouxia Wang, Ziyang Yuan, Xintao Wang, Yaowei Li, Tianshui Chen, Menghan Xia, Ping Luo, and Ying Shan.
\newblock Motionctrl: A unified and flexible motion controller for video generation.
\newblock In {\em ACM SIGGRAPH 2024 Conference Papers}, pages 1--11, 2024.

\bibitem{wu2026geometryforcing}
Haoyu Wu, Diankun Wu, Tianyu He, Junliang Guo, Yang Ye, Yueqi Duan, and Jiang Bian.
\newblock Geometry forcing: Marrying video diffusion and 3d representation for consistent world modeling.
\newblock In {\em The Fourteenth International Conference on Learning Representations}, 2026.

\bibitem{wu2025video}
Tong Wu, Shuai Yang, Ryan Po, Yinghao Xu, Ziwei Liu, Dahua Lin, and Gordon Wetzstein.
\newblock Video world models with long-term spatial memory.
\newblock {\em arXiv preprint arXiv:2506.05284}, 2025.

\bibitem{yang2025fast3r}
Jianing Yang, Alexander Sax, Kevin~J Liang, Mikael Henaff, Hao Tang, Ang Cao, Joyce Chai, Franziska Meier, and Matt Feiszli.
\newblock Fast3r: Towards 3d reconstruction of 1000+ images in one forward pass.
\newblock In {\em Proceedings of the Computer Vision and Pattern Recognition Conference}, pages 21924--21935, 2025.

\bibitem{yang2026neoverse}
Yuxue Yang, Lue Fan, Ziqi Shi, Junran Peng, Feng Wang, and Zhaoxiang Zhang.
\newblock Neoverse: Enhancing 4d world model with in-the-wild monocular videos.
\newblock {\em arXiv preprint arXiv:2601.00393}, 2026.

\bibitem{ye2026dreamzero}
Seonghyeon Ye, Yunhao Ge, Kaiyuan Zheng, Shenyuan Gao, Sihyun Yu, George Kurian, Suneel Indupuru, You~Liang Tan, Chuning Zhu, Jiannan Xiang, et~al.
\newblock World action models are zero-shot policies.
\newblock {\em arXiv preprint arXiv:2602.15922}, 2026.

\bibitem{yeshwanth2023scannetpp}
Chandan Yeshwanth, Yueh-Cheng Liu, Matthias Nie{\ss}ner, and Angela Dai.
\newblock Scannet++: A high-fidelity dataset of 3d indoor scenes.
\newblock In {\em Proceedings of the IEEE/CVF International Conference on Computer Vision (ICCV)}, 2023.

\bibitem{you2024nvs}
Meng You, Zhiyu Zhu, Hui Liu, and Junhui Hou.
\newblock Nvs-solver: Video diffusion model as zero-shot novel view synthesizer.
\newblock {\em arXiv preprint arXiv:2405.15364}, 2024.

\bibitem{yu2024viewcrafter}
Wangbo Yu, Jinbo Xing, Li~Yuan, Wenbo Hu, Xiaoyu Li, Zhipeng Huang, Xiangjun Gao, Tien-Tsin Wong, Ying Shan, and Yonghong Tian.
\newblock Viewcrafter: Taming video diffusion models for high-fidelity novel view synthesis.
\newblock {\em arXiv preprint arXiv:2409.02048}, 2024.

\bibitem{yuille2006vision}
Alan Yuille and Daniel Kersten.
\newblock Vision as bayesian inference: analysis by synthesis?
\newblock {\em Trends in cognitive sciences}, 10(7):301--308, 2006.

\bibitem{zhang2024raydiffusion}
Jason~Y Zhang, Amy Lin, Moneish Kumar, Tzu-Hsuan Yang, Deva Ramanan, and Shubham Tulsiani.
\newblock Cameras as rays: Pose estimation via ray diffusion.
\newblock In {\em International Conference on Learning Representations (ICLR)}, 2024.

\bibitem{zhang2024monst3r}
Junyi Zhang, Charles Herrmann, Junhwa Hur, Varun Jampani, Trevor Darrell, Forrester Cole, Deqing Sun, and Ming-Hsuan Yang.
\newblock Monst3r: A simple approach for estimating geometry in the presence of motion.
\newblock {\em arXiv preprint arxiv:2410.03825}, 2024.

\bibitem{zhang2024worldWVD}
Qihang Zhang, Shuangfei Zhai, Miguel Ángel Bautista, Kevin Miao, Alexander Toshev, Joshua Susskind, and Jiatao Gu.
\newblock World-consistent video diffusion with explicit 3d modeling.
\newblock In {\em arXiv}, 2024.

\bibitem{Zhang_2018_CVPR}
Richard Zhang, Phillip Isola, Alexei~A. Efros, Eli Shechtman, and Oliver Wang.
\newblock The unreasonable effectiveness of deep features as a perceptual metric.
\newblock In {\em Proceedings of the IEEE Conference on Computer Vision and Pattern Recognition (CVPR)}, June 2018.

\bibitem{zhang2025ttsdiffusion}
Xiangcheng Zhang, Haowei Lin, Haotian Ye, James Zou, Jianzhu Ma, Yitao Liang, and Yilun Du.
\newblock Inference-time scaling of diffusion models through classical search.
\newblock {\em arXiv preprint arXiv:2505.23614}, 2025.

\bibitem{zhou2025seva}
Jensen~(Jinghao) Zhou, Hang Gao, Vikram Voleti, Aaryaman Vasishta, Chun-Han Yao, Mark Boss, Philip Torr, Christian Rupprecht, and Varun Jampani.
\newblock Stable virtual camera: Generative view synthesis with diffusion models.
\newblock {\em arXiv preprint arXiv:2503.14489}, 2025.

\bibitem{zhou2025learning}
Siyuan Zhou, Yilun Du, Yuncong Yang, Lei Han, Peihao Chen, Dit-Yan Yeung, and Chuang Gan.
\newblock Learning 3d persistent embodied world models.
\newblock {\em arXiv preprint arXiv:2505.05495}, 2025.

\bibitem{zhou2018realestate}
Tinghui Zhou, Richard Tucker, John Flynn, Graham Fyffe, and Noah Snavely.
\newblock Stereo magnification: Learning view synthesis using multiplane images.
\newblock {\em ACM Transactions on Graphics (TOG), SIGGRAPH}, 37(4):1--12, 2018.

\bibitem{zhou2025omniworld}
Yang Zhou, Yifan Wang, Jianjun Zhou, Wenzheng Chang, Haoyu Guo, Zizun Li, Kaijing Ma, Xinyue Li, et~al.
\newblock Omniworld: A multi-domain and multi-modal dataset for 4d world modeling.
\newblock {\em arXiv preprint arXiv:2509.12201}, 2025.

\bibitem{zhu2025iccv-aether}
Haoyi Zhu, Yifan Wang, Jianjun Zhou, Wenzheng Chang, Yang Zhou, Zizun Li, Junyi Chen, Chunhua Shen, Jiangmiao Pang, and Tong He.
\newblock {Aether: Geometric-Aware Unified World Modeling}.
\newblock In {\em International Conference on Computer Vision}, pages 8535--8546, 2025.

\bibitem{zhu2025motioncrafter}
Ruijie Zhu, Jiahao Lu, Wenbo Hu, Xiaoguang Han, Jianfei Cai, Ying Shan, and Chuanxia Zheng.
\newblock Motioncrafter: Dense geometry and motion reconstruction with a 4d vae.
\newblock {\em arXiv preprint arXiv:2602.08961}, 2026.

\end{thebibliography}

\newpage
\appendix

\section{Visualizations}
Comprehensive visualizations across inference settings are available on our \href{https://ugg-model.github.io/}{project page}, including camera-conditioned generations, camera pose estimation, and depth estimation across datasets.

\section{Implementation Details}
\label{app:implementation}

\subsection{Datasets and Preprocessing}
\label{app:datasets}

\paragraph{Datasets.}
We train a single model on a heterogeneous mixture of real and synthetic video datasets:
RealEstate10K~\cite{zhou2018realestate}, DL3DV-10K~\cite{ling2024dl3dv},
TartanAir~\cite{wang2020tartanair}, TartanAir V2, Spring~\cite{mehl2023spring},
BEDLAM~\cite{black2023bedlam}, BEDLAM2.0~\cite{tesch2025bedlam2},
Virtual KITTI 2~\cite{cabon2020vkitti}, Aria Synthetic Environments~\cite{avetisyan2024scenescript},
OmniWorld-Game~\cite{zhou2025omniworld}, SceneNet RGB-D~\cite{mccormac2017scenenet},
MVS-Synth~\cite{huang2018deepmvs}, Mid-Air~\cite{fonder2019midair}, and MatrixCity~\cite{li2023matrixcity}.
The datasets provide different subsets of RGB, depth, camera pose, and language supervision.
This heterogeneity is handled directly by diffusion forcing: unavailable modalities are treated as
unobserved targets and their losses are masked out, so datasets with incomplete annotations can still
contribute to the shared RGB and cross-modal representation.

\paragraph{Available modalities.}
For RealEstate10K, we use the provided RGB frames and camera poses; for DL3DV-10K, we use RGB frames
and the official COLMAP camera poses. Neither provides depth. For the synthetic datasets, we use the
rendered RGB, depth, and camera-pose annotations, except for OmniWorld-Game and MatrixCity, where we use
RGB and depth only. We use language conditioning throughout: every training clip is paired with a
scene-level text description, and a null prompt is used when no description is available.

\paragraph{Camera representation.}
We represent camera pose as a dense ray map aligned with the image plane. For each pixel $(u,v)$,
the ray map stores a Pl\"ucker ray
\begin{equation}
    R_i(u,v) = \left(\mathbf{d}_i(u,v), \mathbf{m}_i(u,v)\right),
    \qquad
    \mathbf{m}_i(u,v) = \mathbf{o}_i \times \mathbf{d}_i(u,v),
\end{equation}
where $\mathbf{o}_i$ is the camera center and $\mathbf{d}_i(u,v)$ is the unit viewing direction in
the world coordinate frame, which we define by the first camera of the clip. The ray direction encodes
orientation and intrinsics, while the ray moment encodes the camera center up to the common scene scale.
In implementation, ray maps are stored as six-channel tensors with channel order $[\mathbf{d}, \mathbf{m}]$
at $36 \times 48$ resolution and are bilinearly downsampled to the latent spatial resolution.

\paragraph{Cross-dataset normalization.}
To make supervision from different datasets comparable, we normalize all geometric quantities with
a shared clip-level scale. For each video clip, we compute a mutual scale $s$, the maximum ray-moment
magnitude over the clip (floored at $1$), and use it consistently for camera ray moments and depth;
for clips without camera supervision, $s$ is the median valid depth. Ray directions are already
unit-normalized, while ray moments are divided by $s$ and clamped to $[-1,1]$. For depth, we use a disparity-like
parameterization
\[
\tilde{D} = \frac{2}{1 + D/s} - 1,
\]
where $D$ is the raw metric or reconstructed depth. This maps near depths to values close to $1$
and far depths toward $-1$, yielding a bounded representation that is numerically compatible with
RGB and latent diffusion training. With this convention, RGB, depth, and ray-map channels are all
represented in approximately the same normalized range $[-1,1]$.

\paragraph{Missing-modality handling.}
Each training sample is represented in a common format containing RGB, ray maps, and depth.
When a dataset does not provide ray maps or depth, the corresponding tensors are filled with random
Gaussian placeholders and marked by modality-availability flags. During noise injection, missing
ray-map and depth blocks are forced to the maximum noise level, so the model never receives
structured false geometry. During loss computation, unavailable ray-map and depth targets are masked
out. This prevents RGB-only or RGB-plus-pose datasets from corrupting geometric supervision while
still allowing them to train the shared video-generation backbone.

\subsection{Model Architecture}
\label{app:model_architecture}

\paragraph{Base model.}
We initialize from Wan2.2-TI2V-5B and extend its diffusion transformer into a multimodal
Mixture-of-Transformers backbone. The pretrained video VAE and text encoder are kept frozen,
and we finetune only the diffusion transformer. The model jointly processes four latent streams,
\begin{equation}
    [z^I,\; z^{R_d},\; z^{R_m},\; z^D],
\end{equation}
corresponding to RGB video, ray directions, ray moments, and depth, respectively. The two ray
latents are concatenated along the channel dimension and embedded as a single ray token stream,
so the transformer processes three token streams: RGB, ray, and depth.

\paragraph{Tokenization.}
All clips contain $50$ frames resized to $288 \times 384$. RGB frames are encoded by the frozen
Wan2.2 VAE into latent tensors with $48$ channels and spatial size $18 \times 24$. Depth maps are
first repeated to three channels and are encoded by the same frozen VAE. Ray maps bypass the VAE:
we split each ray map into direction and moment fields and pack them directly into latent tensors
at the same spatial resolution as the VAE latents. For the default 50-frame setting, the first and
last frames each occupy their own latent frame and the $48$ interior frames are compressed $4\times$
in time, so the temporal latent length per modality is $T_{\mathrm{lat}} = 14$.
Thus, each modality stream has shape $[B,48,14,18,24]$, and the concatenated four-stream latent
has shape $[B,48,56,18,24]$. We use a 3D patch size of $[1,2,2]$, giving a per-stream patch grid
of $14 \times 9 \times 12$, or $1512$ tokens per stream. Since the two ray latents share one token
stream, the transformer operates on $3 \times 1512 = 4536$ tokens in total.

\paragraph{Ray latent packing.}
For ray directions and ray moments, four consecutive 3-channel ray frames are packed into the first
$12$ channels of a $48$-channel latent tensor, with the remaining channels padded by zeros. In the
first--last-frame setting, the first latent slot repeats the first frame, the middle slots pack groups of
interior frames, and the final latent slot repeats the last frame. The $12$ packed channels of the
direction and moment latents are concatenated into a $24$-channel input to the ray patch embedding.
This packing keeps ray-map tokens spatially and temporally aligned with RGB and depth VAE tokens while
avoiding unnecessary VAE compression of camera geometry.

\paragraph{Mixture-of-Transformers backbone.}
The transformer uses modality-specific input projections and feed-forward layers, together with
joint multimodal self-attention. Concretely, we use separate 3D patch embeddings for RGB, ray maps,
and depth. Queries, keys, values, output projections, and query/key normalization layers are
stream-specific for the RGB, ray, and depth streams, but the resulting tokens are concatenated and
attend to each other in a single multimodal attention operation. Each transformer block also contains
stream-specific feed-forward networks, while text cross-attention is shared across streams. Each stream
has its own output head. This design preserves modality-specific feature statistics while
allowing RGB, depth, and camera geometry to exchange information through shared
attention.

\paragraph{Model dimensions.}
The diffusion transformer follows the Wan2.2-TI2V-5B configuration:
hidden dimension $3072$, feed-forward dimension $14336$, $24$ attention heads, and $30$ transformer
layers. The VAE latent channel dimension is $48$, the VAE stride is $[4,16,16]$ in time, height, and
width, and the transformer patch size is $[1,2,2]$. Text conditioning uses the Wan umT5 text encoder
with prompt embeddings of dimension $4096$ and a maximum length of $512$ tokens.

\paragraph{Initialization of new modality branches.}
When initializing from the RGB-only Wan2.2 checkpoint, all compatible transformer weights are
copied from the pretrained model. The RGB patch embedding initializes the depth patch embedding,
and the RGB output head initializes the ray and depth heads. The ray patch embedding has a different
number of input channels and is trained from scratch. Attention, cross-attention, feed-forward,
time-embedding, and text-conditioning weights are copied into the corresponding
modality streams. This initialization preserves the pretrained video prior
while allowing the new modality branches to specialize during finetuning.

\subsection{Training Objective and Diffusion Forcing}
\label{app:training_details}

\paragraph{Continuous flow matching.}
We train with continuous flow matching. We sample raw diffusion times $t\in[0,1]$ for the context and target groups, as described below, and apply the shifted schedule
\begin{equation}
    t' = \frac{\alpha t}{1 + (\alpha - 1)t},
    \qquad \alpha = 3.0.
\end{equation}
Given a clean latent $z_0$ and Gaussian noise $\epsilon \sim \mathcal{N}(0,I)$, the noisy input is
\begin{equation}
    z_{t'} = (1-t')z_0 + t'\epsilon,
\end{equation}
and the model predicts the flow target
\begin{equation}
    v^\star = \epsilon - z_0.
\end{equation}
The loss is the mean-squared error between the predicted and target flow, computed separately for
RGB, ray direction, ray moment, and depth:
\begin{equation}
    \mathcal{L}
    =
    \frac{1}{4}
    \left(
        \mathcal{L}_{I}
        + \mathcal{L}_{R_d}
        + \mathcal{L}_{R_m}
        + \mathcal{L}_{D}
    \right).
\end{equation}
For samples without ray-map or depth supervision, the corresponding losses are masked out.

\paragraph{Diffusion-forcing modes.}
We specify training tasks by assigning different diffusion levels to different modality-frame groups.
Clean or low-noise tokens serve as conditioning context, while high-noise tokens are prediction
targets. We use five conditioning patterns:
\begin{enumerate}
    \item First RGB latent as history; remaining RGB, ray maps, and depth as targets;
    \item All RGB latents as history; ray maps and depth as targets;
    \item First and last RGB latents together with all ray-map latents as history; middle RGB and
          depth as targets;
    \item First RGB latent together with all ray-map latents as history; remaining RGB and depth as
          targets;
    \item First RGB latent together with all depth latents as history; remaining RGB and ray maps as
          targets.
\end{enumerate}
These patterns correspond to image-conditioned scene completion, RGB-to-geometry prediction,
camera-conditioned video generation, camera-conditioned extrapolation, and depth-conditioned
generation/prediction. We sample these modes during training and restrict them according to the
modalities available in each dataset. For example, depth-conditioned modes are used only for datasets
with depth supervision, and ray-conditioned modes are used only for datasets with camera-pose or
ray-map supervision.

\paragraph{History noise.}
For history tokens, we use clean context with probability $0.5$. Otherwise, history tokens share
a single uniformly sampled noise level. Target tokens share a separately sampled noise level. This
teaches the model to use both perfectly clean context and imperfect context, which improves robustness
when conditioning on generated or partially corrupted histories.

\paragraph{Optimization.}
We train on 50-frame clips at $288 \times 384$ resolution, sampled at a dataset-specific frame rate
between 8 and 30 FPS. Optimization uses AdamW with betas $(0.9,0.95)$, weight decay $0.05$, and
gradient clipping at $1.0$. The learning rate warms up over $250$ steps to a peak of
$4 \times 10^{-5}$ and is decayed to $6 \times 10^{-6}$ by the end of training.
Training is performed with \texttt{bf16} mixed precision and FSDP on $128$ NVIDIA H200 GPUs for
two weeks, or approximately $100{,}000$ optimizer steps. We use batch size $8$ per GPU without
gradient accumulation, giving a global batch size of $1024$. We use gradient checkpointing for
memory efficiency.

\paragraph{Dataset sampling weights.}
The training mixture uses relative dataset weights, which are multiplied by dataset size and normalized
by the mixture loader. The resulting sampling proportions are $18.8\%$ TartanAir, $15.0\%$ RealEstate10K,
$10.0\%$ TartanAir V2, $8.3\%$ Aria Synthetic Environments, $7.3\%$ BEDLAM, $7.3\%$ BEDLAM2.0,
$7.2\%$ DL3DV-10K, $6.2\%$ SceneNet RGB-D, $5.2\%$ OmniWorld-Game, $4.2\%$ Virtual KITTI 2,
$4.2\%$ MatrixCity, $2.6\%$ MVS-Synth, $2.6\%$ Mid-Air, and $1.0\%$ Spring.

\subsection{Ablation Setup}
\label{app:ablation_setup}

\paragraph{Task-removal ablations.}
When we remove a task from training, we keep the training clips
and optimization schedule unchanged. We disable its loss and
replace its input tokens with noise, keeping the remaining loss
weights fixed. When removing RGB generation, RGB remains available
as clean conditioning input. The global-camera variant uses camera poses as conditioning without predicting them. The no-pose variant is an image-only baseline, without camera conditioning or pose supervision. Depth and pose estimation use the dense evaluation setting where all RGB frames are provided as input. 

\begin{table}[H]
\centering
\caption{\textbf{History-guidance weight.}
We report ATE~($\downarrow$) and RPE$_\mathrm{r}$~($\downarrow$,
degrees) on ScanNetV2 and ScanNet++.
Shading marks the best result in each column.}
\label{tab:history_guidance}

\begingroup
\small
\setlength{\tabcolsep}{6pt}
\renewcommand{\arraystretch}{1.05}

\begin{tabular}{@{}c*{4}{r}@{}}
\toprule
& \multicolumn{2}{c}{ScanNetV2}
& \multicolumn{2}{c}{ScanNet++} \\
\cmidrule(lr){2-3} \cmidrule(l){4-5}
$w_{\mathrm{hist}}$
& ATE & RPE$_\mathrm{r}$
& ATE & RPE$_\mathrm{r}$ \\
\midrule
0 & 0.0796 & \best{2.46} & 0.2170 & \best{5.77} \\
1 & 0.0741 & 2.83 & \best{0.2089} & 6.09 \\
2 & \best{0.0740} & 3.24 & 0.2091 & 6.45 \\
4 & 0.0765 & 4.22 & 0.2141 & 7.70 \\
\bottomrule
\end{tabular}

\endgroup
\end{table}

\paragraph{History guidance.}
\label{app:history_guidance}
History guidance improves trajectory alignment at the cost
of rotation accuracy. Moderate weights
($w_{\mathrm{hist}}\in\{1,2\}$) reduce ATE relative to
unguided inference, but rotational RPE increases with
guidance strength (Table~\ref{tab:history_guidance}).
Increasing the weight to 4 worsens both metrics relative
to moderate guidance.

\subsection{Inference}
\label{app:inference_details}

\paragraph{Sampling.}
At inference time, the same clean/noisy pattern specifies the task. Observed RGB, depth, or ray-map
tokens are clamped to their clean values at every denoising step and assigned zero diffusion level.
Unknown tokens are initialized from Gaussian noise and progressively denoised. We use UniPC sampling with $40$ denoising steps and the same shifted time schedule used during training. Under this setting, the denoising loop for one 50-frame clip takes approximately 20 seconds on a single NVIDIA H100 or H200 GPU.

\paragraph{History guidance.}
We use history guidance to strengthen consistency with observed context. For a set of observed
history tokens $H$, we compute a history-conditioned flow $\hat{v}_{H}$ with $H$ clamped to the
observed values and a history-dropped flow $\hat{v}_{\emptyset}$ in which the same context is replaced
by noise. The guided flow is
\begin{equation}
    \hat{v}_{\mathrm{HG}}
    =
    \hat{v}_{H}
    +
    w_{\mathrm{hist}}
    \left(\hat{v}_{H} - \hat{v}_{\emptyset}\right).
\end{equation}
This is analogous to classifier-free guidance, but the dropped condition is the observed history/context rather than the text prompt. We set $w_{\mathrm{hist}}=2.0$ in our default configuration.

\paragraph{Decoding.}
After denoising, RGB and depth latents are decoded with the frozen Wan VAE. Ray-direction and
ray-moment latents are unpacked into dense ray maps. For camera-pose evaluation, the predicted
ray maps are converted back to camera parameters using the same ray-map-to-camera recovery
procedure across all datasets.

\paragraph{Inference-time search.}
For test-time search, we sample multiple candidate completions from different random seeds while
clamping the same observed context. Each candidate produces RGB, depth, and ray-map predictions.
We score candidates against RAFT~\citep{teed2020raft} correspondences between the observed frames,
without ground-truth depth or camera pose: depth candidates by how well their scale-and-shift-aligned
depth explains the observed correspondences, and pose candidates by the agreement between their relative
translation directions and those estimated from the correspondences. This implements the
inference-time search procedure described in Sec.~\ref{sec:inference_time_search}.

\subsection{Depth Evaluation}
\label{app:depth_evaluation}

We decode our disparity predictions to depth before evaluation.
All methods use the same affine-invariant evaluation protocol:
we fit a single scale and shift per clip by minimizing an L1
objective over valid pixels pooled across the scored frames,
\begin{equation}
(s^*,t^*)=\arg\min_{s,t}
\sum_{p\in\mathcal V}|s\hat d_p+t-d_p|,
\qquad
\widetilde d_p=\max(s^*\hat d_p+t^*,10^{-6}),
\end{equation}
where $\hat d_p$ and $d_p$ are predicted and ground-truth depths.
Ground-truth depths are used in metres without additional
normalization. The shared fit across frames preserves sensitivity
to temporal depth consistency.

We report absolute relative error and threshold accuracy:
\begin{align}
\mathrm{AbsRel}
&=\frac{1}{|\mathcal V|}
\sum_{p\in\mathcal V}
\frac{|\widetilde d_p-d_p|}{d_p},\\
\delta_{<1.25}
&=\frac{1}{|\mathcal V|}
\sum_{p\in\mathcal V}
\mathbf{1}\!\left[
\max\!\left(
\frac{\widetilde d_p}{d_p},
\frac{d_p}{\widetilde d_p}
\right)<1.25
\right].
\end{align}
Both metrics average valid pixels jointly across the clip,
using a strict threshold for $\delta_{<1.25}$.

Dense evaluation scores all 50 frames; sparse evaluation fits
and scores only the first and last observed frames.
On Sintel, Bonn, and KITTI, dense evaluation uses a
$240\times320$ grid, while sparse evaluation uses the
cropped ground-truth resolution.
We report an unweighted mean of per-clip scores for each dataset.

\subsection{Camera Pose Evaluation}
\label{app:pose_evaluation}

\paragraph{Dense setting.}
In the dense setting, models condition on the full RGB sequence.
Following Geo4D~\cite{jiang2025geo4d}, we align predicted trajectories
to ground truth using Sim(3) Umeyama alignment. We report absolute
trajectory error (ATE), relative pose error in translation
(RPE$_\mathrm{t}$), and in rotation (RPE$_\mathrm{r}$).
ATE measures global trajectory agreement after alignment, while
RPE measures local frame-to-frame motion consistency.

\paragraph{Sparse setting.}
In the sparse setting, models condition on only the first and last
RGB frames, and we evaluate their relative pose. With only two
camera centers, Sim(3) alignment can match both positions exactly,
making the standard aligned ATE uninformative. For sparse setting
results we therefore report translation angular error (tAE),
measuring the angular difference between the predicted and ground
truth relative translation directions, and RPE$_\mathrm{r}$ for
relative rotation error.

Let $T_i=[R_i\mid\mathbf c_i]$ denote camera-to-world poses.
For dense evaluation, we align each predicted trajectory to ground
truth using a single Sim(3) Umeyama fit over all camera centers,
yielding aligned predictions $\widetilde T_i$. We define
\begin{equation}
\Delta_{ij}=T_i^{-1}T_j,\qquad
E_{ij}=\Delta_{ij}^{-1}
(\widetilde T_i^{-1}\widetilde T_j).
\end{equation}
Following~\cite{sturm2012tumdynamics}, we report
\begin{align}
\mathrm{ATE}
&=\sqrt{\frac{1}{N}\sum_{i=0}^{N-1}
\|\widetilde{\mathbf c}_i-\mathbf c_i\|_2^2},\\
\mathrm{RPE}_{\mathrm t}
&=\sqrt{\frac{1}{N-1}\sum_{i=0}^{N-2}
\|\mathbf t(E_{i,i+1})\|_2^2},\\
\mathrm{RPE}_{\mathrm r}
&=\sqrt{\frac{1}{N-1}\sum_{i=0}^{N-2}
\theta(R(E_{i,i+1}))^2},
\end{align}
where $\mathbf t(E)$ and $R(E)$ are the translation and rotation
components, and
$\theta(R)=\frac{180}{\pi}
\arccos((\operatorname{tr}(R)-1)/2)$.
Translation errors remain in ground-truth units, without
additional normalization.

For sparse evaluation, we use the unaligned first--last relative
poses $\Delta_{0,N-1}=[R\mid\mathbf t]$ and
$\widehat\Delta_{0,N-1}=[\widehat R\mid\widehat{\mathbf t}]$:
\begin{equation}
\mathrm{RPE}_{\mathrm r}=\theta(R^\top\widehat R),
\qquad
\mathrm{tAE}=\frac{180}{\pi}\arccos\!\left(
\frac{\widehat{\mathbf t}^{\top}\mathbf t}
{\|\widehat{\mathbf t}\|_2\|\mathbf t\|_2}
\right).
\end{equation}
Both are invariant to global similarity transformations, and tAE
lies in $[0^\circ,180^\circ]$. All inverse-cosine arguments are
clipped to $[-1,1]$. We report an unweighted mean of per-clip
scores for each dataset.

\subsection{Inference-time Search Details}
\label{app:tts}

This section gives the equations and hyperparameters of the search in Sec.~\ref{sec:inference_time_search}. The verifier uses only the observed RGB frames, RAFT-Large optical flow~\citep{teed2020raft}, DINOv2 ViT-S/14 features~\citep{oquab2023dinov2}, and each candidate's own predicted depth, cameras and intrinsics. It never uses ground-truth depth or pose, never uses another depth or pose model, and never averages or compares candidates against each other except when ranking them. All computation is at the model resolution ($288\times384$), and residuals are evaluated on a stride-2 pixel grid.

\paragraph{Conventions.}
Cameras follow the OpenCV convention ($x$ right, $y$ down, $z$ forward) and pixel coordinates are at $288\times384$. For a frame pair $(i,j)$, the relative camera $(\mathbf{R}_{ij},\mathbf{t}_{ij})$ maps camera-$i$ coordinates to camera-$j$ coordinates, $\mathbf{X}_j=\mathbf{R}_{ij}\mathbf{X}_i+\mathbf{t}_{ij}$. For a candidate this is $\hat{\mathbf{T}}_{ij}=\hat{\mathbf{T}}_j^{-1}\hat{\mathbf{T}}_i$, where $\hat{\mathbf{T}}_i$ is its predicted camera-to-world matrix. Image-estimated translations are defined only up to scale and have unit norm.

\paragraph{Forward-backward consistency.}
RAFT is run in both directions to give flows $\mathbf{F}_{ij}$ and $\mathbf{F}_{ji}$. The correspondence $\mathbf{x}\!\mapsto\!\mathbf{x}'=\mathbf{x}+\mathbf{F}_{ij}(\mathbf{x})$ is kept only if $\mathbf{x}'$ lies inside frame $j$ and the round trip returns close to $\mathbf{x}$~\citep{sundaram2010dense}:
\begin{equation}
\big\|\mathbf{F}_{ij}(\mathbf{x})+\mathbf{F}_{ji}(\mathbf{x}')\big\|^2
\le 0.01\big(\|\mathbf{F}_{ij}(\mathbf{x})\|^2+\|\mathbf{F}_{ji}(\mathbf{x}')\|^2\big)+0.5 .
\label{eq:tts_fb}
\end{equation}
A pixel whose match in frame $j$ maps back somewhere else is occluded, moving, or a flow failure. Such pixels are discarded before any camera estimate or residual is computed.

\paragraph{Frame-pair selection.}
A fixed frame gap gives almost no parallax on slow video and breaks RAFT on fast video. For dense observations, we therefore choose each gap so that the pair has a fixed target motion $\tau$:
\begin{equation}
g(i)=\mathrm{clip}\!\Big(\mathrm{round}\big(\tau\,/\,\mathrm{med}_{\mathbf{x}}\|\mathbf{F}_{i,i+1}(\mathbf{x})\|\big),\,1,\,12\Big),
\label{eq:tts_gap}
\end{equation}
with $\tau=16$\,px for every task except dense pose, where $\tau=32$\,px. The same values are used on every benchmark.
\emph{Dense depth}: 12 anchors $i$ spaced evenly over the clip, each paired with $i+g(i)$.
\emph{Dense pose}: 16 anchors, each used in both directions (32 pairs).
\emph{Sparse depth}: only the two given frames are used. Correspondences are computed directly between them, and only the candidate's depth predictions at the observed frames are scored.
\emph{Sparse pose}: only the two given frames are used, testing the candidate's predicted first--last relative camera.

\paragraph{Relative camera from RGB (RANSAC).}
On up to 6{,}000 consistent correspondences (subsampled uniformly) we estimate the essential matrix with the five-point algorithm inside RANSAC (OpenCV \texttt{findEssentialMat}: confidence 0.999, 1\,px threshold). We then decompose it with \texttt{recoverPose}, which picks the solution that places points in front of both cameras. For \emph{depth}, this camera is computed once per pair and shared by all candidates, so the candidates differ only in their depth. Using each candidate's own predicted camera instead put the ranking at chance, because a wrong camera can be compensated by a wrong depth. For \emph{pose}, the reference camera is recomputed with each candidate's own predicted intrinsics $\hat{\mathbf{K}}$.

\paragraph{Depth residuals.}
For pair $(i,j)$ and candidate inverse depth $w$ at frame $i$, let $\mathbf{f}_{\mathrm{rot}}$ be the flow induced by rotation alone (depth $\to\infty$). We linearize reprojection around zero inverse depth as
$\mathbf{x}'(w)\approx\mathbf{x}+\mathbf{f}_{\mathrm{rot}}+w\mathbf{b}$,
where $\mathbf{b}=\left.\partial\mathbf{x}'(w)/\partial w\right|_{w=0}$
depends only on the shared camera and pixel position.
Let $\mathbf{c}=w\mathbf{b}$ be the candidate's predicted parallax,
used for parallax filtering, and let
$\hat{\mathbf{b}}=\mathbf{b}/\|\mathbf{b}\|$.
Let $\mathbf{p}=\mathbf{F}_{ij}-\mathbf{f}_{\mathrm{rot}}$ be the observed parallax.
We split $\mathbf{p}$ into $a=\mathbf{p}\!\cdot\!\hat{\mathbf{b}}$,
the part a depth can explain, and the perpendicular remainder
$e_\perp=\|\mathbf{p}-a\hat{\mathbf{b}}\|$, the part no depth can explain.
\emph{Gauge}: with $\hat w=w/\mathrm{med}(w)$, a closed-form weighted least-squares fit gives the scale and shift
$(\alpha,\beta)=\arg\min_{\alpha,\beta}\sum_{\mathbf{x}}\omega(\mathbf{x})
\|\mathbf{p}-(\alpha\hat w+\beta)\,\mathbf{b}\|^2$,
where the sum is over kept pixels and
$\omega(\mathbf{x})=\|\nabla I_i(\mathbf{x})\|$.
We write the aligned inverse depth as $\tilde w=\alpha\hat w+\beta$.
\emph{Epipolar filtering}: we keep only pixels that pass Eq.~\ref{eq:tts_fb}, have predicted parallax $\|\mathbf{c}\|>0.5$\,px, and have $e_\perp$ at or below its 80th percentile in the pair.The epipolar residual is used only for filtering. Kept pixels are weighted by image gradient magnitude $\|\nabla I_i\|$.
The aligned warp $\mathbf{x}'=\mathbf{x}+\mathbf{f}_{\mathrm{rot}}+\tilde w\,\mathbf{b}$
defines three pointwise residuals:
\begin{align}
r_{\mathrm{corr}}(\mathbf{x}) &= \big|a-\tilde w\|\mathbf{b}\|\big|, \nonumber\\
r_{\mathrm{photo}}(\mathbf{x}) &= \big\|I_i(\mathbf{x})-I_j(\mathbf{x}')\big\|_1, \nonumber\\
r_{\mathrm{feat}}(\mathbf{x}) &= 1-\cos\!\big(\phi_i(\mathbf{x}),\phi_j(\mathbf{x}')\big).
\label{eq:tts_res}
\end{align}
Each residual is aggregated as an image-gradient-weighted mean over kept pixels.

where $\phi$ are DINOv2 ViT-S/14 patch features, bilinearly upsampled to image resolution and sampled at $\mathbf{x}'$. Sparse depth uses $r_{\mathrm{corr}}$.. Dense depth uses $\{r_{\mathrm{corr}},r_{\mathrm{feat}},r_{\mathrm{photo}}\}$.

\paragraph{Pose residual.}
For each pair, $\theta^{(\ell)}_{ij}=\angle\big(\hat{\mathbf{t}}^{(\ell)}_{ij},\,\mathbf{t}^{\mathrm{RGB}}_{ij}\big)$ in degrees, where $\mathbf{t}^{\mathrm{RGB}}_{ij}$ is the RANSAC translation computed with candidate $\ell$'s intrinsics.

\paragraph{Selection score.}
Depth and pose are selected \emph{independently}. There is no weighting between them, and the two results can come from different candidates.
\emph{Depth.} For pair $p$, each residual $k$ in the rule set $\mathcal{K}$ is z-normalized across the $L$ candidates. The normalized residuals are averaged into $s^{(\ell)}_p=\frac{1}{|\mathcal{K}|}\sum_{k}z_p^{(\ell)}(r_k)$, and candidates are ranked within the pair, $\rho^{(\ell)}_p\in[0,1]$. Pairs are weighted by how strongly they separate the candidates, $\omega_p=\mathrm{IQR}_\ell(s^{(\ell)}_p)$, because a pair on which all candidates score alike carries no information:
\begin{equation}
E^{(\ell)}_{\mathrm{depth}}=\frac{\sum_p\omega_p\,\rho^{(\ell)}_p}{\sum_p\omega_p}.
\label{eq:tts_edepth}
\end{equation}
Using ranks means a single pathological pair cannot dominate.
\emph{Pose.}
\begin{equation}
E^{(\ell)}_{\mathrm{pose}}=\frac{1}{|\mathcal{P}|}\sum_{(i,j)\in\mathcal{P}}\theta^{(\ell)}_{ij}.
\label{eq:tts_epose}
\end{equation}
The selected candidate is $\ell^*=\arg\min_\ell E^{(\ell)}_{\mathrm{task}}$. If no pair survives the filtering for a scene, the verifier abstains and the scene is scored as a uniformly random pick from the pool.

\paragraph{Choosing the rules.}
The residual set and the target $\tau$ for each task and protocol were chosen by leave-one-benchmark-out: the rule is selected on the other benchmarks and evaluated on the held-out one. Sparse depth and sparse pose select the same rule in every fold.

\section{Existing Assets and Licenses}
\label{app:assets_licenses}

We use existing public datasets and pretrained model assets. Table~\ref{tab:assets_licenses}
summarizes the role of each asset in this work and the corresponding license or access terms.
All datasets are used for academic research purposes only. We do not redistribute the original
datasets, videos, game assets, or other third-party data. Model checkpoints are
released under terms compatible with the upstream data and
model licenses.

\begin{table}[t]
\centering
\caption{\textbf{Datasets and pretrained assets.}
We summarize the role of each asset and its license or access terms.
License names are abbreviated, original agreements and third-party restrictions apply.}
\label{tab:assets_licenses}
\begingroup
\fontsize{9}{11}\selectfont
\setlength{\tabcolsep}{5pt}
\renewcommand{\arraystretch}{1.0}
\setlength{\aboverulesep}{0.4ex}
\setlength{\belowrulesep}{0.65ex}
\begin{tabularx}{\linewidth}{@{}
  >{\raggedright\arraybackslash}p{0.27\linewidth}
  >{\raggedright\arraybackslash}p{0.31\linewidth}
  >{\raggedright\arraybackslash}X@{}}
\toprule
\textbf{Asset} & \textbf{Use in this work} & \textbf{License / access terms} \\
\midrule
\multicolumn{3}{@{}l}{\textbf{Training datasets}} \\
\addlinespace[3pt]
DL3DV-10K~\cite{ling2024dl3dv}
& RGB and pose; generation evaluation
& \mbox{CC BY-NC 4.0}; dataset terms; research and education only \\
\addlinespace[3pt]
RealEstate10K~\cite{zhou2018realestate}
& RGB and pose; generation and pose evaluation
& \mbox{CC BY 4.0} \\
\addlinespace[3pt]
TartanAir / TartanAir V2~\cite{wang2020tartanair}
& Synthetic RGB, depth, and pose
& \mbox{CC BY 4.0} \\
\addlinespace[3pt]
Spring~\cite{mehl2023spring}
& Synthetic RGB, depth, and pose
& \mbox{CC BY 4.0} \\
\addlinespace[3pt]
BEDLAM~\cite{black2023bedlam}
& Human-centric RGB, depth, and pose
& BEDLAM license; noncommercial research and education \\
\addlinespace[3pt]
BEDLAM2.0~\cite{tesch2025bedlam2}
& Human-centric RGB, depth, and pose
& Perceiving Systems License 1.0; noncommercial research; third-party terms \\
\addlinespace[3pt]
Virtual KITTI 2~\cite{cabon2020vkitti}
& Driving RGB, depth, and pose
& \mbox{CC BY-NC-SA 3.0} \\
\addlinespace[3pt]
Aria Synthetic Environments~\cite{avetisyan2024scenescript}
& Indoor RGB, depth, and pose
& Aria dataset agreement; noncommercial research \\
\addlinespace[3pt]
OmniWorld-Game~\cite{zhou2025omniworld}
& Game-scene RGB and depth
& \mbox{CC BY-NC-SA 4.0}; additional game-asset terms and notices \\
\addlinespace[3pt]
SceneNet RGB-D~\cite{mccormac2017scenenet}
& Indoor RGB, depth, and pose
& SceneNet terms; noncommercial use \\
\addlinespace[3pt]
MVS-Synth~\cite{huang2018deepmvs}
& Urban RGB, depth, and pose
& Research release; GTA V terms apply \\
\addlinespace[3pt]
Mid-Air~\cite{fonder2019midair}
& Drone RGB, depth, and pose
& \mbox{CC BY-NC-SA 4.0} \\
\addlinespace[3pt]
MatrixCity~\cite{li2023matrixcity}
& City-scale RGB and depth
& \mbox{CC BY-NC 4.0} \\
\addlinespace[2pt]
\midrule
\multicolumn{3}{@{}l}{\textbf{Evaluation-only datasets}} \\
\addlinespace[3pt]
SpatialVID~\cite{wang2025spatialvid}
& Camera-controlled generation
& \mbox{CC BY-NC-SA 4.0} \\
\addlinespace[3pt]
ScanNet++~\cite{yeshwanth2023scannetpp}
& Indoor pose and depth prediction
& ScanNet++ terms; noncommercial research and education; redistribution restricted \\
\addlinespace[3pt]
7-Scenes~\cite{shotton2013scene}
& Indoor RGB-D depth prediction
& Microsoft 7-Scenes agreement; noncommercial use \\
\addlinespace[2pt]
\midrule
\multicolumn{3}{@{}l}{\textbf{Pretrained model}} \\
\addlinespace[3pt]
Wan2.2-TI2V-5B~\cite{wan2025wan}
& Backbone initialization; transformer finetuning; frozen VAE and text encoder
& Apache 2.0 \\
\bottomrule
\end{tabularx}
\par\vspace{4pt}
\begin{minipage}{\linewidth}
\fontsize{8}{10}\selectfont
\raggedright
CC: Creative Commons; BY: attribution; NC: noncommercial; SA: share-alike.
\end{minipage}
\endgroup
\end{table}

\clearpage
\section*{NeurIPS Paper Checklist}

\begin{enumerate}

\item {\bf Claims}
    \item[] Question: Do the main claims made in the abstract and introduction accurately reflect the paper's contributions and scope?
    \item[] Answer: \answerYes{}
    \item[] Justification: The abstract and introduction state the paper’s intended scope: a unified generative-predictive model over RGB, depth, and camera pose for cross-modal generation, geometric prediction, and inference-time search. These claims are evaluated in Sec.~\ref{sec:experiments} through camera-controlled generation, camera pose estimation, depth prediction, and test-time search experiments.

\item {\bf Limitations}
    \item[] Question: Does the paper discuss the limitations of the work performed by the authors?
    \item[] Answer: \answerYes{}
    \item[] Justification: The paper includes a limitations discussion in Sec.~\ref{sec:conclusion}. It states that specialized methods remain stronger on some pixel-aligned novel-view-synthesis metrics and indoor depth metrics, that RGB reprojection is an imperfect proxy for test-time search, and that the current model is trained at moderate scale and resolution.

\item {\bf Theory assumptions and proofs}
    \item[] Question: For each theoretical result, does the paper provide the full set of assumptions and a complete (and correct) proof?
    \item[] Answer: \answerNA{}
    \item[] Justification: The paper does not present theoretical results, theorems, or formal proofs. The mathematical notation is used to define the scene representation, diffusion objective, and inference procedure.

    \item {\bf Experimental result reproducibility}
    \item[] Question: Does the paper fully disclose all the information needed to reproduce the main experimental results of the paper to the extent that it affects the main claims and/or conclusions of the paper (regardless of whether the code and data are provided or not)?
    \item[] Answer: \answerYes{}
    \item[] Justification: The paper discloses the model representation, architecture, training objective, conditioning patterns, datasets, evaluation settings, metrics, baselines, and test-time search procedure. Appendix~\ref{app:implementation} further specifies preprocessing, tokenization, model dimensions, optimization hyperparameters, sampling, and inference details needed to reproduce the main experimental setup.

\item {\bf Open access to data and code}
    \item[] Question: Does the paper provide open access to the data and code, with sufficient instructions to faithfully reproduce the main experimental results, as described in supplemental material?
    \item[] Answer: \answerYes{}
    \item[] Justification: The current submission provides all open-source code, model checkpoints, reproduction scripts at the project website. The experiments use existing datasets and pretrained model assets cited in the paper; access to those assets follows the terms of their original providers.

\item {\bf Experimental setting/details}
    \item[] Question: Does the paper specify all the training and test details (e.g., data splits, hyperparameters, how they were chosen, type of optimizer) necessary to understand the results?
    \item[] Answer: \answerYes{}
    \item[] Justification: Sec.~\ref{sec:experiments} and Appendix~\ref{app:implementation} specify the training datasets, modality availability, preprocessing, normalization, model architecture, diffusion-forcing modes, optimizer, learning rate, batch size, gradient accumulation, sampling procedure, evaluation datasets, metrics, and baselines. These details are sufficient to understand the experimental protocol and the reported results.

\item {\bf Experiment statistical significance}
    \item[] Question: Does the paper report error bars suitably and correctly defined or other appropriate information about the statistical significance of the experiments?
    \item[] Answer: \answerNo{}
    \item[] Justification: The paper reports standard benchmark metrics across multiple datasets and tasks, but does not report error bars, confidence intervals, or statistical significance tests. Repeated full training runs are computationally expensive for the 5B-parameter generative model, so the paper instead supports its claims through evaluation across diverse datasets, dense/sparse settings, and multiple tasks.

\item {\bf Experiments compute resources}
    \item[] Question: For each experiment, does the paper provide sufficient information on the computer resources (type of compute workers, memory, time of execution) needed to reproduce the experiments?
    \item[] Answer: \answerYes{}
    \item[] Justification: Appendix~\ref{app:implementation} reports the training resolution, clip length, optimizer settings, mixed-precision training, number and type of GPUs, optimizer steps, batch size, gradient accumulation, gradient checkpointing, and inference-time sampling cost. These details describe the main compute resources needed to reproduce training and inference.
    
\item {\bf Code of ethics}
    \item[] Question: Does the research conducted in the paper conform, in every respect, with the NeurIPS Code of Ethics \url{https://neurips.cc/public/EthicsGuidelines}?
    \item[] Answer: \answerYes{}
    \item[] Justification: The research conforms to the NeurIPS Code of Ethics. It uses existing datasets and pretrained model assets, does not involve new human-subject experiments or crowdsourcing, and reports assumptions, limitations, and evaluation results transparently.

\item {\bf Broader impacts}
    \item[] Question: Does the paper discuss both potential positive societal impacts and negative societal impacts of the work performed?
    \item[] Answer: \answerYes{}
    \item[] Justification: This work is a technical study of unified RGB, depth, and camera-pose modeling for 4D scene understanding, evaluated on standard research benchmarks. It does not introduce a deployed system, collect new human-subject data, or release a new dataset. We therefore do not anticipate substantial direct negative societal impacts beyond general risks associated with generative visual models and geometric scene reconstruction.
    
\item {\bf Safeguards}
    \item[] Question: Does the paper describe safeguards that have been put in place for responsible release of data or models that have a high risk for misuse (e.g., pre-trained language models, image generators, or scraped datasets)?
    \item[] Answer: \answerYes{}
    \item[] Justification: The release includes intended-use documentation and responsible-use guidance, warning about inaccurate geometry and synthetic content and advising against deceptive, privacy-invasive, or unvalidated safety-critical uses.

\item {\bf Licenses for existing assets}
    \item[] Question: Are the creators or original owners of assets (e.g., code, data, models), used in the paper, properly credited and are the license and terms of use explicitly mentioned and properly respected?
    \item[] Answer: \answerYes{}
    \item[] Justification: Appendix~\ref{app:assets_licenses} summarizing all existing datasets and pretrained model assets used in the work, including their citations, role in the experiments, and license or access terms. We use the assets for academic research, comply with non-commercial and attribution requirements where applicable, and do not redistribute third-party datasets, videos, game assets, or restricted source media. Our release of code or checkpoints is made only under terms compatible with the upstream asset licenses.

\item {\bf New assets}
    \item[] Question: Are new assets introduced in the paper well documented and is the documentation provided alongside the assets?
    \item[] Answer: \answerYes{}
    \item[] Justification: We release UGG source code and model weights at \url{https://github.com/amanikiruga/UGG}. The accompanying README and model card document installation, dependencies, supported tasks, inference examples, training configurations, hardware requirements, licensing, and known limitations.

\item {\bf Crowdsourcing and research with human subjects}
    \item[] Question: For crowdsourcing experiments and research with human subjects, does the paper include the full text of instructions given to participants and screenshots, if applicable, as well as details about compensation (if any)? 
    \item[] Answer: \answerNA{}
    \item[] Justification: The paper does not involve crowdsourcing experiments or newly conducted research with human subjects. It uses existing datasets and pretrained assets.

\item {\bf Institutional review board (IRB) approvals or equivalent for research with human subjects}
    \item[] Question: Does the paper describe potential risks incurred by study participants, whether such risks were disclosed to the subjects, and whether Institutional Review Board (IRB) approvals (or an equivalent approval/review based on the requirements of your country or institution) were obtained?
    \item[] Answer: \answerNA{}
    \item[] Justification: The paper does not conduct new human-subject studies or crowdsourcing experiments, so IRB or equivalent review is not applicable.

\item {\bf Declaration of LLM usage}
    \item[] Question: Does the paper describe the usage of LLMs if it is an important, original, or non-standard component of the core methods in this research? Note that if the LLM is used only for writing, editing, or formatting purposes and does \emph{not} impact the core methodology, scientific rigor, or originality of the research, declaration is not required.
    \item[] Answer: \answerNA{}
    \item[] Justification: The core method does not use LLMs or agents as an important, original, or non-standard component. Any writing, editing, or formatting assistance does not affect the scientific methodology, rigor, or originality of the work.

\end{enumerate}

\end{document}